\documentclass[10pt,twocolumn,letterpaper]{article}

\usepackage{iccv}
\usepackage{times}
\usepackage{epsfig}
\usepackage{graphicx}
\usepackage{amsmath}
\usepackage{amssymb}
\usepackage{placeins}
\usepackage{subcaption}
\usepackage{siunitx}

\usepackage[
backend=biber, 
style=numeric,
isbn=false,
url=false,
doi=false,
sorting=none,
]{biblatex}
\usepackage[pagebackref=false,breaklinks=true,letterpaper=true,colorlinks,bookmarks=false]{hyperref}

\usepackage{algpseudocode}
\usepackage[linesnumbered,ruled]{algorithm2e}

\iccvfinalcopy 

\def\iccvPaperID{8} 

\ificcvfinal\fi

\begin{document}

\title{Classification robustness to common optical aberrations}

\author{Patrick Müller\textsuperscript{1,2}, Alexander Braun\textsuperscript{1}, Margret Keuper\textsuperscript{2,3}\\
\small\textsuperscript{1}Hochschule Düsseldorf - University of Applied Sciences, \\
\small\textsuperscript{2}University of Siegen, \textsuperscript{3}Max-Planck-Institute for Informatics, Saarland Informatics Campus\normalfont
}

\maketitle
\ificcvfinal\thispagestyle{empty}\fi

\begin{abstract}
Computer vision using deep neural networks (DNNs) has brought about seminal changes in people's lives. Applications range from automotive, face recognition in the security industry, to industrial process monitoring. In some cases, DNNs infer even in safety-critical situations. 
Therefore, for practical applications, DNNs have to behave in a robust way to disturbances such as noise, pixelation, or blur. Blur directly impacts the performance of DNNs, which are often approximated as a disk-shaped kernel to model defocus. However, optics suggests that there are different kernel shapes depending on wavelength and location caused by optical aberrations. In practice, as the optical quality of a lens decreases, such aberrations increase.
This paper proposes \emph{OpticsBench}, a benchmark for investigating robustness to realistic, practically relevant optical blur effects. 
Each corruption represents an optical aberration (coma, astigmatism, spherical, trefoil) derived from Zernike Polynomials.
%
Experiments on ImageNet show that for a variety of different pre-trained DNNs, the performance varies strongly compared to disk-shaped kernels, indicating the necessity of considering realistic image degradations. 
In addition, we show on ImageNet-100 with \emph{OpticsAugment} that robustness can be increased by using optical kernels as data augmentation. Compared to a conventionally trained ResNeXt50, training with OpticsAugment achieves an average performance gain of 21.7\% points on OpticsBench and 6.8\% points on 2D common corruptions.
\end{abstract}

\section{Introduction}
\label{sec:introduction}
Deep neural networks (DNN) are present in peoples' daily lives in various applications to mobility~\cite{michaelis_benchmarking_2020,yu_bdd100k_2020,caesar_nuscenes_2020}, language and speech processing~\cite{hinton_deep_2012} and vision~\cite{chai_deep_2021,yuille_deep_2021}.
\newcommand{\kernelSize}{0.13}
\begin{figure}[h]
    \centering
    \begin{subfigure}{\kernelSize\linewidth}
    \includegraphics[width=\linewidth]{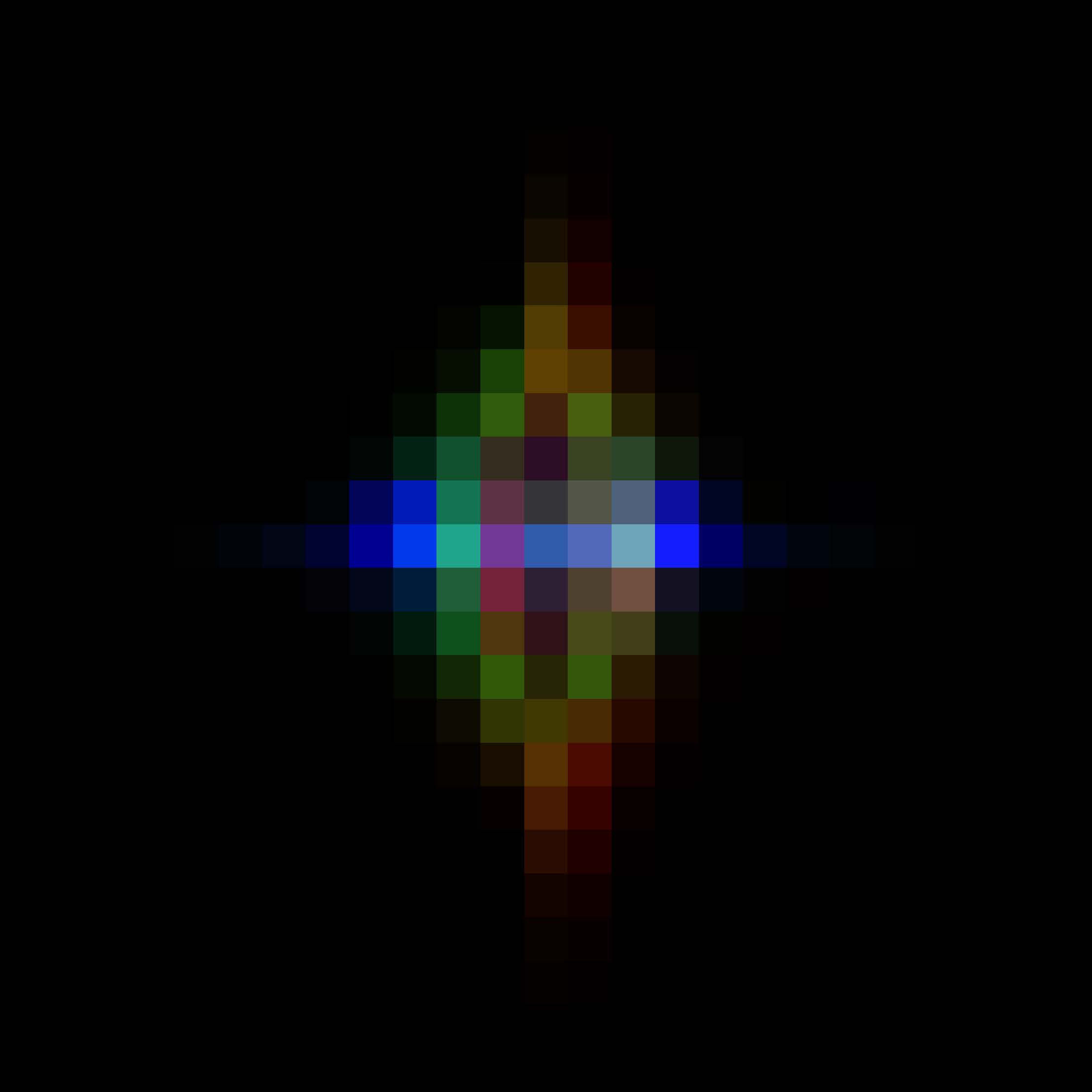}
    \caption{}
    \end{subfigure}
    \begin{subfigure}{\kernelSize\linewidth}
    \includegraphics[width=\linewidth]{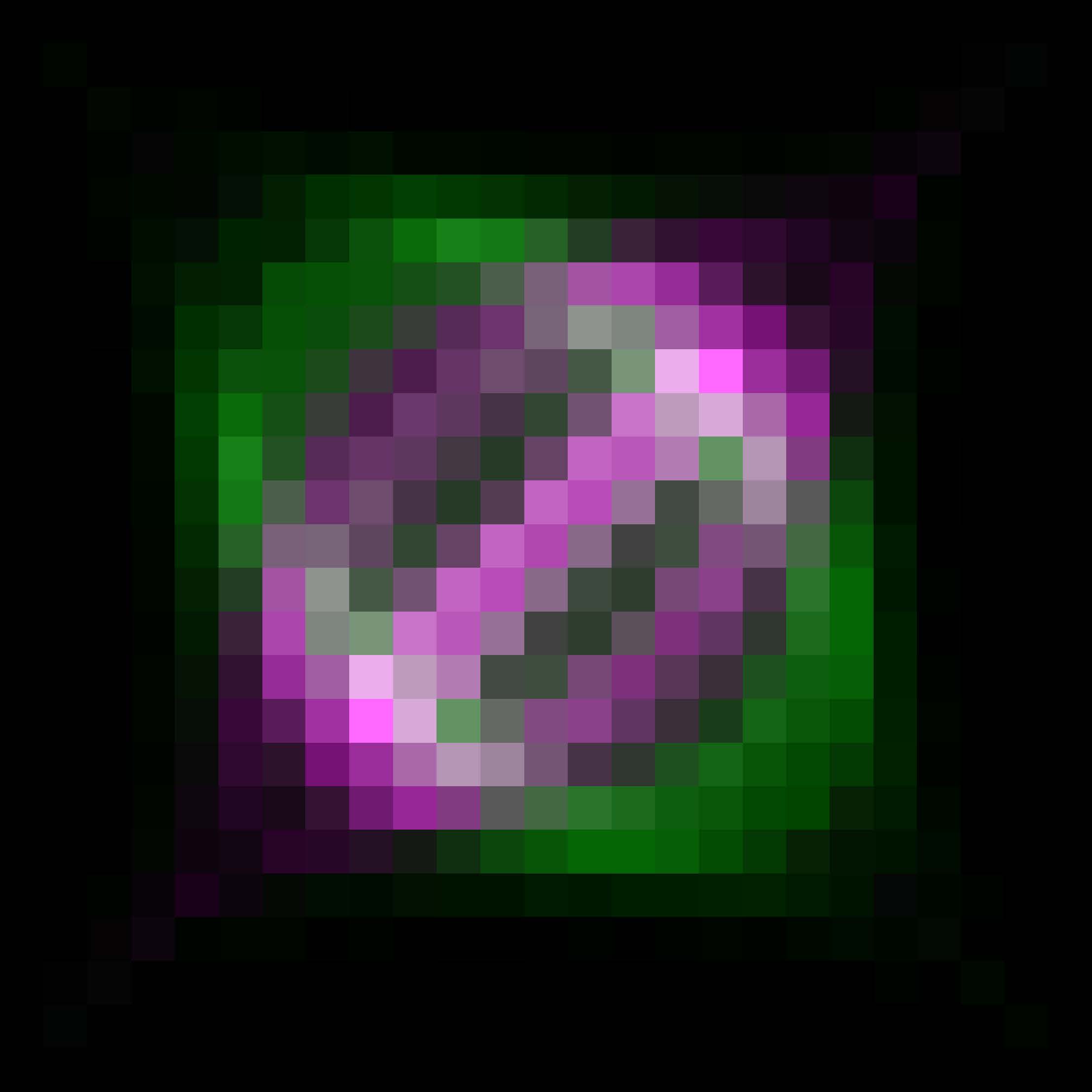}
    \caption{}\label{fig:psf_astigmatism_rg}
    \end{subfigure}
    \begin{subfigure}{\kernelSize\linewidth}
    \includegraphics[width=\linewidth]{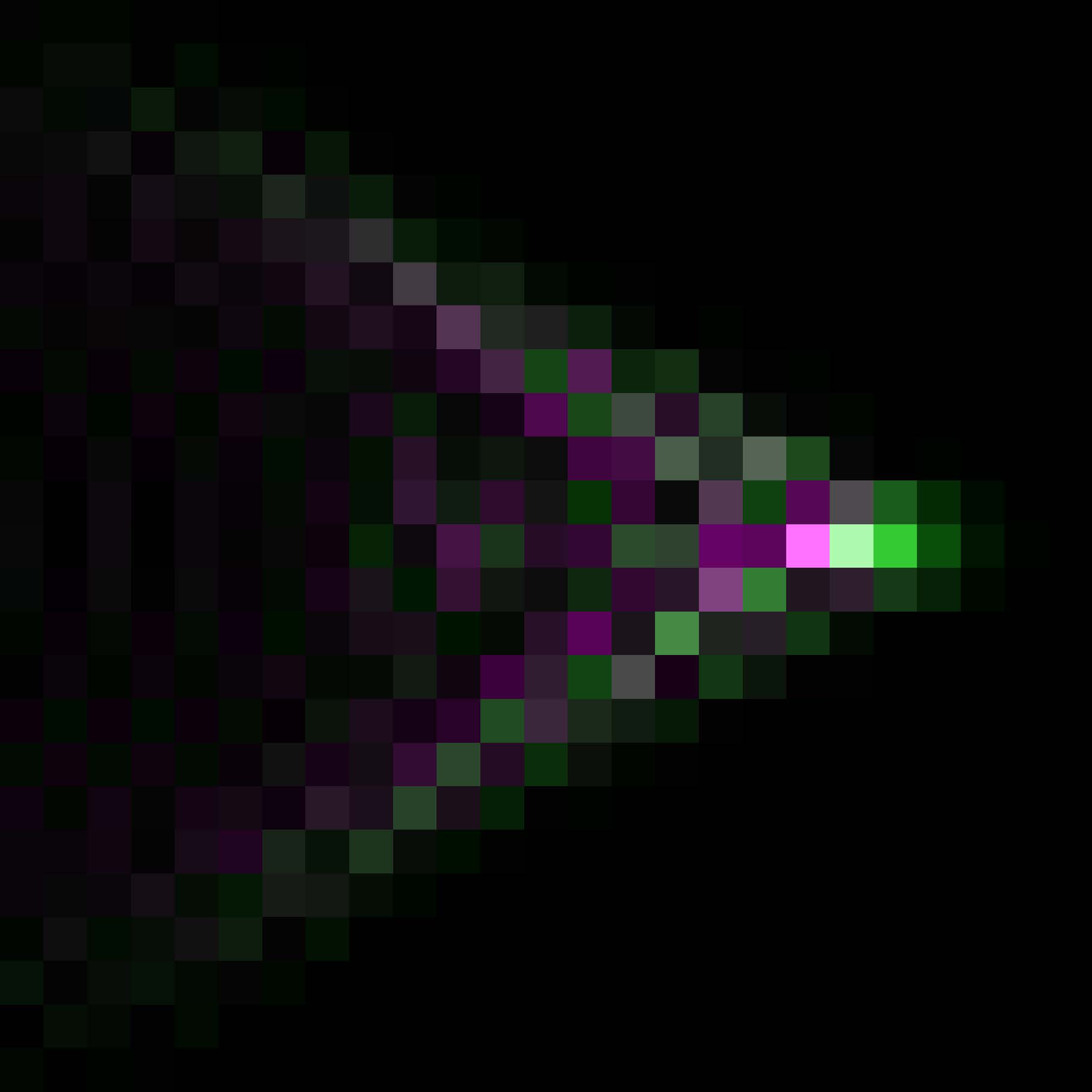}
    \caption{}
    \end{subfigure}
    \begin{subfigure}{\kernelSize\linewidth}
    \includegraphics[width=\linewidth]{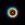}
    \caption{}
    \end{subfigure}
    \begin{subfigure}{\kernelSize\linewidth}
    \includegraphics[width=\linewidth]{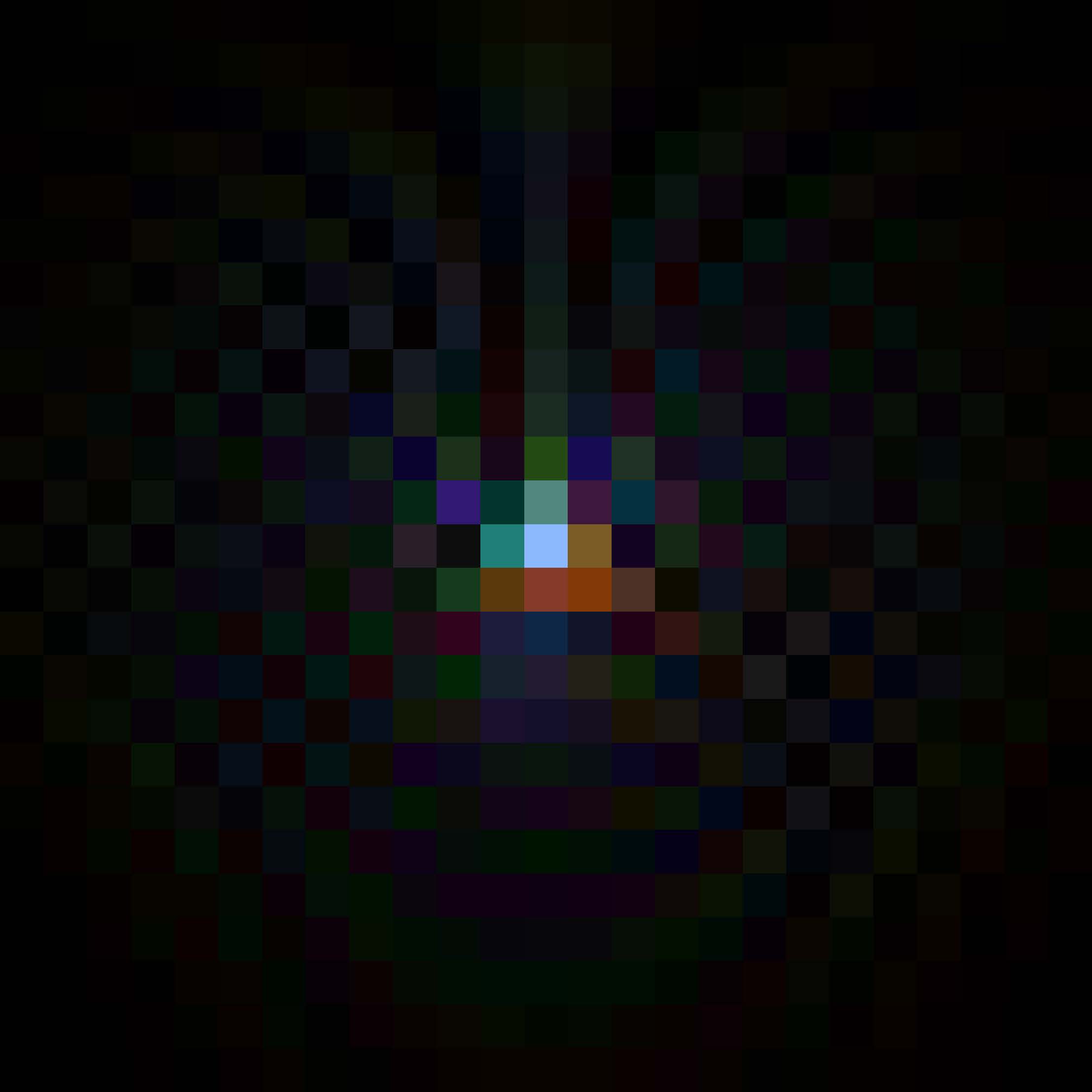}
    \caption{}
    \end{subfigure}
    \begin{subfigure}{\kernelSize\linewidth}
    \includegraphics[width=\linewidth]{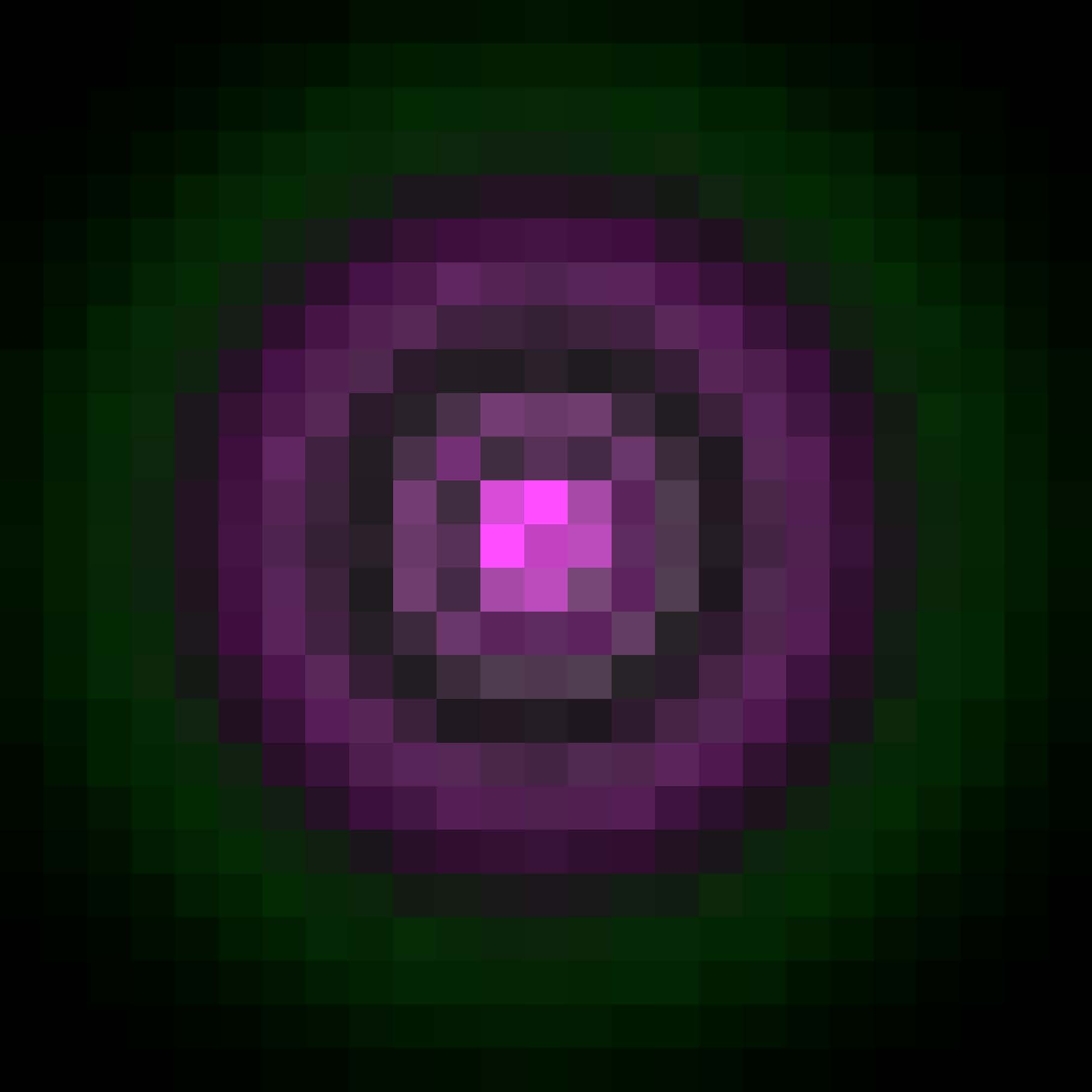}
    \caption{}
    \end{subfigure}
    \begin{subfigure}{\kernelSize\linewidth}
    \includegraphics[width=\linewidth]{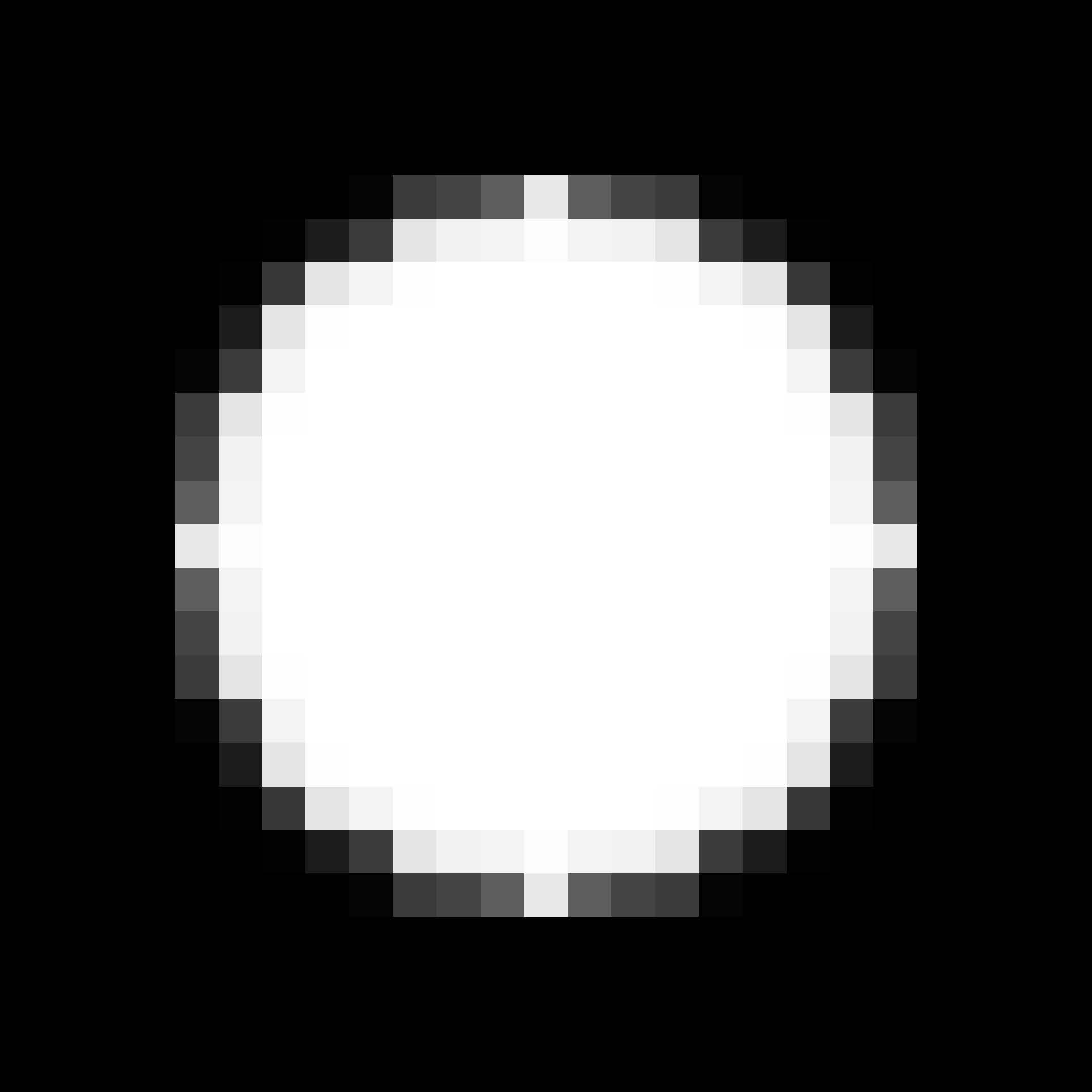}
    \caption{}\label{fig:disk_kernel}
    \end{subfigure}    
    \caption{Kernel samples available with OpticsBench (a-f) and a disk-like blur kernel (g). All kernels model optical aberrations, while the kernels in (a-f) additionally include chromatic aberration, asymmetric and non-constant shapes.}
    \label{fig:psf_samples}
\end{figure}
 
Although, for example, image classification networks produce excellent results on various challenges, it often remains unclear how well the trained models will generalize from the training distribution to practical test scenarios, that may present them with various domain shifts. For practical deployment, this generalization ability and robustness to data degradation is however crucial. 
Measuring robustness is addressed by various benchmarks introducing targeted corruptions~\cite{carlini_adversarial_2017,papernot_limitations_2016,croce_robustbench_2021}, common corruptions and adverse weather conditions~\cite{michaelis_benchmarking_2020,hendrycks_benchmarking_2019,kar_3d_2022} to vision datasets. 
These methods and benchmarks already cover a wide range of corruptions and yet use necessary simplifications. 

In this paper, we focus on a practically highly relevant case that has so far not been addressed in the literature: robustness to realistic optical aberration effects. Under high cost pressure, \eg in automotive mass production on cameras, compromises in optical quality may have to be made and natural production tolerances occur. In practice, even the constant use of a camera can lead to a change in image quality during the period of use, \eg due to thermal expansion. This can lead to increased aberrations~\cite{braun_automotive_2022}, \eg chromatic aberration and astigmatism, which are not considered in current robustness research. 

This paper proposes OpticsBench to close this gap, a benchmark that includes common types of optical aberrations such as coma, astigmatism and spherical aberration. The optical kernels are derived theoretically from an expansion of the wavefront into Zernike polynomials, which can be mixed by the user to create any \emph{realistic} optical kernel. The dataset grounds on 3D kernels (x, y, color) matched in size to the defocus blur corruption of~\cite{hendrycks_benchmarking_2019} as we consider the disk-shaped kernel as the base type of blur kernels.
We evaluate 70 different DNNs on a total of 1M images from the benchmark applied to ImageNet. Our evaluation shows that the performance of ImageNet models varies strongly for different optical degradations and that the disk-shaped blur kernels provide only weak proxies to estimate the models' robustness when confronted with optical degradations. 

Further, since our analysis indicates a lack of robustness to optical degradations, we propose an efficient tool to use optical kernels for data augmentation during DNN training, which we denote \emph{OpticsAugment}.  
In experiments on ImageNet-100, OpticsAugment achieves on average 18\% performance gain compared to conventionally trained DNNs on OpticsBench. 
We also show that OpticsAugment allows to improve on 2D common corruptions~\cite{hendrycks_benchmarking_2019} on average by 5.3\% points on ImageNet-100, \ie the learned robustness transfers to other domain shifts. 

\section{Related work}
\label{sec:related_work}
Vasiljevic et al.~\cite{vasiljevic2016examining} investigate the robustness of CNNs to defocus and camera shake. Hendrycks et al.~\cite{hendrycks_benchmarking_2019} provide a benchmark to 2D common corruptions. The benchmark includes several general modifications to images such as change in brightness or contrast as well as weather influences such as fog, frost and snow. They also include different types of blur, but only consider luminance kernels and more general types such as Gaussian or disk-shaped kernels.
Kar et al.~\cite{kar_3d_2022} build on this work and extend common corruptions to 3D. These include \emph{extrinsic} camera parameter changes such as field of view changes or translation and rotation. Michaelis et al.~\cite{michaelis_benchmarking_2020} and Dong et al.~\cite{dong_benchmarking_2023} provide robustness benchmarks for object detection on common vision datasets.
As the field of research grows, more subtle changes in image quality potentially posing a distribution shift are uncovered. This work aims to build on a more general treatment of blur types, which are known in optics research, but less common in computer vision.

A related field of research investigates robustness to adversarial examples created by targeted~\cite{croce_reliable_2020,moosavi-dezfooli_deepfool_2016} and untargeted~\cite{andriushchenko_square_2020} attacks. The goal is to introduce small perturbations of the input data in a way such that the model makes wrong classifications. Successful attacks pose a security risk to a particular DNN, while human observers would not even notice a difference and safely classify~\cite{carlini_adversarial_2017}.
Croce et al.~\cite{croce_robustbench_2021} provide a robustness benchmark, originally intended for adversarial robustness testing using AutoAttack~\cite{croce_reliable_2020}, while more practical $l_p$-bounds are discussed in~\cite{lorenz2022is}.
However, these methods are model specific in that the particular attack is \emph{optimized} for the model using \eg projected gradient descent in the backward pass. Therefore white-box methods require full knowledge about the underlying model. 
In contrast, model evaluation 
with OpticsBench corruptions can be done by applying simple filters to the validation data and thus requires only a clean image dataset. Therefore, it also works for black-box models. Convolving tensors with kernels is a base task in computer vision and so GPU-optimized implementations exist. These include parallel evaluation of OpticsBench for many models and on-the-fly training with OpticsAugment. 

Since these benchmarks reveal potential distribution shifts or lack of robustness for a given neural network model, concurrently methods are researched to improve robustness~\cite{gowal2021improving,geirhos2018imagenet,cubuk_autoaugment_2019} on various benchmarks. Hendrycks et al.~\cite{hendrycks_augmix_2020,hendrycks_many_2021} propose different data augmentation methods to improve classification robustness towards 2D common corruptions. The work of Saikia~\cite{saikia_improving_2021} further builds on this and achieves high accuracy on both clean and corrupted samples. Similarly, methods exist to improve adversarial robustness~\cite{salman2020adversarially} on benchmarks like~\cite{PINTOR2023109064,croce2021robustbench}. 

Modeling or retrieving Zernike polynomials~\cite{born_principles_1999,noll_zernike_1976,zernike_beugungstheorie_1934} is a common process in ophthalmological optics~\cite{thibos_retinal_2009,iskander_optimal_2001} to investigate aberrations of the human eye. The expansion is also widely used in other areas of optics, such as lens design~\cite{noauthor_opticstudio_nodate}, microscopy~\cite{cumming_direct_2020} and astronomy~\cite{ndiaye_calibration_2013,lane_wave-front_1992}. Therefore several tools~\cite{noauthor_opticstudio_nodate,dube_prysm_2019,kirshner2011} exist to generate PSFs from Zernike coefficients or wavefronts. The design is application specific and intended for optical engineers or physicists such as optical design software like Zemax Optic Studio~\cite{noauthor_opticstudio_nodate}. 
Our OpticsBench and OpticsAugment leverage such optical models to evaluate and improve current models with respect to~realistic optical effects.

\section{Blur kernel generation} 
\label{sec:kernel_generation}

\begin{figure}[h]
    \centering
    \includegraphics[width=\linewidth]{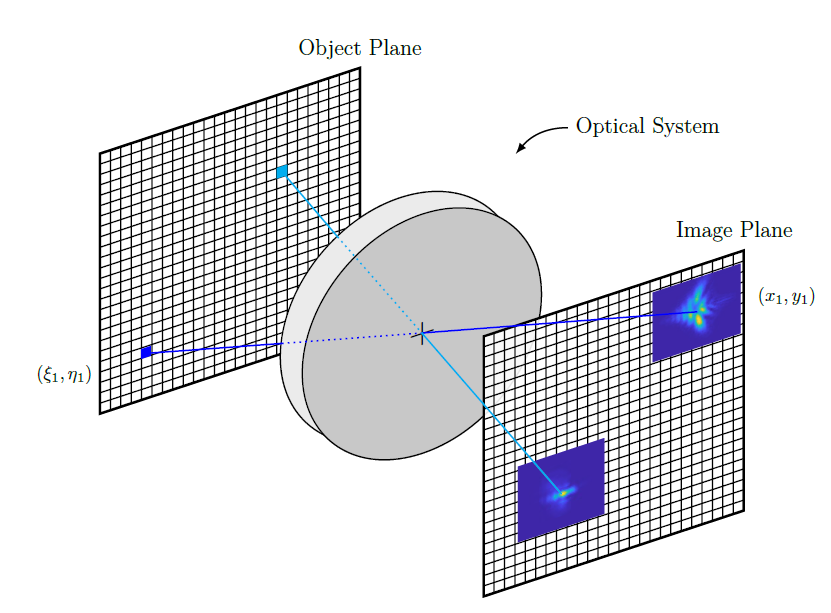}
    \caption{Imaging with space-variant PSF: Point sources in object space that pass the optical system appear spread and differently shaped in image space depending on the angle of incident. Kernel sizes are exaggerated for display purposes.}
    \label{fig:optical_psf}
\end{figure}

The point spread function (PSF) is the linear system response to a spherical point wave source emanating at some point in object space. With an ideal lens, this spherical wave would simply converge at the focus. Real-world systems create a shape specific to the aberrations present at that field position. Typically the wavelength-dependent PSF varies from the center to the edge and may also depend on azimuth as shown in Fig.~\ref{fig:optical_psf}. Point sources from different locations lead to different PSFs~\cite{born_principles_1999,goodman_introduction_2017}.
This directly impacts the imaged object: depending on the azimuth, the object appears differently blurred.
In this article, we follow a similar approach to show possible effects of lens blur on the robustness of DNNs. For small images of size $224\times224$ we assume a constant PSF and treat them as if they were regions of interest (ROI) from a larger image: Each image is blurred in different severities and optical aberrations. 
Since the images already contain different corruptions, including blur, the low pass filtering further reduces the image quality. In addition, depth-dependent blur would also control the blur size at a specific pixel. However, if the objects are beyond the hyperfocal distance, the depth-dependent blur variation can be neglected. This hyperfocal distance is for automotive lenses \eg at several meters. Farther objects have the same blur as objects at \emph{infinity}. Blur is more pronounced for these objects than for near objects because they are much smaller.

\subsection{Kernel generation}

To obtain kernel shapes specific to real optical aberrations, we use the linear system model for diffraction and aberration as in \cite{goodman_introduction_2017,born_principles_1999} and expand the wavefront in Zernike polynomials.
The effect of a non-ideal lens on a point source of wavelength $\lambda$ can be compactly summarized by a  Fourier transform $\mathcal{F}$ over a circular region that propagates the aberrated wave from pupil space to the image space. 
Fig.~\ref{fig:image_processing} visualizes the process. An ideal spherical wave passes through a circular pupil, which then deforms according to the  characteristics of the optical system. If there were no aberrations, the phase would be flat. 
\newcommand{\processSz}{0.15\linewidth}
\begin{figure}[h]
    \centering
    \begin{subfigure}{\processSz}
    \centering
        \includegraphics[trim=1cm 0.8cm 2cm 1.4cm,clip,width=\linewidth]{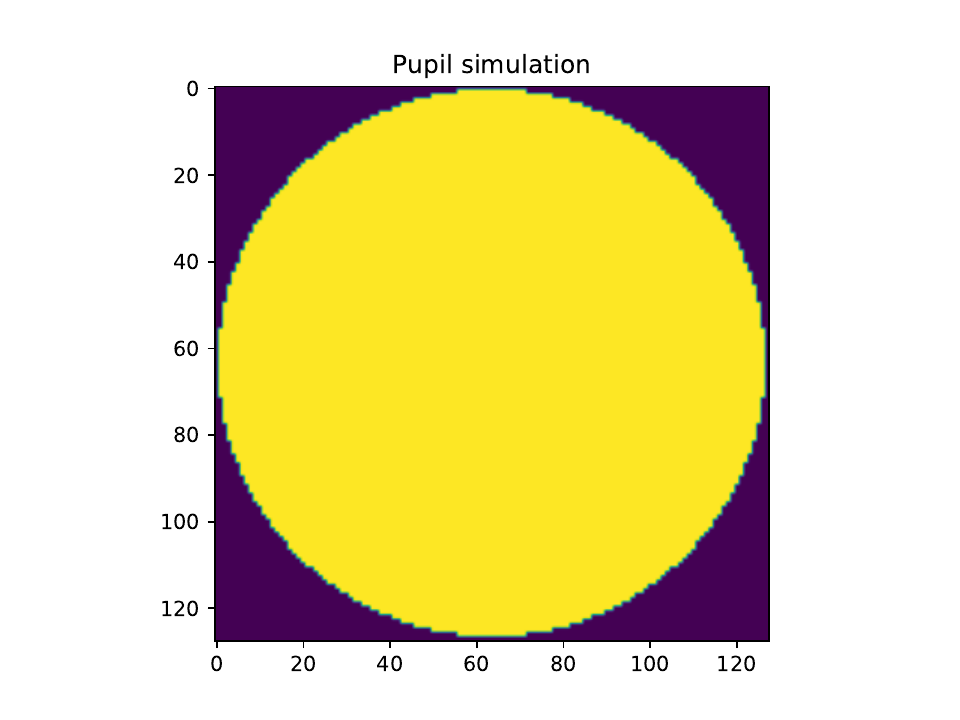}
        \caption{}
    \end{subfigure}
    \begin{subfigure}{\processSz}
    \centering
        \includegraphics[trim=1cm 0.8cm 2cm 1.4cm,clip,width=\linewidth]{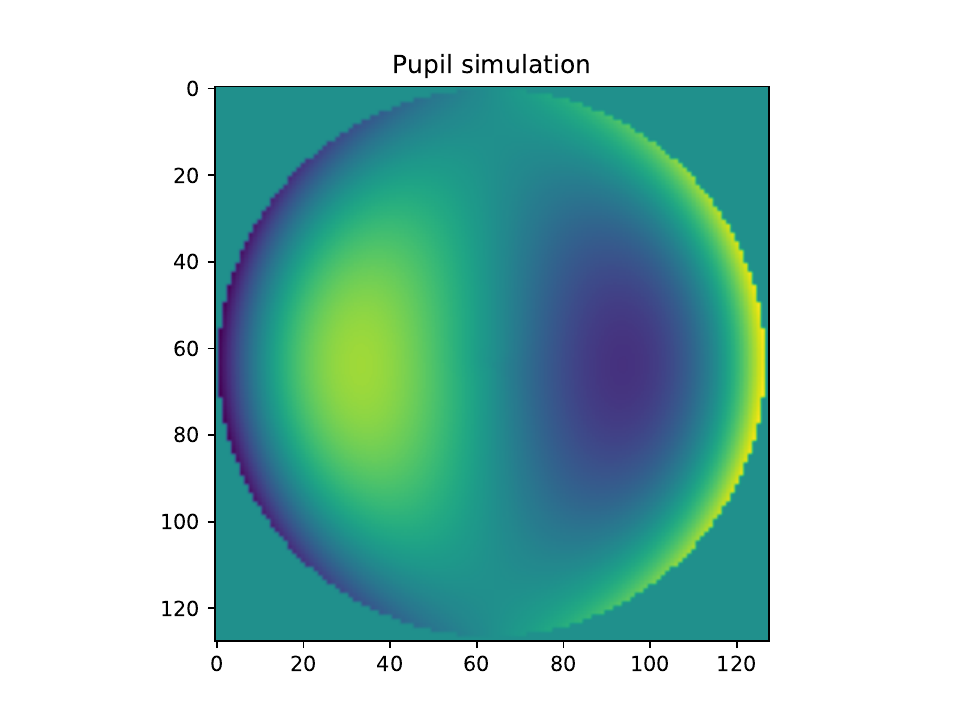}
        \caption{}
    \end{subfigure}
    \begin{subfigure}{\processSz}
    \centering
        \includegraphics[trim=1cm 0.8cm 2cm 1.4cm,clip,width=\linewidth]{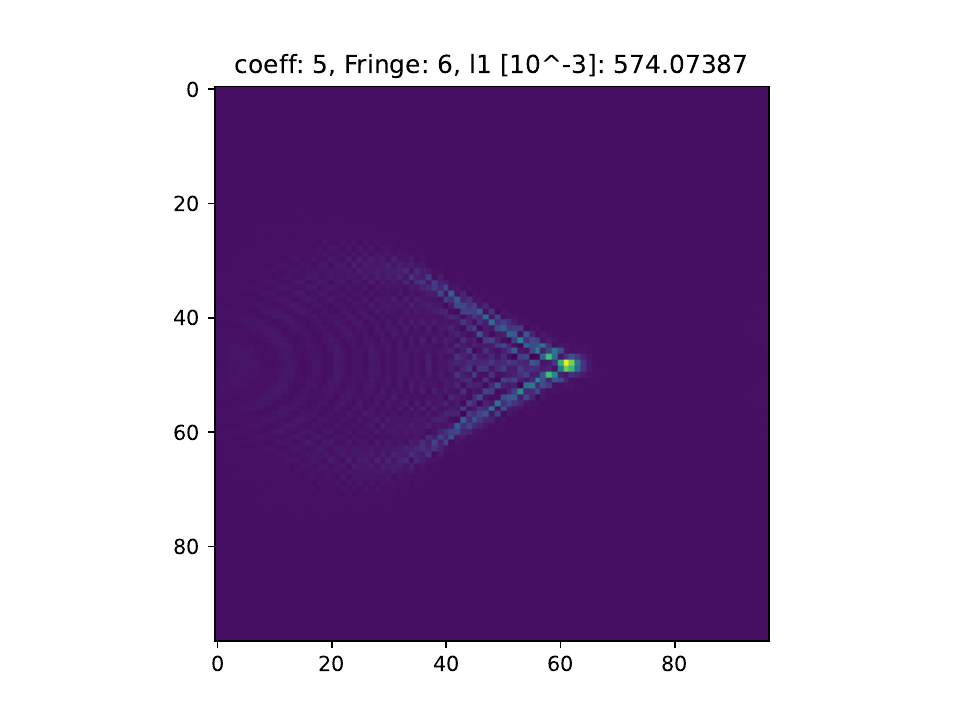}
        \caption{}
    \end{subfigure}\hspace{1cm}
    \begin{subfigure}{\processSz}
    \centering
        \includegraphics[width=\linewidth]{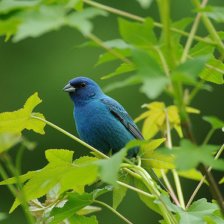}
        \caption{}
    \end{subfigure}
    \begin{subfigure}{\processSz}
    \centering
        \includegraphics[width=\linewidth]{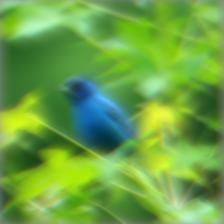}
        \caption{}
    \end{subfigure}
    \caption{Image processing scheme: The circular pupil (a)  contains an aberrated phase (b), which is mapped into image space by a 2D Fourier Transform, yielding a PSF (c). This PSF is convolved with an image (d) to produce a blurred image (e). The process is repeated for each color channel.}
    \label{fig:image_processing}
\end{figure}
The complex phase factor describing the optical path difference $W_\lambda$ is transformed to image space at $z_i$ to yield a PSF for wavelength $\lambda$: \cite{goodman_introduction_2017}
\begin{equation}
    h(u,v,\lambda) = \vert \mathcal{F}\bigl\{(Circ(x,y) \cdot e^{-j\frac{2\pi}{\lambda z_i} \mathbf{W}_\lambda(x,y,\lambda)}\bigr\}\vert^2
    . 
    \label{eq:propagation}
\end{equation}
A blurred image is then created with a 2D convolution of the PSF and the image. The model from Eq.~\eqref{eq:propagation} assumes scalar diffraction theory and therefore no polarization. Since we use an $l_1$-normed discrete PSF to preserve intensity in the final image, any scalar weights are suppressed. 
In this paper, we consider a very small imager $224\times224\times3$ that can be fed with ImageNet data. The model is restricted to represent the lens with a single PSF, no dependence on the object distance and angle, no magnification to concentrate on the kernel's shape nor its possible variation over the field of view. We observe the PSF $h$ in a region $u,v \in [0,25)$ pixels and apply it as a convolution kernel to the images. These restrictions allow for fast processing although an extension to space-variant  optical models is given in \cite{hirsch_efficient_2010,muller_simulating_2022,rerabek_space_2008,nagy_fast_1997}.

Now, to model specific PSFs, the expansion of the wavefront $W_\lambda$ into a complete and orthogonal set of polynomials $Z_n^m$ named after Fritz Zernike \cite{born_principles_1999,zernike_beugungstheorie_1934,lakshminarayanan_zernike_2011} is used:
\begin{equation}
W_{\lambda}(x,y,\lambda) = \lambda \cdot \sum_{n,m} A_n^m(\lambda) \cdot Z_n^m
\label{eq:zernike_expansion}
\end{equation}
Each coefficient $A_n^m$ in multiples of the  wavelengths $\lambda_i$ represents the contribution of a particular type of aberration and therefore different aspects such as the amount of coma, astigmatism or defocus can be turned off or on. In this article, we choose the eight isolated Zernike Fringe Polynomials, including primary and more complex aberrations. Tilt x and tilt y are not considered here. The concrete list is marked as bold text in Tab.~\ref{tab:zernike_fringe_modes}.
\begin{table}[]
    \caption{First twelve Zernike Fringe modes. We select numbers $4$-$11$ for OpticsBench. However, an extension to other (higher) modes is easily accomplished.}
    \centering
    \small
    \begin{tabular}{ll|ll}piston
        \# & name & \# & name \\
         1 & piston   &   \textbf{7} & \textbf{horizontal coma} \\
         2 & tilt x   &   \textbf{8} & \textbf{vertical coma} \\
         3 & tilt y   &   \textbf{9} & \textbf{spherical} \\
         \textbf{4} & \textbf{defocus}  &   \textbf{10} & \textbf{oblique trefoil} \\
         \textbf{5} & \textbf{oblique astigmatism} & \textbf{11} & \textbf{vertical trefoil} \\
         \textbf{6} & \textbf{vertical astigmatism} & 12 & sec. vert. astigmatism
    \end{tabular}
    \normalfont
    \label{tab:zernike_fringe_modes}
\end{table}
Conversely to Seidel aberrations not occurring isolated but mixed in practice, we select isolated Zernike modes, combining Seidel aberrations by definition, such as astigmatism \& defocus. For convenience, we refer to these aberrations here briefly as their Seidel aberration equivalent. However, lens aberrations of real systems usually consist of a bunch of different Zernike modes, see e.g. the Zemax sample objectives, we select here single Zernike modes to allow for categorization and encourage researchers to combine different modes and investigate its impact on model robustness. Although we do not show distortion, tilt, and field curvature here, the same framework can be used to model or process entire lens representations. This is particularly useful for image datasets with larger images, as the lens corruptions then may vary significantly over the field of view, \eg the Berkeley Deep Drive (BDD100k).~\cite{yu_bdd100k_2020}
The PSF generation is done in Python and the pupil modeling is based on~\cite{dube_prysm_2019}.

\subsection{Kernel matching}

To compare the impact of different kernel types on each other, it is crucial to have size-matched representations. As a baseline, we use the simple disk-shaped kernel prototype from Hendrycks et al.~\cite{hendrycks_benchmarking_2019} shown in Fig.~\ref{fig:disk_kernel}. Then, with an educated initial guess of coefficient values two kernels are evaluated on different metrics and optimized by offsetting the coefficients in steps of $\pm0.1\lambda$. From this, we take the best overall fit as the kernel pair.

First, to compare two kernels, the Modulation Transfer Function (MTF) is obtained and evaluated in differently orientated slices (0°, 45°, 90°, 135°)~\cite{boreman_modulation_2001,goodman_introduction_2017}. From this, the frequency value at $50\%$ (MTF50) and the area under the curve (AUC) are obtained. The difference between these metrics provides a distance in  optical quality. Further, SSIM and PSNR are reported~\cite{wang_image_2004}. These metrics aim to compare the shape. 
Secondly, the kernels are convolved and then analyzed on established test images used in benchmarking camera image quality as another matching criterion.~\cite{phillips_camera_2018,noauthor_ieee_2017,noauthor_iso122332017_2017} 
We evaluate SSIM and PSNR on a slanted edge test chart and the scale-invariant spilled coins test chart, both at the required ImageNet target resolution of $224\times224$. The spilled coins chart, shown in Fig.~\ref{fig:spilled_coins}, consists of randomly generated disks of different sizes and is designed to measure texture loss as an image-level MTF.~\cite{mcelvain_texture-based_2010,burns_refined_2013} We report here the acutance and MTF50 value as evaluated with the algorithm of Burns~\cite{burns_refined_2013} from the spilled coins chart and compare once again between the two kernel representations. The evaluated image and optical quality on the spilled coins test chart serves here as proxy for other images from the image datasets.

Although the obtained optical kernels match also for higher severities with the defocus blur corruption~\cite{hendrycks_benchmarking_2019}, blur kernel sizes observed in the real world would be smaller, so this poses rather an upper bound on the  observable blur corruption for an entire image. However, small objects such as pedestrians in object detection datasets such as \cite{yu_bdd100k_2020} may suffer from such severe blurring as the object size decreases to tens of pixels.

\newsavebox{\imagebox}
\savebox{\imagebox}{\includegraphics[page=18,trim=0cm 8cm 21.6cm 3.5cm,clip,width=0.355\linewidth]{figures/matching/report_sheets_cooke_rgb_iccv.pdf}}%

\begin{figure}
\begin{subfigure}[t]{0.2\linewidth}   
\centering
\raisebox{\dimexpr.5\ht\imagebox-.5\height}{ \includegraphics[width=\linewidth]{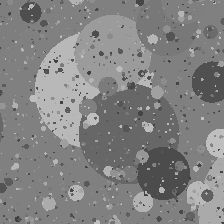}}
\caption{}
\label{fig:spilled_coins}
\end{subfigure}\hspace{.2cm}
\begin{subfigure}[t]{0.37\linewidth}
\centering\usebox{\imagebox}
\caption{}
\end{subfigure}
\begin{subfigure}[t]{0.38\linewidth}
\centering
\includegraphics[page=18,trim=9cm 8cm 12cm 3.5cm,clip,width=\linewidth]{figures/matching/report_sheets_cooke_rgb_iccv.pdf}    
\caption{}
\end{subfigure}
    \centering
    \caption{(a) Spilled coins test chart generated with Imatest target generator tool~\cite{noauthor_imatest_nodate}. (b) MTF curves obtained from degraded versions of (a) and the corresponding PSF based MTF (c). The red curves refer to the baseline kernel at severity 3 and black to the corresponding astigmatism.}
    \label{fig:my_label}
\end{figure}

\section{OpticsBench - benchmarking robustness to selected optical aberrations}
\label{sec:benchmark}
Similar to Hendrycks et al.~\cite{hendrycks_benchmarking_2019} we define five different severities for each optical aberration. To obtain comparable corruptions, the defocus blur corruption from \cite{hendrycks_benchmarking_2019} serves as baseline for OpticsBench. 
For each severity, our kernels are matched to the defocus blur kernel types using the method and metrics described above. We generate two sets of kernels with different chromatic aberrations: OpticsBenchRG consists of the same kernels as OpticsBench, but with only the red and green channels spread as in Fig.~\ref{fig:psf_astigmatism_rg} and thus producing severe chromatic aberration. For brevity, we refer to our kernels as \emph{optical kernels}, although the disk-shaped kernel type  from~\cite{hendrycks_benchmarking_2019} is a model for defocus obtained from geometric optics.

Each set of kernels consists of 40 kernels for the eight different Zernike modes. 
We combine Zernike modes that result in similar shapes into a single corruption as in Tab.~\ref{tab:corruptions}.
\begin{table}[]
    \caption{Corruption types used in OpticsBench with their corresponding Zernike mode equivalent from Tab.~\ref{tab:zernike_fringe_modes}. 
    }
    \centering
    \begin{tabular}{llll}
       Astigmatism  & $(5,6)$ & Trefoil & $(10,11)$\\
       Coma & $(7,8)$ & Defocus \& spherical & $(4,9)$\\
    \end{tabular}
    \label{tab:corruptions}
\end{table}
This results in four different optical corruptions. Each corrupted dataset is then obtained by randomly assigning one of the two kernel types. A predefined seed of the pseudo-random number generator ensures reproducibility. 
Before applying the blur kernels, the images are resized to $256\times256$ and center cropped to $224\times224$, to avoid any reduction of the effect.

A set of DNNs is inferred for each severity and corruption, and the classification accuracy (acc@1) is reported. Additionally, the baseline and the clean dataset are evaluated. 

To compare accuracies with the baseline, we define the deviation of a specific corruption $c_i$ from the baseline $b$ (defocus blur):
\begin{equation}
    \Delta_{c_i, b} = Acc_{c_i} - Acc_{b}
\end{equation}
Additionally, the Kendall Tau rank coefficient~\cite{kendall_treatment_1945} is evaluated on the DNN model ranking to further investigate whether the optical kernels elicit different behavior compared to the disk-shaped kernels for different architectures.

The Python code to re-create both the benchmark inference scripts and the benchmark datasets for all optical corruptions is provided\footnote{\url{https://github.com/PatMue/classification_robustness}}. This includes the sets consisting of 40 kernels for OpticsBench and OpticsBenchRG. The script is intended for ImageNet-1k and ImageNet-100, but can be extended to other datasets. Since the benchmark is intended to be modifiable to specific user investigations (e.g. different aberrations), the source code to create kernels is also available but not as part of this contribution.
Generating the OpticsBench datasets (five severities, four corruptions) from the 50k ImageNet-1k validation images takes about 120 minutes with six PyTorch workers and batch size 128 on an 8-core i7-CPU and 32GB RAM equipped with NVIDIA GeForce 3080-Ti 12GB GPU. Using smaller kernels than ours ($25\times25\times3$) may speed up the process.

\section{Experiments on ImageNet} 
\label{sec:experiments_imagenet}
OpticsBench is built on the ImageNet validation dataset,
consisting of 50k images distributed in 1000 classes. All four corruptions (astigmatism, coma, defocus blur \& spherical and trefoil) are divided into 5 different severities each. We evaluate the 65 DNNs from the torchvision model zoo including a wide range of architectures as listed in  Tab.~\ref{tab:architectures}. All models are pre-trained on ImageNet-1k. MobileNet\_v3 is trained using  AutoAugment~\cite{cubuk_autoaugment_2019}, EffcientNet training uses CutMix~\cite{yun_cutmix_2019} and MixUp~\cite{zhang_mixup_2018} and ConvNeXt and VisionTransformers use a combination of these.
In addition to these networks, we evaluate ResNet50 models from the RobustBench~\cite{croce2021robustbench} leaderboard, which are reportedly robust against common corruptions~\cite{hendrycks_augmix_2020,hendrycks_many_2021,erichson_noisymix_2022}.  
Accuracies on the validation set are available in the supplementary material.
\begin{table}[h]
\caption{Selected architectures used from PyTorch vision model zoo in small and large variants.}
    \centering
    \small
    \begin{tabular}{l l l}
    ConvNeXt~\cite{liu2022convnet}& Inception\_v3~\cite{szegedy_going_2015}& ResNet~\cite{he_deep_2016}\\
    DenseNet~\cite{huang_densely_2017} & MobileNet~\cite{howard_mobilenets_2017} &  ResNeXt~\cite{xie_aggregated_2017} \\
    EfficientNet~\cite{tan_efficientnet_2019} & RegNet~\cite{radosavovic_designing_2020} &  ViTransformer~\cite{ranftl_vision_2021}
    \end{tabular}
    \normalfont
\label{tab:architectures}    
    \centering
\end{table}

\begin{figure}[h]
    \centering
    \includegraphics[width=\linewidth]{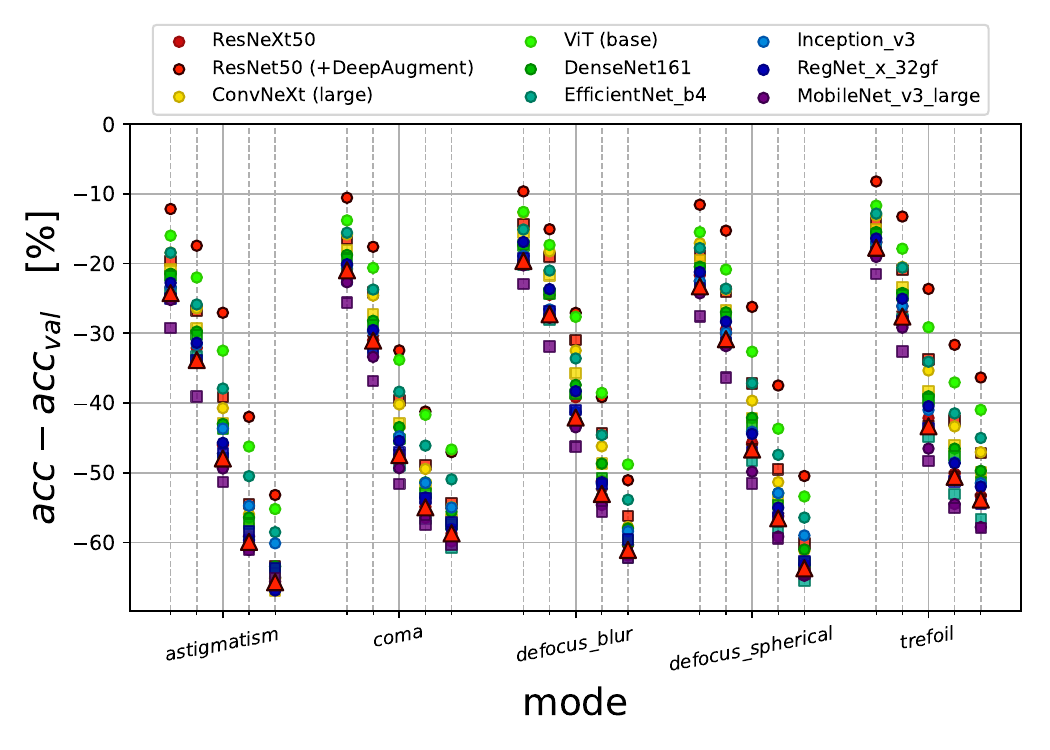}
    \caption{Difference in Accuracy on OpticsBench compared to the clean data. For each corruption five levels of severity are shown from left to right. We also include defocus blur~\cite{hendrycks_benchmarking_2019}, used as baseline to match our kernel sizes. Each color represents a DNN from Table~\ref{tab:architectures}.
    }
    \label{fig:acc_vs_mode_sub_val}
\end{figure}

Fig.~\ref{fig:acc_vs_mode_sub_val} shows the difference in accuracy compared to the validation data for all corruptions.
The severities are plotted from left to right for a single corruption. The different colors refer to representatives from Table~\ref{tab:architectures}. The evaluation of all 70 models is found in the supplementary material~\ref{app:ranking}.
In Fig.~\ref{fig:acc_vs_mode_sub_val} the ResNet50 architecture, marked with red triangles for the standard training and with red circles for ResNet50+DeepAugment, clearly benefits from the training with DeepAugment. It drops by almost 65\% points for severity 5 and astigmatism compared to the clean data accuracy. The robust model with DeepAugment looses 52\% points. In general, the performance at a particular severity and corruption depends on the DNN.

\begin{figure}[h]
    \centering
    \includegraphics[width=\linewidth]{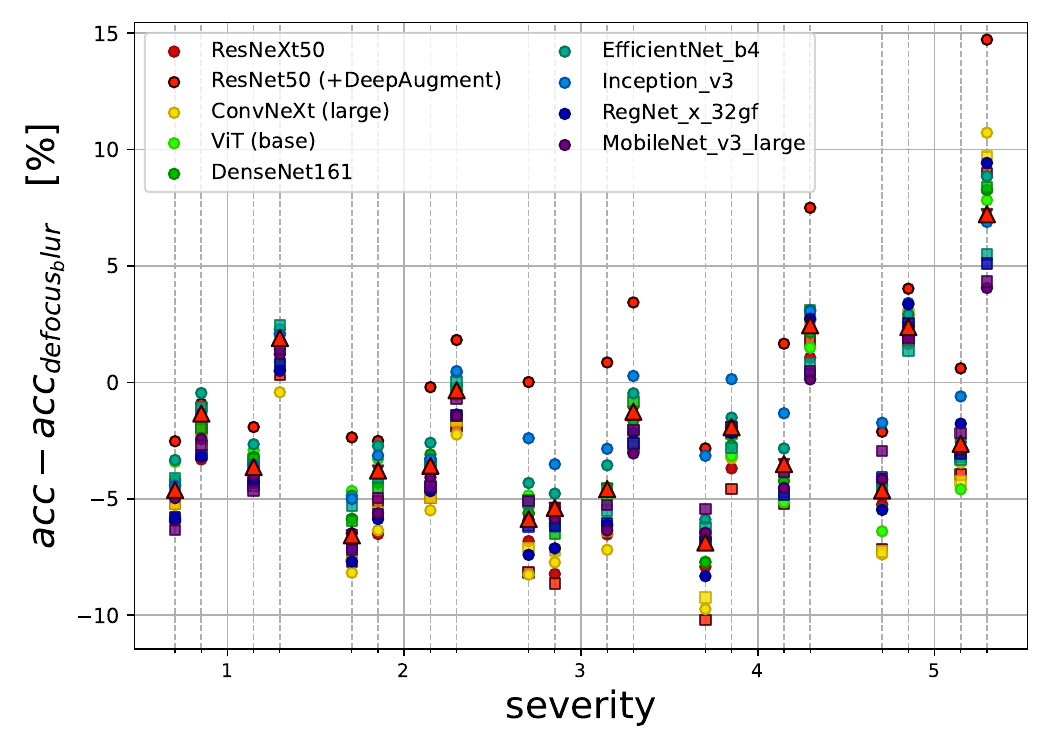}
    \caption{Comparison of Accuracy for the baseline and the different corruptions and severities. The corruptions per severity are from left to right: astigmatism, coma, defocus \& spherical and trefoil. 
    }
    \label{fig:acc_vs_severity_sub_defocus_blur}
\end{figure}

Fig.~\ref{fig:acc_vs_severity_sub_defocus_blur} compares the achieved accuracy at a specific severity and mode with the disk-shaped kernel equivalent from~\cite{hendrycks_benchmarking_2019}. 
For each severity the four OpticsBench corruptions are shown from left to right (astigmatism, coma, defocus \& spherical, trefoil). The red circles show for the robust ResNet50 model that the particular relative performance depends not only on the severity but also on the corruption: the various kernel types challenge each DNN differently. Especially the robust ResNet50 (red circle) and Inception (light blue circle) differ from the mean deviation. This is further justified by the ranking comparisons in the supplementary material~\ref{app:ranking}. 
The different rankings show that the blur types are processed differently by the DNNs and that each blur type needs to be considered.

\section{OpticsAugment - augmenting with optical kernels}
\label{sec:augmentation}
Since the above results show a clear drop in performance due to the different corruptions, and motivated by the general benefit of training with data augmentation, we propose here a data augmentation method with kernels from the OpticsBench kernel generation, which supplements existing methods. 

OpticsAugment follows a similar approach as AugMix~\cite{hendrycks_augmix_2020} and is described in Alg.~\ref{alg:optics_augment}. During dataloading in the training process each image is convolved with an individual RGB-kernel from the kernel stack containing e.g. 40 kernels for the different corruptions and severities. 
The random selection of the kernel follows a uniform distribution. Additionally, each resulting image is a weighted combination of the original image and the blurred image, following a beta-distribution as in~\cite{hendrycks_augmix_2020}, which controls the amount of augmentation. Thus, the total number of images per epoch stays the same. The exact combination varies each epoch, which facilitates generalization to the blur types. Since the processing is accelerated by GPU parallelization, the blurred images do not need to be computed in advance, but can be computed on-the-fly with comparable overhead to other methods such as~\cite{kar_3d_2022} and a kernel size of 25x25x3. For this, Alg.~\ref{alg:optics_augment} is implemented using parallelization on full image batches, which results in strong acceleration. For an image batch of 128 images an overhead of about 250ms is generated. Using dataloaders takes much longer since the number of workers is typically about 4-8, while the batch size is typically between 32 and 128. The implementation relies heavily on GPU acceleration using cuda and pytorch. The processed image batch is then normalized to the dataset specific mean and standard deviation. This is crucial, otherwise the training loss won't converge. 
\begin{algorithm}[h]
    \SetKwInOut{Input}{Input}
    \SetKwInOut{Output}{Output}
    \underline{augment} $(\boldsymbol{x},kernels,severity=3,\alpha=1.0)$ :
    \Input{Clean tensor $\boldsymbol{x}$ consisting of $N$ images, kernel stack $kernels$ and severity between 1-5, intensity $\alpha$}
    \Output{Randomly blurred image batch $\boldsymbol{x}$}
    \caption{OpticsAugment}
    \label{alg:optics_augment}
    \For{$n=0,1,...,N$}{
        $blurred=zeros\_like(x)$
        $h\gets choice(kernels)$ randomly select a kernel from $kernels$ from one of the available augmentation types (e.g. oblique astigmatism). \\
        \For{color=0,1,..,3}{
            $blurred[n,color]\gets \boldsymbol{x}[n,color] * h[color]$ 2D convolution for image $x$ and $color$\\
        }
        $p \gets realization(\alpha)$ sample from a $\beta$-distribution controlling amount of augmentation\\
        $x\gets (1-p) \cdot x + p \cdot blurred$ \\
    }
    $\boldsymbol{x} \gets normalize(\boldsymbol{x})$ adapt image batch to dataset specific mean and standard deviation\\
    return $\boldsymbol{x}$
\end{algorithm}

Since AugMix~\cite{hendrycks_augmix_2020} provides low-cost data augmentation and promising results but no diverse blur kernel augmentation, we also try pipelining AugMix and OpticsAugment. Since both methods can feed the training algorithm with highly corrupted images, direct chaining leads to non-convergence and low accuracy. 
Therefore, we model the probability of augmentation with a flat Dirichlet distribution in four dimensions: The first two variables are the probabilities for AugMix and OpticsAugment. The other variables are auxiliary variables and ensure that the overall probability of each augmentation remains uniformly distributed. The output of OpticsAugment is normalized.

\section{Experiments on ImageNet-100} 
\label{sec:experiments_imagenet100}
On ImageNet-100, which is a smaller dataset with only 100 classes from ImageNet-1k, five different DNNs are trained with OpticsAugment. In addition, we train baseline models on ImageNet-100 with the same hyperparameter settings, but without the data augmentation. These DNNs are then compared to each other on OpticsBench and 2D Common Corruptions~\cite{hendrycks_benchmarking_2019} applied to ImageNet-100. We select five different architectures:  EfficientNet\_b0, MobileNet\_v3\_large, DenseNet161, ResNeXt50 and ResNet101.

The train split is divided into $5\%$ validation images and $95\%$ train images to ensure that the benchmark data only contains unseen data. On top of the trained DNNs all models are also trained with same settings but include OpticsAugment with severity 3 during training and the amount of augmentations is uniformly distributed, so $\alpha=1.0$.
The hyperparameter settings follow the standard training recipes as reported in the pytorch references~\cite{noauthor_visionreferencesclassification_nodate} using cross-entropy loss, stochastic gradient descent and learning rate scheduling but no data augmentation. However, no additional data augmentation such as CutMix~\cite{yun_cutmix_2019} or AutoAugment~\cite{cubuk_autoaugment_2019} are used. So, the achieved accuracy for MobileNet would be improved with AutoAugment. The DNNs are trained using the same batch size and number of epochs for clean and OpticsAugment training. Fine-tuning the hyperparameters can further improve the observed benefit. 

\begin{table}[]
    \caption{Performance \emph{gain} with OpticsAugment on all ImageNet-100 OpticsBench corruptions. Average difference in accuracy across all corruptions in \%-points for each severity.  Details are given in supplementary~\ref{app:imagenet100_and_c}, tables~\ref{tab:tab:imagenet100_corruptions_DenseNet_revisited}-\ref{tab:tab:imagenet100_corruptions_ResNeXt50_revisited}.}
    \centering
    \begin{tabular}{l l l l l l}
        DNN & 1 & 2 & 3 & 4 & 5\\
        \hline 
        DenseNet161  & 14.77 &  21.96 &  27.26  & 20.98 & 13.84 \\
        ResNeXt50 & 17.40 &  24.49 &  29.56 & 22.32 &  14.76\\
        ResNet101 & 9.97 & 16.24 & 21.15 & 17.39 & 12.07\\
        MobileNet & 8.12 & 12.73 & 13.80 & 10.09 & 7.27\\
        EfficientNet  & 8.45 & 12.60 & 12.90 & 9.43 & 7.35\\
    \end{tabular}
    \label{tab:optics_augment_opticsbench}
\end{table}

\begin{figure}[h]
    \centering
    \begin{subfigure}{\linewidth}
    \includegraphics[width=\linewidth]{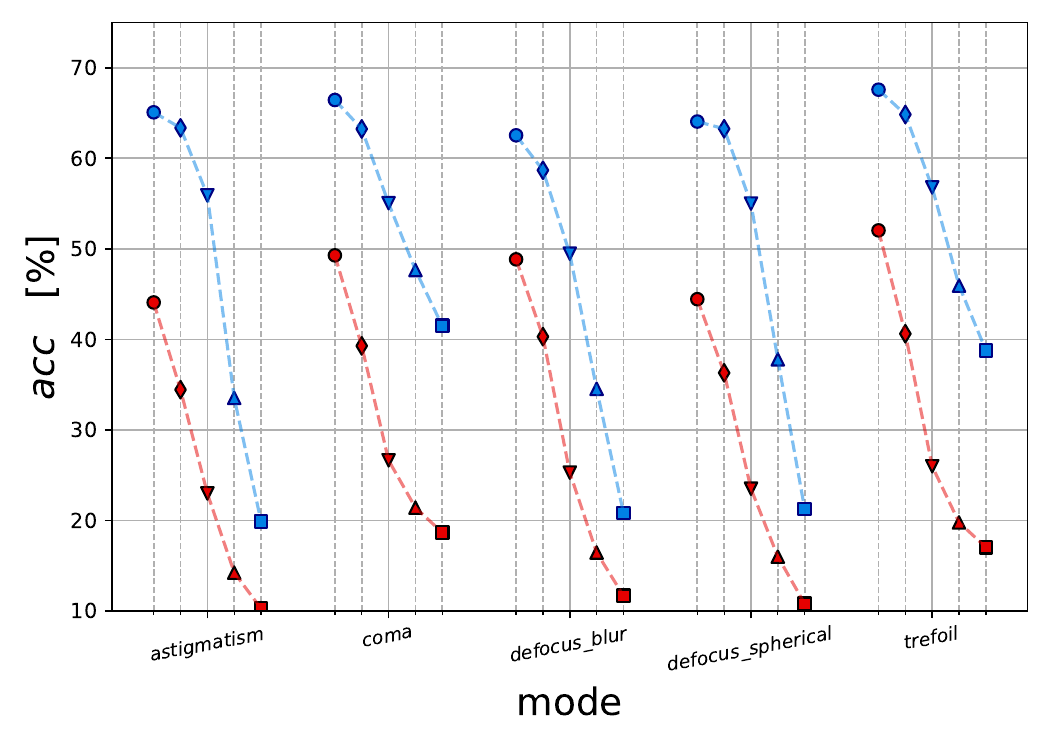}
    \caption{ResNext50}
    \label{fig:imagenet100_resnext50}
    \end{subfigure}
    \begin{subfigure}{\linewidth}
    \includegraphics[width=\linewidth]{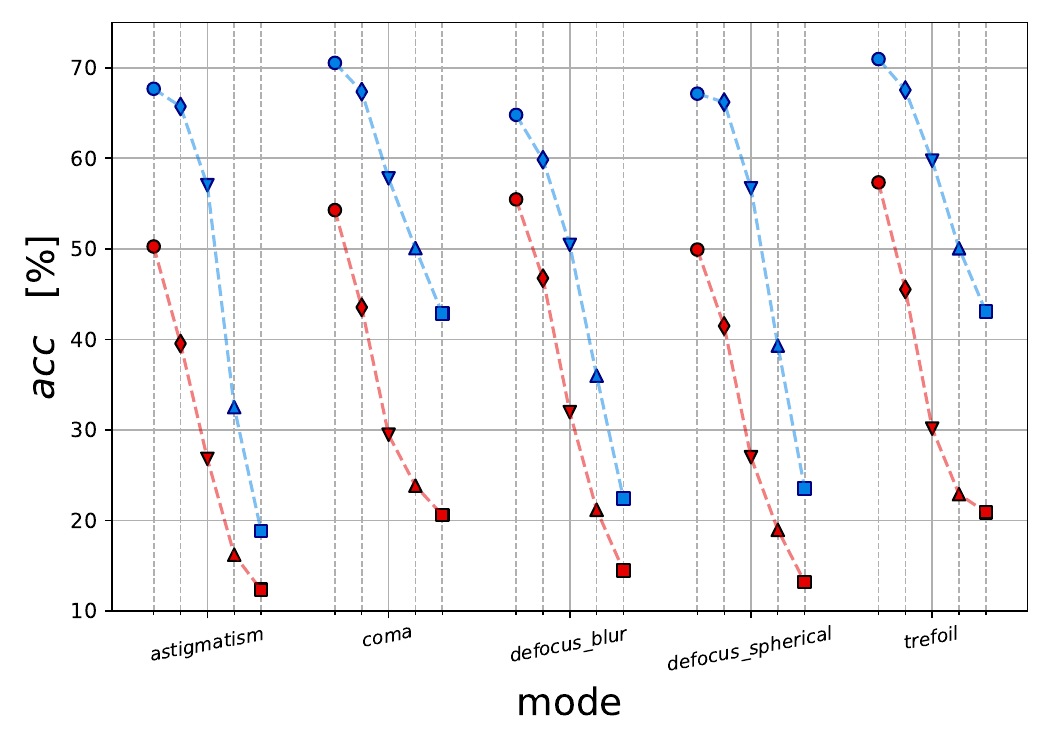}
    \caption{DenseNet161}
    \label{fig:imagenet100_densenet161}
    \end{subfigure}
    \caption{Accuracy on OpticsBench-ImageNet-100 for  DNNs with (blue) and without (red) OpticsAugment training. The x-axis shows for each of the five corruptions (Astigmatism, Coma, Defocus~\cite{hendrycks_benchmarking_2019}, Defocus \& Spherical, Trefoil) five different severities (from left to right).
    (a) ResNeXt50 (b) DenseNet161.  The accuracy increases by on average $29.6$\% points for ResNeXt50 compared to the clean trained DNN (red) \textbf{with OpticsAugment (blue)} and severity $3$ (down-pointing triangles). The performance gain for defocus blur~\cite{hendrycks_benchmarking_2019} is the lowest for all severities.}
    \label{fig:optics_augment}
\end{figure}

Tab.~\ref{tab:optics_augment_opticsbench} gives an overview of the improvement on ImageNet-100 OpticsBench with OpticsAugment. The smaller DNNs (MobileNet, EfficientNet\_b0) show significantly lower improvements in accuracy, while ResNeXt50 gains up to $29.6\%$ points with OpticsAugment.
Fig.~\ref{fig:optics_augment} compares the accuracies with/without OpticsAugment during training for each corruption. The ResNeXt50 in Fig.~\ref{fig:imagenet100_resnext50} improves with OpticsAugment (blue) on average by 21.7\% points compared to the default model (red). The robustness to coma is significantly improved, especially for higher severities.
The results for DenseNet161 in Fig.~\ref{fig:imagenet100_densenet161} are similar with an average improvement of 19.8\% points. The severity 3 accuracies in Fig.~\ref{fig:imagenet100_densenet161} for the default DenseNet161 model are comparable, while OpticsAugment is significantly more robust to OpticsBench corruptions. For instance, trefoil (last data point) is handled overly well while the performance gain for defocus blur is significantly lower. This seems to hold across different severities and DNNs: The corresponding disk-shaped kernel~\cite{hendrycks_benchmarking_2019} was not present during training, suggesting that diverse blur kernels need to be considered during training. Together with tables supporting this claim the results for the other DNNs can be found in the supplementary material~\ref{app:imagenet100_and_c}.

Additionally, the tuples of DNNs  (baseline, baseline+OpticsAugment) are evaluated on 2D common corruptions~\cite{hendrycks_benchmarking_2019}. The benchmark consists of 19 different corruptions including various blur types, noise and weather conditions. The average improvement with OpticsAugment is shown in Tab.~\ref{tab:optics_augment_imagenet100_c}. On average all DNNs still improve with OpticsAugment under the influence of the 2D common corruptions. 
\begin{table}[b]
    \caption{Performance \emph{gain} with OpticsAugment on all 2D common corruptions \cite{hendrycks_benchmarking_2019} as average difference in accuracy across all corruptions in \%-points for each severity. Details are given in supplementary~\ref{app:imagenet100_and_c}, tables~\ref{tab:tab:imagenet100c_corruptions_DenseNet}-\ref{tab:tab:imagenet100c_corruptions_ResNeXt50}.}
    \centering
    \begin{tabular}{l l l l l l}
        DNN & 1 & 2 & 3 & 4 & 5\\
        \hline 
        DenseNet161  & 5.08 &  7.55 &  8.73  & 7.30 & 5.38 \\
        ResNeXt50 & 5.11 &  7.63 &  8.68 & 7.18 &  5.27\\
        ResNet101 & 1.25 & 3.07 & 4.55 & 4.90 & 4.10\\
        MobileNet & 3.58 & 4.92 & 4.78 & 3.69 & 3.07\\
        EfficientNet  & 4.35 & 6.32 & 6.70 & 4.62 & 3.69\\
    \end{tabular}
\label{tab:optics_augment_imagenet100_c}
\end{table}
As an example, we discuss here the results for ResNeXt50 with and without OpticsAugment in Fig.~\ref{fig:common2d_modes_resnext50}. (Results for more models are in the supplementary material~\ref{app:imagenet100_and_c}). 
Fig.~\ref{fig:common2d_modes_resnext50} compares the accuracies for ResNeXt50 solely trained on ImageNet-100 (red) and augmented with OpticsAugment (blue) respectively. 
\begin{figure*}[ht]
    \centering 
    \includegraphics[trim=0cm 1cm 0cm 0cm,clip,width=0.7\linewidth]{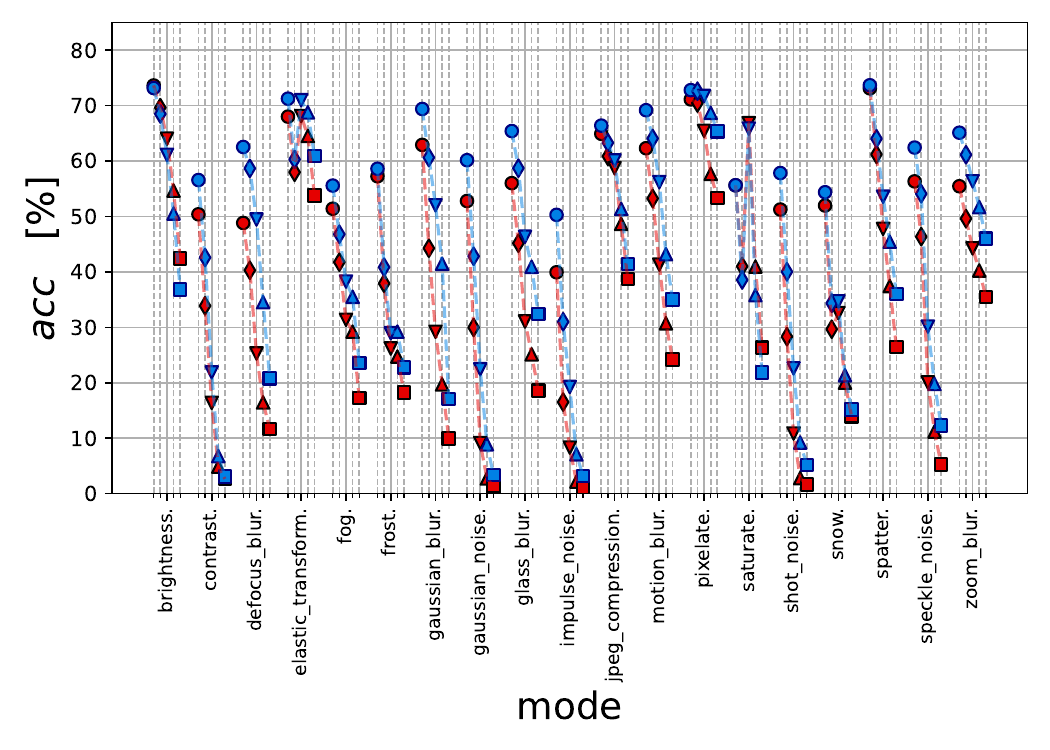}
    \caption{Accuracy for ResNeXt50 evaluated on ImageNet-100-C 2D common corruptions w/wo OpticsAugment training and all severities 1-5 (circle, diamond, triangles and square markers) at each corruption. \textbf{OpticsAugment (blue)} accuracy compared to the conventionally trained DNN (red). Results for more DNNs are available in supplementary~\ref{app:imagenet100_and_c},  Figs.~\ref{fig:app_imagenet100c_1}-~\ref{fig:app_imagenet100c_3}.}
    \label{fig:common2d_modes_resnext50}
\end{figure*}
For most of the corruptions the augmentation is beneficial, including blur, weather corruptions (fog, frost) and pixelation.  
However, some corruptions are not compensated by   OpticsAugment, e.g. JPEG compression and contrast. Noise robustness may be improved by chaining OpticsAugment with AugMix. 

In addition, pipelining AugMix with OpticsAugment further improves robustness  on 2D Common corruptions (especially to gaussian and speckle noise, cf. Table~\ref{tab:tab:imagenet100c_corruptions_EfficientNet}), but reduces the accuracy on OpticsBench. Table~\ref{tab:pipelining_overview} shows the average improvement for a cascaded application of AugMix \& OpticsAugment and the different severities. The accuracy of EfficientNet increases by an additional $3.4\%$ points on average for 2D common corruptions. 
\begin{table}[h]
\caption{Additional average improvement for DNNs with pipelining evaluated on ImageNet-100-c 2D common corruptions~\cite{hendrycks_benchmarking_2019}.}
\centering\begin{tabular}{llllll}
\footnotesize{DNN} & \footnotesize{$1$} & \footnotesize{$2$} & \footnotesize{$3$} & \footnotesize{$4$} & \footnotesize{$5$} 
\\\hline\footnotesize{EfficientNet} & \footnotesize{+2.79} & \footnotesize{+3.89} & \footnotesize{+4.34} & \footnotesize{+3.43} & \footnotesize{+2.61}\\
\footnotesize{MobileNet} & \footnotesize{+1.59} & \footnotesize{+2.27} & \footnotesize{+1.75} & \footnotesize{+0.57} & \footnotesize{-0.32}
\end{tabular}
\label{tab:pipelining_overview}
\end{table}
\FloatBarrier

\noindent\textbf{Limitations. } First, with OpticsBench and the according augmentation, we make only one further step towards realistic benchmarking for robustness. As we can only test for sample classes of aberrations here, we point users to the kernel generation tool to test their specific use case.
While we show that the proposed augmentation is beneficial for other types of common corruptions ~\cite{hendrycks_benchmarking_2019}, we also evaluate towards adversarial robustness on ImageNet-100 test batches for $l_2$  bounded attacks from AutoAttack~\cite{croce_reliable_2020} and $\epsilon =4/255$ in the supplementary material~\ref{app:additional_analysis}, table~\ref{tab:app:adversarial_robustness}. The evaluation is done for both APGD-CE and APGD-DLR attacks using the joint version of APGD and EoT for random defenses~\cite{athalye_obfuscated_2018} and 5 restarts.
However, the analysis does not indicate any benefits from OpticsAugment for adversarial robustness.  
\section{Conclusion}
\label{sec:conclusion}
This paper proposes the use of optical kernels obtained from optics to benchmark model robustness to aberrations. These 3D kernels (x, y, color) have diverse shape compared to disk-shaped kernels and depend on color. To compare their influence to a baseline, they are matched to disk-shaped kernels by minimizing various optical and image quality metrics, and provide OpticsBench, a benchmark aimed at testing for lens aberrations. We show empirically on ImageNet that a large number of DNNs can handle the optical corruptions differently well and conclude that these diverse blur types should be considered.

In addition, we investigate a training method to achieve robustness on OpticsBench: OpticsAugment efficiently generates an augmentation for each image and epoch by randomly selecting a blur kernel and convolving it with the image. Augmentation with OpticsAugment is beneficial beyond OpticsBench, for example for different types of 2D common corruptions. An average performance gain of 6.8\% compared to a DNN without augmentation is achieved across a large number of corruptions, with peaks of up to 29\% at medium severities.

\paragraph*{Acknowledgements}
The computations were supported by the OMNI Cluster of the University of Siegen.
\clearpage
\printbibliography

\clearpage
\appendix
\section*{Supplementary material}
This supplementary provides additional analyses into all evaluated datasets and settings. 

\begin{itemize}
    \item In App.~\ref{app:ranking} the achieved accuracies on OpticsBench (ImageNet) for the 70 different DNNs are visualized. The ranking compares all corruptions to the baseline (defocus blur)~\cite{hendrycks_benchmarking_2019} together with Kendall's $\tau$ rank correlation coefficient.
    \item App.~\ref{app:imagenet100_and_c} further quantifies with tables the benefit for OpticsAugment training on ImageNet-100 OpticsBench. Additionally, the accuracies for 2D common corruptions w/wo OpticsAugment training are listed. On top of that for all DNNs the comparison plots are shown on both benchmarks.
    \item Subsequently, App.~\ref{app:kernels} gives more insight into kernel generation and visualizes all OpticsBench and OpticsBenchRG kernels as used for the presented analysis. 
    \item Image examples can be found in the supplementary material App.~\ref{app:image_examples}
    \item Additional exemplary analysis on OpticsBenchRG with reddish and greenish kernels only is displayed in App.~\ref{app:opticsbenchRG}. The benchmark is both evaluated on ImageNet and ImageNet-100 showing again the ranking for 70 DNNs on ImageNet with Kendall's $\tau$ and on ImageNet-100 the averaged accuracies of w/wo OpticsAugment training.
    \item App.~\ref{app:additional_analysis} concludes this supplementary with general implementation details, hyperparameter settings for training and computational resources. Additionally, a DNN trained on ImageNet using OpticsBench supplements existing analysis and an experiment with pipelining 
    OpticsAugment and AugMix on ImageNet during training shows another possible use-case of the proposed OpticsAugment method as described in the main paper.
\end{itemize}


\section{Ranking comparisons} 
\label{app:ranking}
\begin{figure*}[h]
\begin{tabular}{@{}c@{}}
    \includegraphics[width=0.9\linewidth]{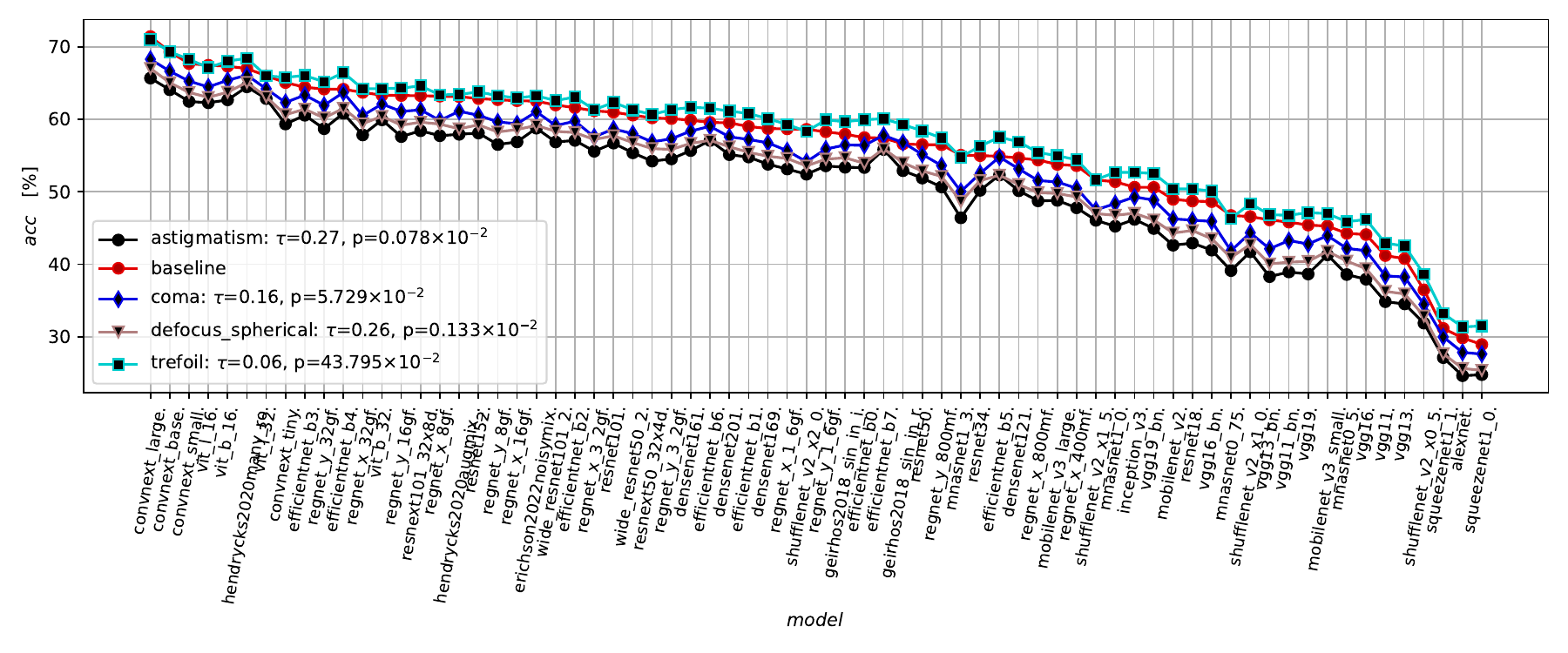}\\
    \footnotesize severity 1\\
    \includegraphics[width=0.9\linewidth]{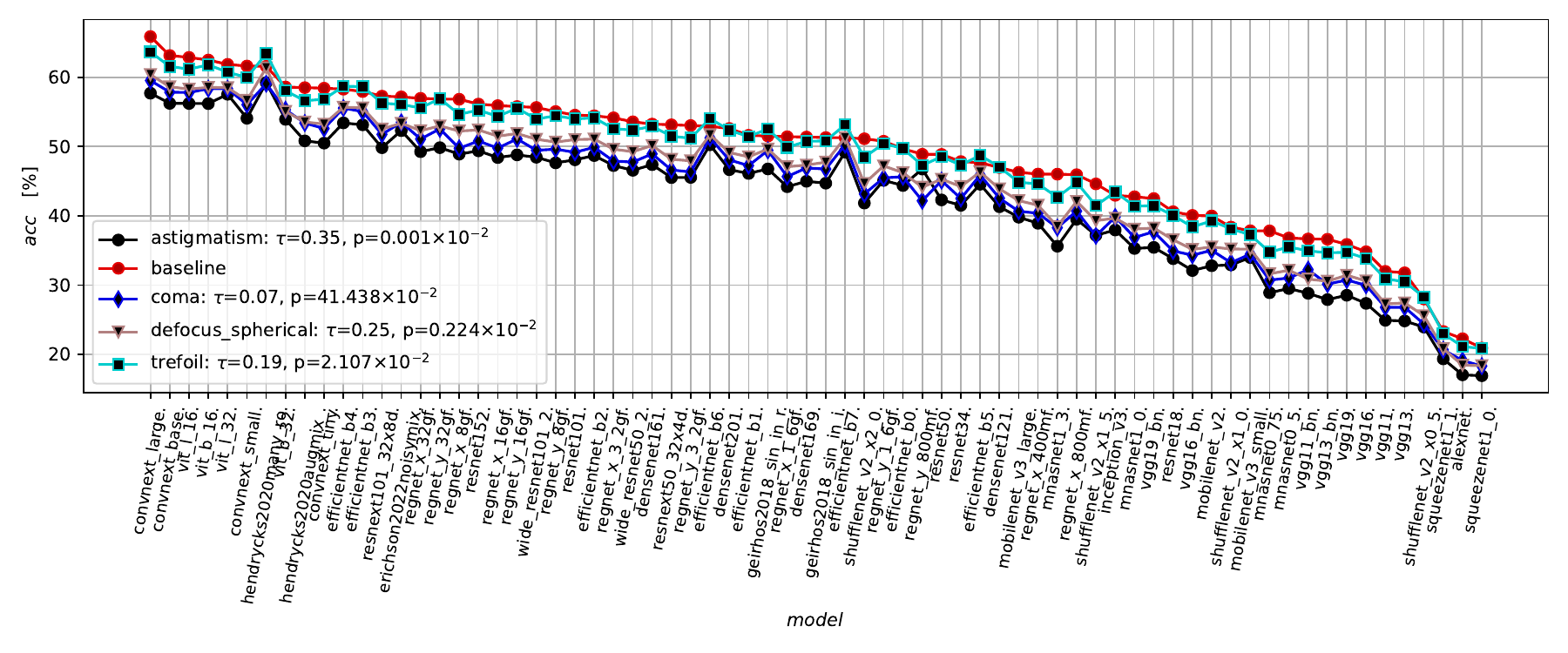}\\
    \footnotesize severity 2\\
     \includegraphics[width=0.9\linewidth]{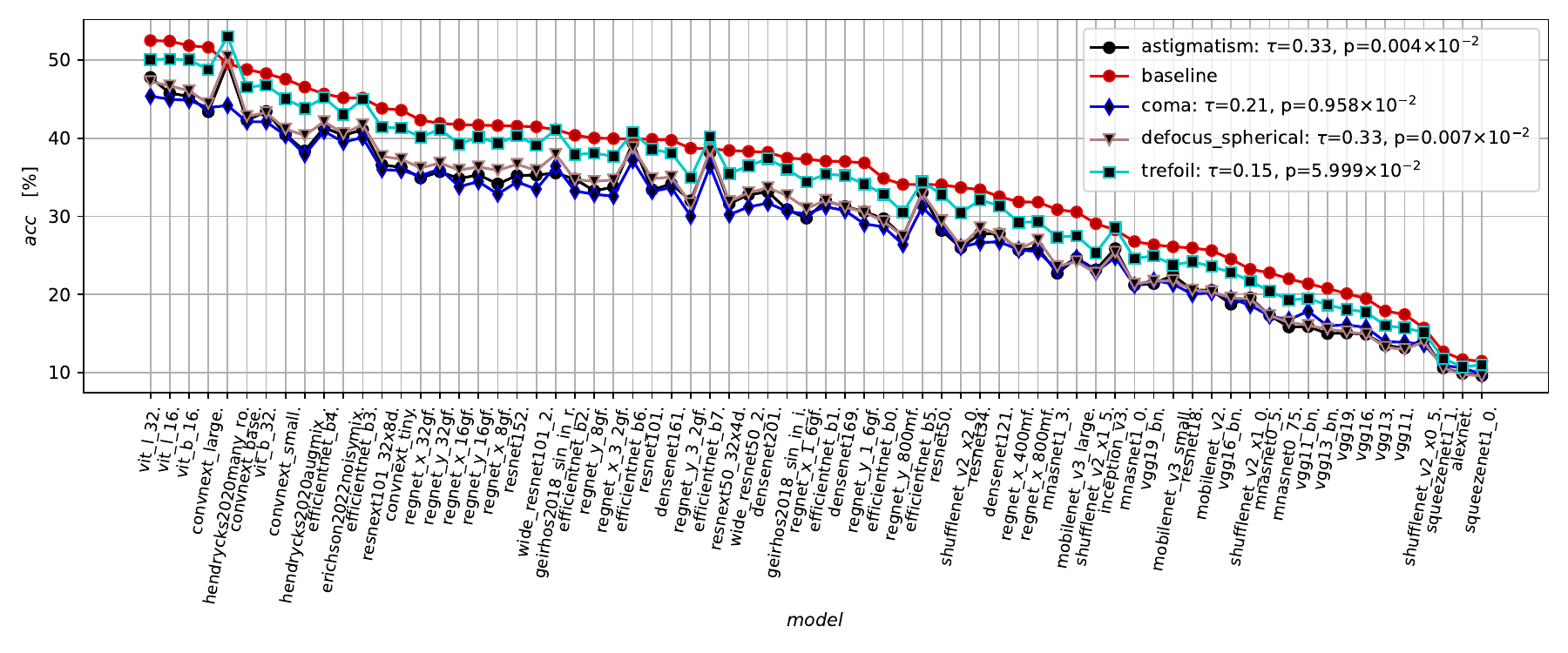}\\
   \footnotesize severity 3\\
    \end{tabular}
    \caption{Ranking comparison of baseline and all corruptions for severities 1-3.}
    \label{fig:ranking_sev12}
\end{figure*}

\begin{figure*}[h]
\begin{tabular}{@{}c@{}}
\footnotesize
    \includegraphics[width=0.9\linewidth]{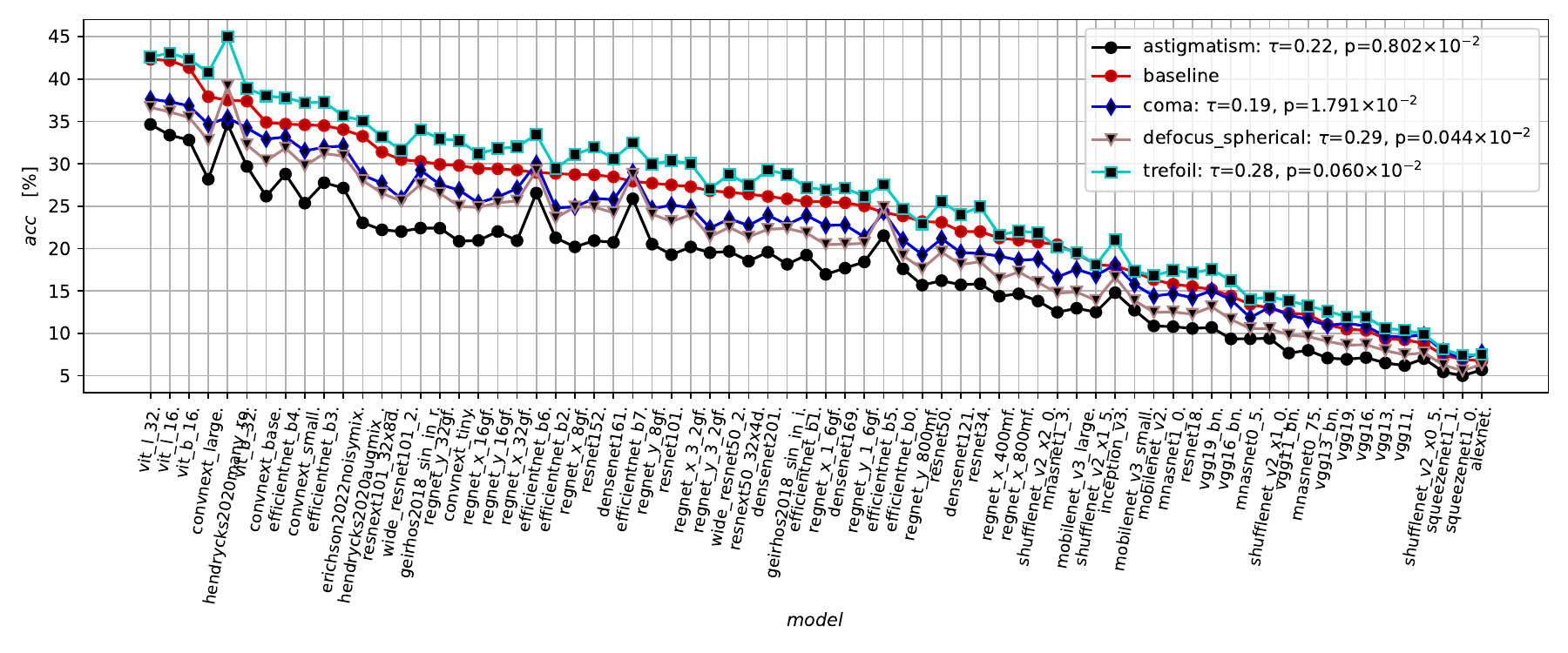}\\
   \footnotesize severity 4\\
    \includegraphics[width=0.9\linewidth]{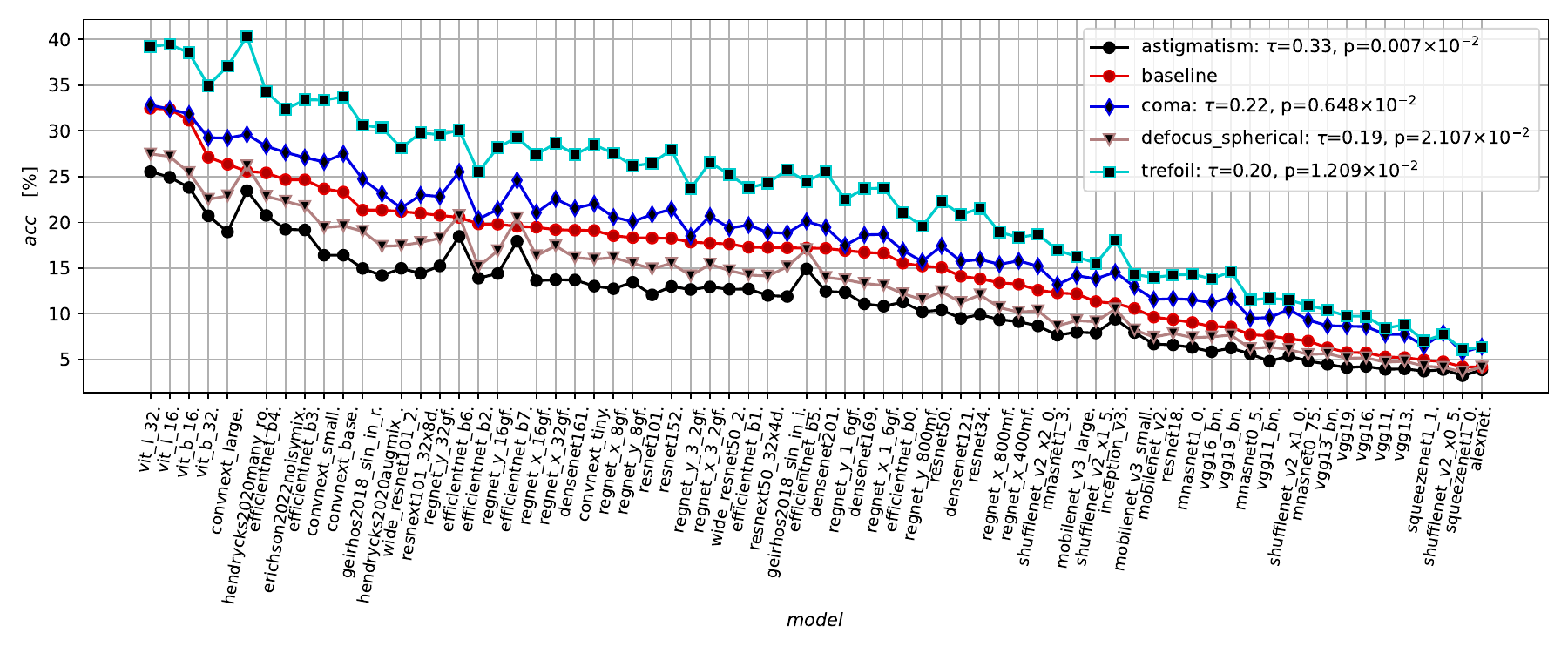}\\
   \footnotesize severity 5\\
    \end{tabular}
    \caption{Ranking comparison of baseline and all corruptions for severities 4-5 .}
    \label{fig:ranking_sev34}
\end{figure*}

The rank comparisons provided in Figure~\ref{fig:ranking_sev12} and Figure~\ref{fig:ranking_sev34} show the correlation between defocus blur~\cite{hendrycks_benchmarking_2019} (baseline) ordering and the different corruptions. This additional analysis confirms that DNNs can handle the blur kernel types from OpticsBench differently well. Additionally to the visualization of the accuracies, the  Kendall's rank coefficient $\tau$ is evaluated~\cite{kendall_treatment_1945}. 
A weak correlation is indicated by Kendall's $\tau \ll 1.0$. The $p$-value denotes the result of a hypothesis test for $\tau=0.0$, alternative $\tau \neq 0.0$. Most of the corruptions $\tau_{mean}=0.276$ are weakly correlated with defocus blur $\tau \leq 0.3$. Robust ResNet50 models from the RobustBench leaderboard using DeepAugment (\emph{hendrycks2020many} in our plots)~\cite{hendrycks_many_2021} or AugMix~\cite{hendrycks_augmix_2020} are among the top 10 DNNs for different severities. Besides this, VisionTransformer DNNs are always among top 5 and also include augmentation during training. The EfficientNet architecture achieves also good results on OpticsBench.
%
\newcommand{\szRanks}{0.9\linewidth}

\pagebreak
\section{OpticsAugment}
\label{app:imagenet100_and_c}

This appendix displays additional tables for the ImageNet-100 OpticsBench and ImageNet-100-C analysis to quantify the benefit with OpticsAugment training. The different tables show for OpticsBench accuaracies together with the improvement in accuracy $\Delta$ between our method and conventionally trained DNNs. We include the same tables for the evaluation on 2D common corruptions. All values are given in $\%$.
First, we report the accuracies on the ImageNet-100 validation dataset in Tab.~\ref{tab:app_validation_accs}.

\begin{table}[t]
    \centering
\footnotesize
    \begin{tabular}{llllll}
    DNN & wo & w (ours) & DNN & wo & w (ours) \\
    \hline
      DenseNet   &  79.2 & \textbf{80.2} &       ResNet101 & \textbf{82.6} & 80.6 \\
      EfficientNet   & 78.0 & \textbf{78.2} &       ResNeXt50 & \textbf{77.4} & 75.6 \\
      MobileNet & 76.2 & \textbf{76.6} &&&\\
    \end{tabular}

    \caption{Validation accuracies on ImageNet-100 for with (w) and without (wo) OpticsAugment trained DNNs. Both training schemes perform similarly well on the unmodified validation dataset.}
    \label{tab:app_validation_accs}
\end{table}

Tab.~\ref{tab:tab:imagenet100_avg_absolute} gives an overview for the achieved accuracies on OpticsBench ImageNet-100 and Tab.~\ref{tab:tab:imagenet100c_avg_absolute} for 2D common corruptions ImageNet-100-C. Although the validation accuracies are similar for w/wo OpticsAugment training, as the images are corrupted, the benefit with OpticsAugment becomes clear. 
The analysis is then extended to OpticsBench corruptions in Tables~\ref{tab:tab:imagenet100_corruptions_DenseNet_revisited}-
\ref{tab:tab:imagenet100_corruptions_ResNeXt50_revisited}. We also report the defocus blur corruption~\cite{hendrycks_benchmarking_2019} for comparison. All DNNs benefit from the OpticsAugment training scheme. However, the performance gain is almost for all settings lowest for defocus blur, which had been out of domain during training. This gives further proof that kernel types matter. 
The tables are visualized in Fig.~\ref{fig:app_optics_augment2} and~\ref{fig:app_optics_augment1}. The visualization is also offered for DenseNet161 and ResNeXt50 to allow for another point of view compared to Fig.~\ref{fig:imagenet100_resnext50} and~\ref{fig:imagenet100_densenet161}.

Tables~\ref{tab:tab:imagenet100c_corruptions_DenseNet}-\ref{tab:tab:imagenet100c_corruptions_ResNeXt50} show for all 19 corruptions from~\cite{hendrycks_benchmarking_2019} the particular accuracies and the difference in accuracy $\Delta$. 
%
\begin{table}[h]\centering
\begin{footnotesize}
\begin{tabular}{llllll}Model & 1 & 2 & 3 & 4 & 5 \\ \hline \\DenseNet \textbf{(ours)} & \textbf{68.22} & \textbf{65.33} & \textbf{56.33} & \textbf{41.60} & \textbf{30.13} \\ DenseNet & 53.45 & 43.37 & 29.07 & 20.62 & 16.30\\EfficientNet \textbf{(ours)} & \textbf{61.00} & \textbf{55.34} & \textbf{42.14} & \textbf{30.27} & \textbf{23.35} \\ EfficientNet & 52.55 & 42.74 & 29.24 & 20.84 & 16.00\\MobileNet \textbf{(ours)} & \textbf{57.59} & \textbf{52.30} & \textbf{38.58} & \textbf{27.51} & \textbf{20.54} \\ MobileNet & 49.47 & 39.57 & 24.78 & 17.42 & 13.27\\ResNet101 \textbf{(ours)} & \textbf{69.90} & \textbf{67.68} & \textbf{61.36} & \textbf{49.04} & \textbf{37.80} \\ ResNet101 & 59.92 & 51.44 & 40.21 & 31.65 & 25.73\\ResNeXt50 \textbf{(ours)} & \textbf{65.14} & \textbf{62.68} & \textbf{54.44} & \textbf{39.90} & \textbf{28.45} \\ ResNeXt50 & 47.74 & 38.19 & 24.88 & 17.58 & 13.69\\\end{tabular}
\end{footnotesize}
\caption{Accuracies w/wo OpticsAugment evaluated on ImageNet-100 OpticsBench. Average over all corruptions.}\label{tab:tab:imagenet100_avg_absolute}\end{table}
\newcommand{\szOptAug}{0.8\linewidth}
\begin{figure}[h]
    \centering
    \begin{subfigure}{\szOptAug}
    \includegraphics[width=\linewidth]{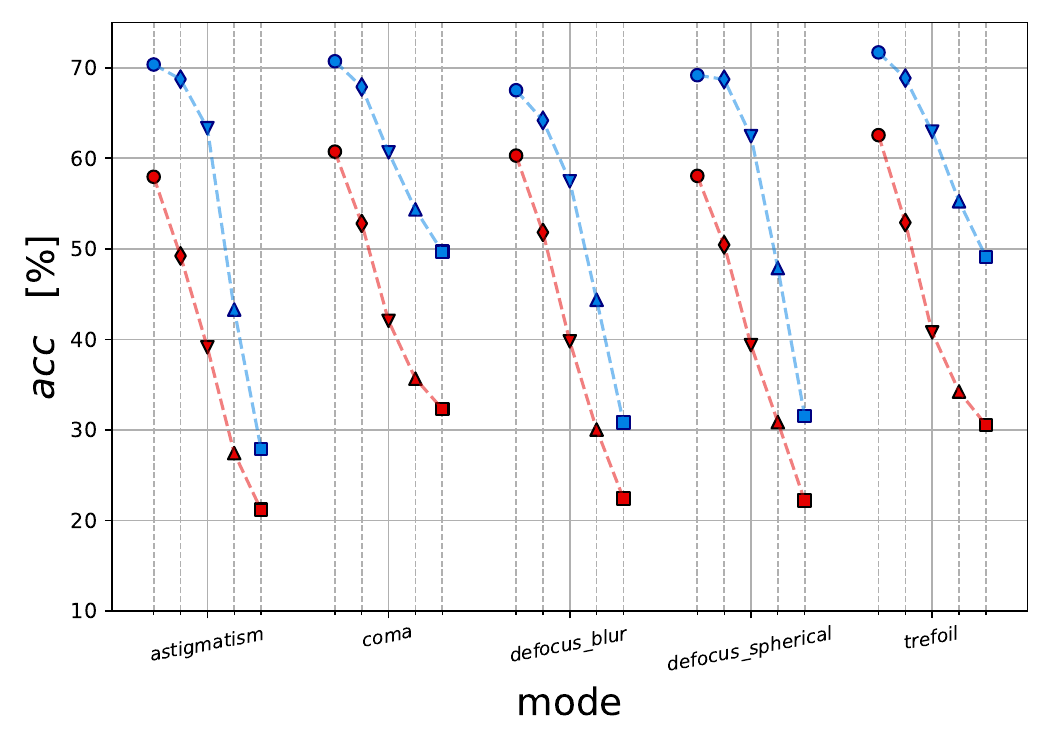}
    \caption{ResNet101}
    \label{fig:app_imagenet100_resnet101}
    \end{subfigure}    
    \begin{subfigure}{\szOptAug}
    \includegraphics[width=\linewidth]{figures/revisited/imagenet100/eval_rgb/report_acc_over_modesplit_corruptions_highlight_resnext50_32x4d_sgd_clean_resnext50_32x4d_sgd_opticsaugment.pdf}
    \caption{ResNeXt50}
    \label{fig:app_imagenet100_resnext50}
    \end{subfigure}
    \caption{Accuracy evaluated on OpticsBench-ImageNet-100 for DNNs w/wo OpticsAugment training and all severities 1-5 (circle, diamond, triangles and square markers) at each corruption. 
    \textbf{OpticsAugment (blue) improves} accuracy compared to the conventionally trained DNN (red).}
    \label{fig:app_optics_augment1}
\end{figure}

\begin{figure}[h]
    \centering
    \begin{subfigure}{\szOptAug}
    \includegraphics[width=\linewidth]{figures/revisited/imagenet100/eval_rgb/report_acc_over_modesplit_corruptions_highlight_densenet161_sgd_clean_densenet161_sgd_opticsaugment.pdf}
    \caption{DenseNet161}
    \label{fig:app_imagenet100_densenet161}
    \end{subfigure} 
    \begin{subfigure}{\szOptAug}
    \includegraphics[width=\linewidth]{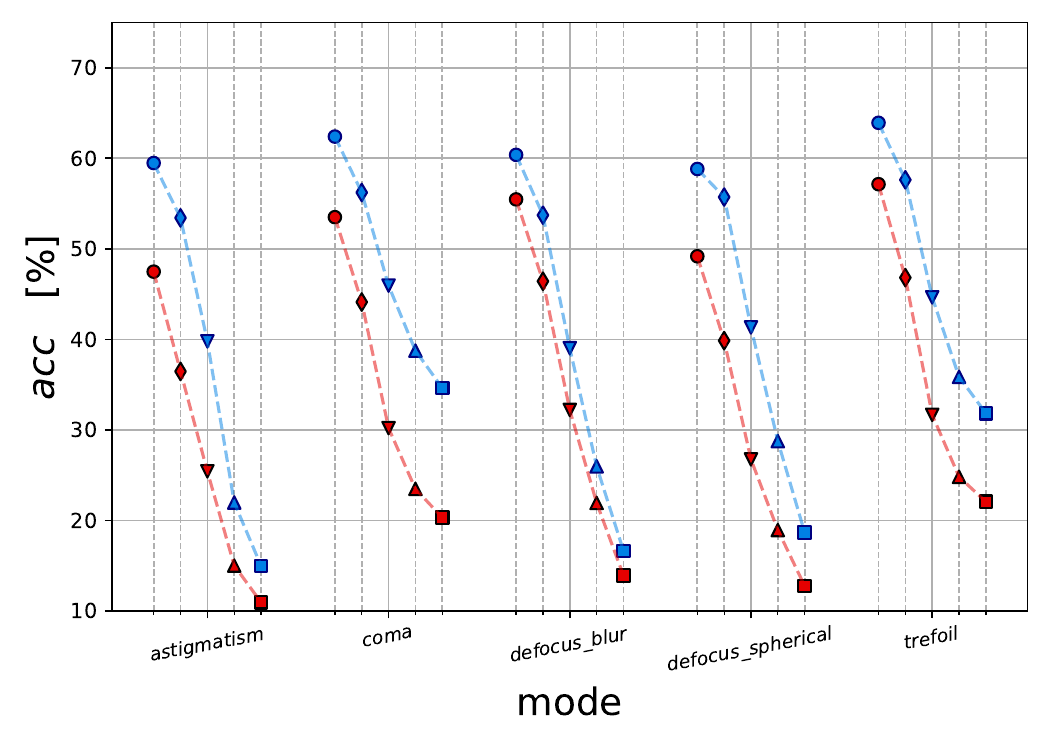}
    \caption{EfficientNet}
    \label{fig:app_imagenet100_efficientnet_b0}
    \end{subfigure}
    \begin{subfigure}{\szOptAug}
    \includegraphics[width=\linewidth]{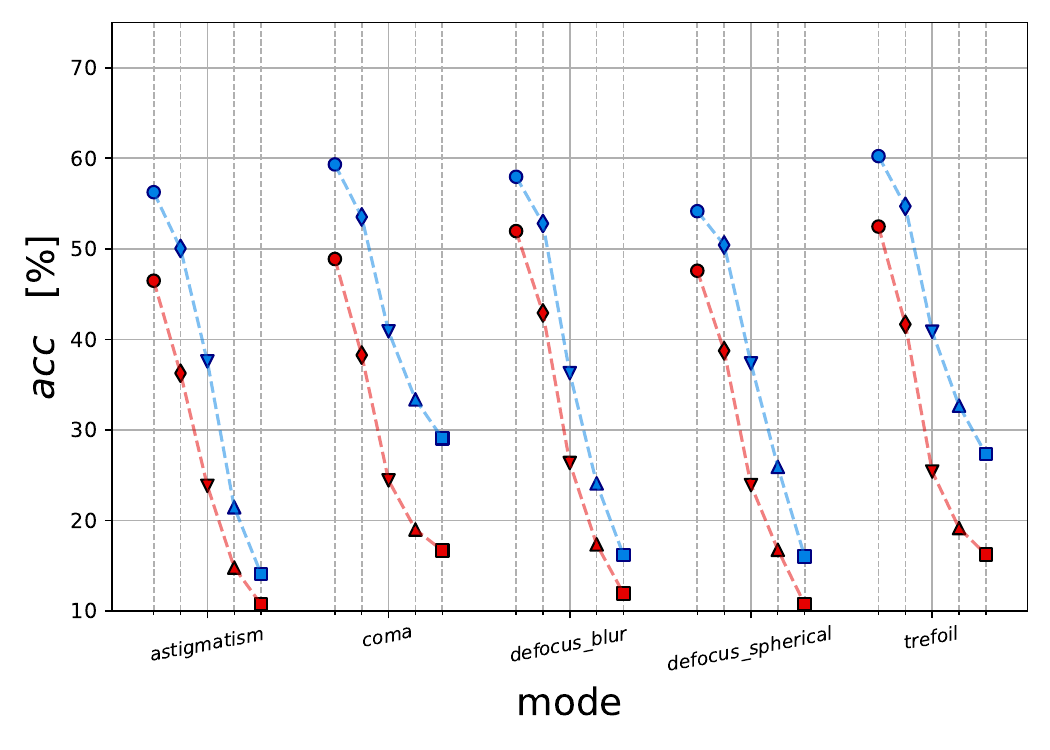}
    \caption{MobileNet}
    \label{fig:app_imagenet100_mobilenet_v3_large}
    \end{subfigure}   
    \caption{ Accuracy evaluated on OpticsBench-ImageNet-100 for DNNs w/wo OpticsAugment training and all severities 1-5 (circle, diamond, triangles and square markers) at each corruption. \textbf{OpticsAugment (blue) improves} accuracy compared to the conventionally trained DNN (red):  (a) DenseNet, (b) EfficientNet, (c) MobileNet.}
    \label{fig:app_optics_augment2}
\end{figure}
\FloatBarrier


\FloatBarrier

\begin{table*}[h]\centering\begin{tabular}{llll|lll|lll|lll|lll}&\multicolumn{3}{c}{\scriptsize{1}}&\multicolumn{3}{c}{\scriptsize{2}}&\multicolumn{3}{c}{\scriptsize{3}}&\multicolumn{3}{c}{\scriptsize{4}}&\multicolumn{3}{c}{\scriptsize{5}}\\\tiny{Corruption} & \tiny{clean} & \tiny{\textbf{ours}} & \tiny{$\Delta$} & \tiny{clean} & \tiny{\textbf{ours}} & \tiny{$\Delta$} & \tiny{clean} & \tiny{\textbf{ours}} & \tiny{$\Delta$} & \tiny{clean} & \tiny{\textbf{ours}} & \tiny{$\Delta$} & \tiny{clean} & \tiny{\textbf{ours}} & \tiny{$\Delta$}\\\hline\tiny{astigmatism} & \tiny{50.26} & \textbf{\tiny{67.68}} & \tiny{17.42} & \tiny{39.54} & \textbf{\tiny{65.72}} & \tiny{26.18} & \tiny{26.80} & \textbf{\tiny{57.02}} & \tiny{30.22} & \tiny{16.22} & \textbf{\tiny{32.52}} & \tiny{16.30} & \tiny{12.34} & \textbf{\tiny{18.84}} & \tiny{6.50}\\\tiny{coma} & \tiny{54.28} & \textbf{\tiny{70.54}} & \tiny{16.26} & \tiny{43.54} & \textbf{\tiny{67.36}} & \tiny{23.82} & \tiny{29.46} & \textbf{\tiny{57.78}} & \tiny{28.32} & \tiny{23.84} & \textbf{\tiny{50.08}} & \tiny{26.24} & \tiny{20.60} & \textbf{\tiny{42.84}} & \tiny{22.24}\\\tiny{defocus\_blur} & \tiny{55.46} & \textbf{\tiny{64.80}} & \tiny{9.34} & \tiny{46.76} & \textbf{\tiny{59.84}} & \tiny{13.08} & \tiny{31.96} & \textbf{\tiny{50.44}} & \tiny{18.48} & \tiny{21.18} & \textbf{\tiny{36.00}} & \tiny{14.82} & \tiny{14.46} & \textbf{\tiny{22.40}} & \tiny{7.94}\\\tiny{defocus\_spherical} & \tiny{49.92} & \textbf{\tiny{67.14}} & \tiny{17.22} & \tiny{41.48} & \textbf{\tiny{66.20}} & \tiny{24.72} & \tiny{26.98} & \textbf{\tiny{56.68}} & \tiny{29.70} & \tiny{18.96} & \textbf{\tiny{39.32}} & \tiny{20.36} & \tiny{13.18} & \textbf{\tiny{23.52}} & \tiny{10.34}\\\tiny{trefoil} & \tiny{57.34} & \textbf{\tiny{70.96}} & \tiny{13.62} & \tiny{45.52} & \textbf{\tiny{67.54}} & \tiny{22.02} & \tiny{30.14} & \textbf{\tiny{59.74}} & \tiny{29.60} & \tiny{22.90} & \textbf{\tiny{50.06}} & \tiny{27.16} & \tiny{20.90} & \textbf{\tiny{43.06}} & \tiny{22.16}\\\tiny{{{$\Sigma$}}} & \tiny{53.45} & \textbf{\tiny{68.22}} & \tiny{14.77} & \tiny{43.37} & \textbf{\tiny{65.33}} & \tiny{21.96} & \tiny{29.07} & \textbf{\tiny{56.33}} & \tiny{27.26} & \tiny{20.62} & \textbf{\tiny{41.60}} & \tiny{20.98} & \tiny{16.30} & \textbf{\tiny{30.13}} & \tiny{13.84}\end{tabular}\caption{Accuracies for DenseNet w/wo OpticsAugment evaluated on ImageNet-100 OpticsBench.}\label{tab:tab:imagenet100_corruptions_DenseNet_revisited}\end{table*}
\begin{table*}[h]\centering\begin{tabular}{llll|lll|lll|lll|lll}&\multicolumn{3}{c}{\scriptsize{1}}&\multicolumn{3}{c}{\scriptsize{2}}&\multicolumn{3}{c}{\scriptsize{3}}&\multicolumn{3}{c}{\scriptsize{4}}&\multicolumn{3}{c}{\scriptsize{5}}\\\tiny{Corruption} & \tiny{clean} & \tiny{\textbf{ours}} & \tiny{$\Delta$} & \tiny{clean} & \tiny{\textbf{ours}} & \tiny{$\Delta$} & \tiny{clean} & \tiny{\textbf{ours}} & \tiny{$\Delta$} & \tiny{clean} & \tiny{\textbf{ours}} & \tiny{$\Delta$} & \tiny{clean} & \tiny{\textbf{ours}} & \tiny{$\Delta$}\\\hline\tiny{astigmatism} & \tiny{47.48} & \textbf{\tiny{59.48}} & \tiny{12.00} & \tiny{36.46} & \textbf{\tiny{53.42}} & \tiny{16.96} & \tiny{25.42} & \textbf{\tiny{39.78}} & \tiny{14.36} & \tiny{15.02} & \textbf{\tiny{21.98}} & \tiny{6.96} & \tiny{10.96} & \textbf{\tiny{14.98}} & \tiny{4.02}\\\tiny{coma} & \tiny{53.50} & \textbf{\tiny{62.40}} & \tiny{8.90} & \tiny{44.12} & \textbf{\tiny{56.22}} & \tiny{12.10} & \tiny{30.20} & \textbf{\tiny{45.94}} & \tiny{15.74} & \tiny{23.48} & \textbf{\tiny{38.76}} & \tiny{15.28} & \tiny{20.32} & \textbf{\tiny{34.64}} & \tiny{14.32}\\\tiny{defocus\_blur} & \tiny{55.46} & \textbf{\tiny{60.38}} & \tiny{4.92} & \tiny{46.42} & \textbf{\tiny{53.72}} & \tiny{7.30} & \tiny{32.20} & \textbf{\tiny{39.02}} & \tiny{6.82} & \tiny{21.94} & \textbf{\tiny{25.98}} & \tiny{4.04} & \tiny{13.92} & \textbf{\tiny{16.64}} & \tiny{2.72}\\\tiny{defocus\_spherical} & \tiny{49.18} & \textbf{\tiny{58.82}} & \tiny{9.64} & \tiny{39.86} & \textbf{\tiny{55.72}} & \tiny{15.86} & \tiny{26.74} & \textbf{\tiny{41.32}} & \tiny{14.58} & \tiny{18.94} & \textbf{\tiny{28.76}} & \tiny{9.82} & \tiny{12.76} & \textbf{\tiny{18.66}} & \tiny{5.90}\\\tiny{trefoil} & \tiny{57.14} & \textbf{\tiny{63.92}} & \tiny{6.78} & \tiny{46.82} & \textbf{\tiny{57.62}} & \tiny{10.80} & \tiny{31.66} & \textbf{\tiny{44.64}} & \tiny{12.98} & \tiny{24.82} & \textbf{\tiny{35.86}} & \tiny{11.04} & \tiny{22.06} & \textbf{\tiny{31.84}} & \tiny{9.78}\\\tiny{{{$\Sigma$}}} & \tiny{52.55} & \textbf{\tiny{61.00}} & \tiny{8.45} & \tiny{42.74} & \textbf{\tiny{55.34}} & \tiny{12.60} & \tiny{29.24} & \textbf{\tiny{42.14}} & \tiny{12.90} & \tiny{20.84} & \textbf{\tiny{30.27}} & \tiny{9.43} & \tiny{16.00} & \textbf{\tiny{23.35}} & \tiny{7.35}\end{tabular}\caption{Accuracies for EfficientNet w/wo OpticsAugment evaluated on ImageNet-100 OpticsBench.}\label{tab:tab:imagenet100_corruptions_EfficientNet_revisited}\end{table*}
\begin{table*}[h]\centering\begin{tabular}{llll|lll|lll|lll|lll}&\multicolumn{3}{c}{\scriptsize{1}}&\multicolumn{3}{c}{\scriptsize{2}}&\multicolumn{3}{c}{\scriptsize{3}}&\multicolumn{3}{c}{\scriptsize{4}}&\multicolumn{3}{c}{\scriptsize{5}}\\\tiny{Corruption} & \tiny{clean} & \tiny{\textbf{ours}} & \tiny{$\Delta$} & \tiny{clean} & \tiny{\textbf{ours}} & \tiny{$\Delta$} & \tiny{clean} & \tiny{\textbf{ours}} & \tiny{$\Delta$} & \tiny{clean} & \tiny{\textbf{ours}} & \tiny{$\Delta$} & \tiny{clean} & \tiny{\textbf{ours}} & \tiny{$\Delta$}\\\hline\tiny{astigmatism} & \tiny{46.48} & \textbf{\tiny{56.26}} & \tiny{9.78} & \tiny{36.26} & \textbf{\tiny{50.04}} & \tiny{13.78} & \tiny{23.82} & \textbf{\tiny{37.56}} & \tiny{13.74} & \tiny{14.80} & \textbf{\tiny{21.46}} & \tiny{6.66} & \tiny{10.74} & \textbf{\tiny{14.10}} & \tiny{3.36}\\\tiny{coma} & \tiny{48.88} & \textbf{\tiny{59.32}} & \tiny{10.44} & \tiny{38.26} & \textbf{\tiny{53.52}} & \tiny{15.26} & \tiny{24.42} & \textbf{\tiny{40.88}} & \tiny{16.46} & \tiny{19.00} & \textbf{\tiny{33.38}} & \tiny{14.38} & \tiny{16.68} & \textbf{\tiny{29.08}} & \tiny{12.40}\\\tiny{defocus\_blur} & \tiny{51.96} & \textbf{\tiny{57.96}} & \tiny{6.00} & \tiny{42.92} & \textbf{\tiny{52.80}} & \tiny{9.88} & \tiny{26.34} & \textbf{\tiny{36.26}} & \tiny{9.92} & \tiny{17.36} & \textbf{\tiny{24.10}} & \tiny{6.74} & \tiny{11.94} & \textbf{\tiny{16.18}} & \tiny{4.24}\\\tiny{defocus\_spherical} & \tiny{47.58} & \textbf{\tiny{54.16}} & \tiny{6.58} & \tiny{38.74} & \textbf{\tiny{50.42}} & \tiny{11.68} & \tiny{23.90} & \textbf{\tiny{37.34}} & \tiny{13.44} & \tiny{16.76} & \textbf{\tiny{25.92}} & \tiny{9.16} & \tiny{10.74} & \textbf{\tiny{16.02}} & \tiny{5.28}\\\tiny{trefoil} & \tiny{52.46} & \textbf{\tiny{60.24}} & \tiny{7.78} & \tiny{41.66} & \textbf{\tiny{54.70}} & \tiny{13.04} & \tiny{25.40} & \textbf{\tiny{40.84}} & \tiny{15.44} & \tiny{19.16} & \textbf{\tiny{32.68}} & \tiny{13.52} & \tiny{16.24} & \textbf{\tiny{27.32}} & \tiny{11.08}\\\tiny{{{$\Sigma$}}} & \tiny{49.47} & \textbf{\tiny{57.59}} & \tiny{8.12} & \tiny{39.57} & \textbf{\tiny{52.30}} & \tiny{12.73} & \tiny{24.78} & \textbf{\tiny{38.58}} & \tiny{13.80} & \tiny{17.42} & \textbf{\tiny{27.51}} & \tiny{10.09} & \tiny{13.27} & \textbf{\tiny{20.54}} & \tiny{7.27}\end{tabular}\caption{Accuracies for MobileNet w/wo OpticsAugment evaluated on ImageNet-100 OpticsBench.}\label{tab:tab:imagenet100_corruptions_MobileNet_revisited}\end{table*}
\begin{table*}[h]\centering\begin{tabular}{llll|lll|lll|lll|lll}&\multicolumn{3}{c}{\scriptsize{1}}&\multicolumn{3}{c}{\scriptsize{2}}&\multicolumn{3}{c}{\scriptsize{3}}&\multicolumn{3}{c}{\scriptsize{4}}&\multicolumn{3}{c}{\scriptsize{5}}\\\tiny{Corruption} & \tiny{clean} & \tiny{\textbf{ours}} & \tiny{$\Delta$} & \tiny{clean} & \tiny{\textbf{ours}} & \tiny{$\Delta$} & \tiny{clean} & \tiny{\textbf{ours}} & \tiny{$\Delta$} & \tiny{clean} & \tiny{\textbf{ours}} & \tiny{$\Delta$} & \tiny{clean} & \tiny{\textbf{ours}} & \tiny{$\Delta$}\\\hline\tiny{astigmatism} & \tiny{57.96} & \textbf{\tiny{70.36}} & \tiny{12.40} & \tiny{49.22} & \textbf{\tiny{68.74}} & \tiny{19.52} & \tiny{39.12} & \textbf{\tiny{63.32}} & \tiny{24.20} & \tiny{27.44} & \textbf{\tiny{43.30}} & \tiny{15.86} & \tiny{21.18} & \textbf{\tiny{27.88}} & \tiny{6.70}\\\tiny{coma} & \tiny{60.74} & \textbf{\tiny{70.72}} & \tiny{9.98} & \tiny{52.80} & \textbf{\tiny{67.88}} & \tiny{15.08} & \tiny{42.04} & \textbf{\tiny{60.66}} & \tiny{18.62} & \tiny{35.66} & \textbf{\tiny{54.36}} & \tiny{18.70} & \tiny{32.32} & \textbf{\tiny{49.68}} & \tiny{17.36}\\\tiny{defocus\_blur} & \tiny{60.30} & \textbf{\tiny{67.52}} & \tiny{7.22} & \tiny{51.82} & \textbf{\tiny{64.18}} & \tiny{12.36} & \tiny{39.78} & \textbf{\tiny{57.46}} & \tiny{17.68} & \tiny{30.04} & \textbf{\tiny{44.38}} & \tiny{14.34} & \tiny{22.42} & \textbf{\tiny{30.80}} & \tiny{8.38}\\\tiny{defocus\_spherical} & \tiny{58.06} & \textbf{\tiny{69.18}} & \tiny{11.12} & \tiny{50.44} & \textbf{\tiny{68.72}} & \tiny{18.28} & \tiny{39.36} & \textbf{\tiny{62.44}} & \tiny{23.08} & \tiny{30.86} & \textbf{\tiny{47.88}} & \tiny{17.02} & \tiny{22.18} & \textbf{\tiny{31.54}} & \tiny{9.36}\\\tiny{trefoil} & \tiny{62.56} & \textbf{\tiny{71.70}} & \tiny{9.14} & \tiny{52.90} & \textbf{\tiny{68.86}} & \tiny{15.96} & \tiny{40.76} & \textbf{\tiny{62.94}} & \tiny{22.18} & \tiny{34.24} & \textbf{\tiny{55.26}} & \tiny{21.02} & \tiny{30.56} & \textbf{\tiny{49.12}} & \tiny{18.56}\\\tiny{{{$\Sigma$}}} & \tiny{59.92} & \textbf{\tiny{69.90}} & \tiny{9.97} & \tiny{51.44} & \textbf{\tiny{67.68}} & \tiny{16.24} & \tiny{40.21} & \textbf{\tiny{61.36}} & \tiny{21.15} & \tiny{31.65} & \textbf{\tiny{49.04}} & \tiny{17.39} & \tiny{25.73} & \textbf{\tiny{37.80}} & \tiny{12.07}\end{tabular}\caption{Accuracies for ResNet101 w/wo OpticsAugment evaluated on ImageNet-100 OpticsBench.}\label{tab:tab:imagenet100_corruptions_ResNet101_revisited}\end{table*}
\begin{table*}[h]\centering\begin{tabular}{llll|lll|lll|lll|lll}&\multicolumn{3}{c}{\scriptsize{1}}&\multicolumn{3}{c}{\scriptsize{2}}&\multicolumn{3}{c}{\scriptsize{3}}&\multicolumn{3}{c}{\scriptsize{4}}&\multicolumn{3}{c}{\scriptsize{5}}\\\tiny{Corruption} & \tiny{clean} & \tiny{\textbf{ours}} & \tiny{$\Delta$} & \tiny{clean} & \tiny{\textbf{ours}} & \tiny{$\Delta$} & \tiny{clean} & \tiny{\textbf{ours}} & \tiny{$\Delta$} & \tiny{clean} & \tiny{\textbf{ours}} & \tiny{$\Delta$} & \tiny{clean} & \tiny{\textbf{ours}} & \tiny{$\Delta$}\\\hline\tiny{astigmatism} & \tiny{44.08} & \textbf{\tiny{65.08}} & \tiny{21.00} & \tiny{34.44} & \textbf{\tiny{63.36}} & \tiny{28.92} & \tiny{22.98} & \textbf{\tiny{55.90}} & \tiny{32.92} & \tiny{14.24} & \textbf{\tiny{33.54}} & \tiny{19.30} & \tiny{10.26} & \textbf{\tiny{19.88}} & \tiny{9.62}\\\tiny{coma} & \tiny{49.28} & \textbf{\tiny{66.44}} & \tiny{17.16} & \tiny{39.28} & \textbf{\tiny{63.24}} & \tiny{23.96} & \tiny{26.64} & \textbf{\tiny{55.04}} & \tiny{28.40} & \tiny{21.42} & \textbf{\tiny{47.66}} & \tiny{26.24} & \tiny{18.66} & \textbf{\tiny{41.50}} & \tiny{22.84}\\\tiny{defocus\_blur} & \tiny{48.84} & \textbf{\tiny{62.54}} & \tiny{13.70} & \tiny{40.30} & \textbf{\tiny{58.68}} & \tiny{18.38} & \tiny{25.28} & \textbf{\tiny{49.46}} & \tiny{24.18} & \tiny{16.46} & \textbf{\tiny{34.54}} & \tiny{18.08} & \tiny{11.68} & \textbf{\tiny{20.82}} & \tiny{9.14}\\\tiny{defocus\_spherical} & \tiny{44.44} & \textbf{\tiny{64.06}} & \tiny{19.62} & \tiny{36.30} & \textbf{\tiny{63.26}} & \tiny{26.96} & \tiny{23.50} & \textbf{\tiny{55.00}} & \tiny{31.50} & \tiny{16.00} & \textbf{\tiny{37.80}} & \tiny{21.80} & \tiny{10.80} & \textbf{\tiny{21.26}} & \tiny{10.46}\\\tiny{trefoil} & \tiny{52.04} & \textbf{\tiny{67.58}} & \tiny{15.54} & \tiny{40.62} & \textbf{\tiny{64.84}} & \tiny{24.22} & \tiny{25.98} & \textbf{\tiny{56.78}} & \tiny{30.80} & \tiny{19.78} & \textbf{\tiny{45.94}} & \tiny{26.16} & \tiny{17.04} & \textbf{\tiny{38.78}} & \tiny{21.74}\\\tiny{{{$\Sigma$}}} & \tiny{47.74} & \textbf{\tiny{65.14}} & \tiny{17.40} & \tiny{38.19} & \textbf{\tiny{62.68}} & \tiny{24.49} & \tiny{24.88} & \textbf{\tiny{54.44}} & \tiny{29.56} & \tiny{17.58} & \textbf{\tiny{39.90}} & \tiny{22.32} & \tiny{13.69} & \textbf{\tiny{28.45}} & \tiny{14.76}\end{tabular}\caption{Accuracies for ResNeXt50 w/wo OpticsAugment evaluated on ImageNet-100 OpticsBench.}\label{tab:tab:imagenet100_corruptions_ResNeXt50_revisited}\end{table*}

\begin{table*}[ht]\footnotesize\centering\begin{tabular}{@{}llllll@{}}Model & 1 & 2 & 3 & 4 & 5 \\ \hline \\DenseNet \textbf{(ours)} & \textbf{67.99} & \textbf{57.65} & \textbf{49.24} & \textbf{38.18} & \textbf{28.35} \\ DenseNet & 62.91 & 50.09 & 40.50 & 30.88 & 22.97\\EfficientNet \textbf{(ours)} & \textbf{63.89} & \textbf{53.36} & \textbf{45.14} & \textbf{34.66} & \textbf{26.02} \\ EfficientNet & 59.54 & 47.04 & 38.44 & 30.04 & 22.33\\MobileNet \textbf{(ours)} & \textbf{60.87} & \textbf{50.35} & \textbf{42.43} & \textbf{33.07} & \textbf{25.10} \\ MobileNet & 57.29 & 45.43 & 37.66 & 29.38 & 22.03\\ResNet101 \textbf{(ours)} & \textbf{69.13} & \textbf{60.15} & \textbf{52.93} & \textbf{42.70} & \textbf{33.05} \\ ResNet101 & 67.89 & 57.07 & 48.38 & 37.80 & 28.95\\ResNeXt50 \textbf{(ours)} & \textbf{63.19} & \textbf{52.80} & \textbf{45.36} & \textbf{35.29} & \textbf{26.44} \\ ResNeXt50 & 58.07 & 45.17 & 36.68 & 28.11 & 21.17\\\end{tabular}\caption{Average Accuracies w/wo OpticsAugment evaluated on ImageNet-100-c 2D common corruptions~\cite{hendrycks_benchmarking_2019}. Average over all corruptions.}\label{tab:tab:imagenet100c_avg_absolute}\end{table*}

\begin{table*}[h]\centering\begin{tabular}{llll|lll|lll|lll|lll}&\multicolumn{3}{c}{\scriptsize{1}}&\multicolumn{3}{c}{\scriptsize{2}}&\multicolumn{3}{c}{\scriptsize{3}}&\multicolumn{3}{c}{\scriptsize{4}}&\multicolumn{3}{c}{\scriptsize{5}}\\\tiny{Corruption} & \tiny{clean} & \tiny{\textbf{ours}} & \tiny{$\Delta$} & \tiny{clean} & \tiny{\textbf{ours}} & \tiny{$\Delta$} & \tiny{clean} & \tiny{\textbf{ours}} & \tiny{$\Delta$} & \tiny{clean} & \tiny{\textbf{ours}} & \tiny{$\Delta$} & \tiny{clean} & \tiny{\textbf{ours}} & \tiny{$\Delta$}\\\hline\tiny{brightness} & \tiny{76.04} & \textbf{\tiny{78.24}} & \tiny{2.20} & \tiny{72.78} & \textbf{\tiny{75.34}} & \tiny{2.56} & \tiny{67.86} & \textbf{\tiny{70.52}} & \tiny{2.66} & \tiny{59.72} & \textbf{\tiny{63.08}} & \tiny{3.36} & \tiny{48.06} & \textbf{\tiny{52.70}} & \tiny{4.64}\\\tiny{contrast} & \tiny{57.78} & \textbf{\tiny{65.40}} & \tiny{7.62} & \tiny{44.76} & \textbf{\tiny{54.46}} & \tiny{9.70} & \tiny{25.98} & \textbf{\tiny{34.52}} & \tiny{8.54} & \tiny{8.08} & \textbf{\tiny{11.80}} & \tiny{3.72} & \tiny{3.00} & \textbf{\tiny{4.88}} & \tiny{1.88}\\\tiny{defocus\_blur} & \tiny{55.46} & \textbf{\tiny{64.80}} & \tiny{9.34} & \tiny{46.76} & \textbf{\tiny{59.84}} & \tiny{13.08} & \tiny{31.96} & \textbf{\tiny{50.44}} & \tiny{18.48} & \tiny{21.18} & \textbf{\tiny{36.00}} & \tiny{14.82} & \tiny{14.46} & \textbf{\tiny{22.40}} & \tiny{7.94}\\\tiny{elastic\_transform} & \tiny{71.84} & \textbf{\tiny{74.28}} & \tiny{2.44} & \tiny{60.66} & \textbf{\tiny{63.94}} & \tiny{3.28} & \tiny{71.94} & \textbf{\tiny{74.20}} & \tiny{2.26} & \tiny{67.54} & \textbf{\tiny{71.34}} & \tiny{3.80} & \tiny{53.24} & \textbf{\tiny{60.34}} & \tiny{7.10}\\\tiny{fog} & \tiny{55.82} & \textbf{\tiny{64.80}} & \tiny{8.98} & \tiny{47.44} & \textbf{\tiny{58.12}} & \tiny{10.68} & \tiny{36.76} & \textbf{\tiny{49.70}} & \tiny{12.94} & \tiny{34.44} & \textbf{\tiny{46.88}} & \tiny{12.44} & \tiny{21.46} & \textbf{\tiny{32.94}} & \tiny{11.48}\\\tiny{frost} & \tiny{64.48} & \textbf{\tiny{65.92}} & \tiny{1.44} & \tiny{46.40} & \textbf{\tiny{50.12}} & \tiny{3.72} & \tiny{33.50} & \textbf{\tiny{37.72}} & \tiny{4.22} & \tiny{31.62} & \textbf{\tiny{37.02}} & \tiny{5.40} & \tiny{24.70} & \textbf{\tiny{29.38}} & \tiny{4.68}\\\tiny{gaussian\_blur} & \tiny{67.14} & \textbf{\tiny{72.62}} & \tiny{5.48} & \tiny{51.02} & \textbf{\tiny{62.86}} & \tiny{11.84} & \tiny{36.62} & \textbf{\tiny{54.76}} & \tiny{18.14} & \tiny{25.40} & \textbf{\tiny{44.08}} & \tiny{18.68} & \tiny{12.96} & \textbf{\tiny{17.42}} & \tiny{4.46}\\\tiny{gaussian\_noise} & \tiny{59.12} & \textbf{\tiny{64.76}} & \tiny{5.64} & \tiny{34.46} & \textbf{\tiny{47.50}} & \tiny{13.04} & \tiny{10.22} & \textbf{\tiny{25.00}} & \tiny{14.78} & \tiny{2.34} & \textbf{\tiny{10.26}} & \tiny{7.92} & \tiny{1.02} & \textbf{\tiny{3.84}} & \tiny{2.82}\\\tiny{glass\_blur} & \tiny{61.28} & \textbf{\tiny{67.40}} & \tiny{6.12} & \tiny{51.34} & \textbf{\tiny{57.62}} & \tiny{6.28} & \tiny{36.56} & \textbf{\tiny{40.58}} & \tiny{4.02} & \tiny{30.02} & \textbf{\tiny{35.02}} & \tiny{5.00} & \tiny{22.16} & \textbf{\tiny{27.18}} & \tiny{5.02}\\\tiny{impulse\_noise} & \tiny{44.78} & \textbf{\tiny{54.86}} & \tiny{10.08} & \tiny{19.76} & \textbf{\tiny{35.04}} & \tiny{15.28} & \tiny{8.08} & \textbf{\tiny{23.12}} & \tiny{15.04} & \tiny{1.92} & \textbf{\tiny{8.70}} & \tiny{6.78} & \tiny{1.16} & \textbf{\tiny{3.82}} & \tiny{2.66}\\\tiny{jpeg\_compression} & \tiny{67.14} & \textbf{\tiny{69.06}} & \tiny{1.92} & \tiny{63.42} & \textbf{\tiny{65.24}} & \tiny{1.82} & \tiny{60.14} & \textbf{\tiny{61.74}} & \tiny{1.60} & \tiny{49.52} & \textbf{\tiny{52.16}} & \tiny{2.64} & \tiny{37.26} & \textbf{\tiny{40.86}} & \tiny{3.60}\\\tiny{motion\_blur} & \tiny{66.16} & \textbf{\tiny{72.04}} & \tiny{5.88} & \tiny{55.62} & \textbf{\tiny{63.78}} & \tiny{8.16} & \tiny{43.54} & \textbf{\tiny{50.62}} & \tiny{7.08} & \tiny{31.78} & \textbf{\tiny{36.04}} & \tiny{4.26} & \tiny{25.40} & \textbf{\tiny{29.44}} & \tiny{4.04}\\\tiny{pixelate} & \tiny{73.16} & \textbf{\tiny{76.48}} & \tiny{3.32} & \tiny{72.08} & \textbf{\tiny{76.66}} & \tiny{4.58} & \tiny{65.16} & \textbf{\tiny{72.28}} & \tiny{7.12} & \tiny{56.00} & \textbf{\tiny{65.38}} & \tiny{9.38} & \tiny{51.14} & \textbf{\tiny{60.14}} & \tiny{9.00}\\\tiny{saturate} & \tiny{62.00} & \textbf{\tiny{63.80}} & \tiny{1.80} & \tiny{48.16} & \textbf{\tiny{49.34}} & \tiny{1.18} & \tiny{69.68} & \textbf{\tiny{72.90}} & \tiny{3.22} & \tiny{43.26} & \textbf{\tiny{47.48}} & \tiny{4.22} & \tiny{26.76} & \textbf{\tiny{30.56}} & \tiny{3.80}\\\tiny{shot\_noise} & \tiny{57.84} & \textbf{\tiny{64.02}} & \tiny{6.18} & \tiny{31.70} & \textbf{\tiny{44.94}} & \tiny{13.24} & \tiny{11.72} & \textbf{\tiny{26.50}} & \tiny{14.78} & \tiny{2.60} & \textbf{\tiny{10.42}} & \tiny{7.82} & \tiny{1.58} & \textbf{\tiny{5.32}} & \tiny{3.74}\\\tiny{snow} & \tiny{56.42} & \textbf{\tiny{60.88}} & \tiny{4.46} & \tiny{33.52} & \textbf{\tiny{40.44}} & \tiny{6.92} & \tiny{38.68} & \textbf{\tiny{41.56}} & \tiny{2.88} & \tiny{26.24} & \textbf{\tiny{27.98}} & \tiny{1.74} & \tiny{19.70} & \textbf{\tiny{20.52}} & \tiny{0.82}\\\tiny{spatter} & \tiny{75.70} & \textbf{\tiny{77.50}} & \tiny{1.80} & \tiny{65.34} & \textbf{\tiny{68.38}} & \tiny{3.04} & \tiny{51.44} & \textbf{\tiny{57.02}} & \tiny{5.58} & \tiny{39.58} & \textbf{\tiny{45.38}} & \tiny{5.80} & \tiny{27.58} & \textbf{\tiny{35.62}} & \tiny{8.04}\\\tiny{speckle\_noise} & \tiny{62.88} & \textbf{\tiny{68.10}} & \tiny{5.22} & \tiny{51.86} & \textbf{\tiny{60.58}} & \tiny{8.72} & \tiny{21.74} & \textbf{\tiny{35.46}} & \tiny{13.72} & \tiny{11.66} & \textbf{\tiny{23.98}} & \tiny{12.32} & \tiny{6.26} & \textbf{\tiny{14.52}} & \tiny{8.26}\\\tiny{zoom\_blur} & \tiny{60.28} & \textbf{\tiny{66.90}} & \tiny{6.62} & \tiny{54.72} & \textbf{\tiny{61.06}} & \tiny{6.34} & \tiny{48.00} & \textbf{\tiny{56.90}} & \tiny{8.90} & \tiny{43.82} & \textbf{\tiny{52.38}} & \tiny{8.56} & \tiny{38.50} & \textbf{\tiny{46.80}} & \tiny{8.30}\\\tiny{{{$\Sigma$}}} & \tiny{62.91} & \textbf{\tiny{67.99}} & \tiny{5.08} & \tiny{50.09} & \textbf{\tiny{57.65}} & \tiny{7.55} & \tiny{40.50} & \textbf{\tiny{49.24}} & \tiny{8.73} & \tiny{30.88} & \textbf{\tiny{38.18}} & \tiny{7.30} & \tiny{22.97} & \textbf{\tiny{28.35}} & \tiny{5.38}\end{tabular}\caption{Accuracies for DenseNet w/wo OpticsAugment evaluated on ImageNet-100-c 2D common corruptions~\cite{hendrycks_benchmarking_2019}.}\label{tab:tab:imagenet100c_corruptions_DenseNet}\end{table*}
\begin{table*}[h]\centering\begin{tabular}{llll|lll|lll|lll|lll}&\multicolumn{3}{c}{\scriptsize{1}}&\multicolumn{3}{c}{\scriptsize{2}}&\multicolumn{3}{c}{\scriptsize{3}}&\multicolumn{3}{c}{\scriptsize{4}}&\multicolumn{3}{c}{\scriptsize{5}}\\\tiny{Corruption} & \tiny{clean} & \tiny{\textbf{ours}} & \tiny{$\Delta$} & \tiny{clean} & \tiny{\textbf{ours}} & \tiny{$\Delta$} & \tiny{clean} & \tiny{\textbf{ours}} & \tiny{$\Delta$} & \tiny{clean} & \tiny{\textbf{ours}} & \tiny{$\Delta$} & \tiny{clean} & \tiny{\textbf{ours}} & \tiny{$\Delta$}\\\hline\tiny{brightness} & \tiny{74.12} & \textbf{\tiny{74.56}} & \tiny{0.44} & \tiny{70.66} & \textbf{\tiny{72.08}} & \tiny{1.42} & \tiny{64.90} & \textbf{\tiny{68.56}} & \tiny{3.66} & \tiny{57.76} & \textbf{\tiny{63.32}} & \tiny{5.56} & \tiny{46.82} & \textbf{\tiny{54.60}} & \tiny{7.78}\\\tiny{contrast} & \tiny{56.02} & \textbf{\tiny{60.46}} & \tiny{4.44} & \tiny{41.50} & \textbf{\tiny{49.32}} & \tiny{7.82} & \tiny{20.74} & \textbf{\tiny{29.04}} & \tiny{8.30} & \tiny{5.86} & \textbf{\tiny{7.14}} & \tiny{1.28} & \textbf{\tiny{3.26}} & \tiny{2.90} & \tiny{-0.36}\\\tiny{defocus\_blur} & \tiny{55.46} & \textbf{\tiny{60.38}} & \tiny{4.92} & \tiny{46.42} & \textbf{\tiny{53.72}} & \tiny{7.30} & \tiny{32.20} & \textbf{\tiny{39.02}} & \tiny{6.82} & \tiny{21.94} & \textbf{\tiny{25.98}} & \tiny{4.04} & \tiny{13.92} & \textbf{\tiny{16.64}} & \tiny{2.72}\\\tiny{elastic\_transform} & \tiny{69.78} & \textbf{\tiny{69.94}} & \tiny{0.16} & \textbf{\tiny{59.82}} & \tiny{57.96} & \tiny{-1.86} & \textbf{\tiny{69.68}} & \tiny{69.48} & \tiny{-0.20} & \tiny{66.26} & \textbf{\tiny{67.14}} & \tiny{0.88} & \tiny{56.10} & \textbf{\tiny{56.50}} & \tiny{0.40}\\\tiny{fog} & \tiny{53.38} & \textbf{\tiny{60.14}} & \tiny{6.76} & \tiny{43.82} & \textbf{\tiny{53.16}} & \tiny{9.34} & \tiny{33.48} & \textbf{\tiny{45.06}} & \tiny{11.58} & \tiny{30.60} & \textbf{\tiny{40.38}} & \tiny{9.78} & \tiny{21.10} & \textbf{\tiny{28.02}} & \tiny{6.92}\\\tiny{frost} & \tiny{59.98} & \textbf{\tiny{64.04}} & \tiny{4.06} & \tiny{42.96} & \textbf{\tiny{47.40}} & \tiny{4.44} & \tiny{32.08} & \textbf{\tiny{36.56}} & \tiny{4.48} & \tiny{30.80} & \textbf{\tiny{35.72}} & \tiny{4.92} & \tiny{25.16} & \textbf{\tiny{28.08}} & \tiny{2.92}\\\tiny{gaussian\_blur} & \tiny{66.44} & \textbf{\tiny{68.24}} & \tiny{1.80} & \tiny{50.66} & \textbf{\tiny{56.92}} & \tiny{6.26} & \tiny{36.06} & \textbf{\tiny{43.50}} & \tiny{7.44} & \tiny{24.90} & \textbf{\tiny{29.60}} & \tiny{4.70} & \tiny{12.30} & \textbf{\tiny{13.32}} & \tiny{1.02}\\\tiny{gaussian\_noise} & \tiny{52.06} & \textbf{\tiny{61.34}} & \tiny{9.28} & \tiny{28.36} & \textbf{\tiny{45.36}} & \tiny{17.00} & \tiny{9.32} & \textbf{\tiny{24.46}} & \tiny{15.14} & \tiny{3.72} & \textbf{\tiny{9.34}} & \tiny{5.62} & \tiny{1.60} & \textbf{\tiny{3.56}} & \tiny{1.96}\\\tiny{glass\_blur} & \tiny{58.52} & \textbf{\tiny{61.02}} & \tiny{2.50} & \tiny{49.28} & \textbf{\tiny{51.16}} & \tiny{1.88} & \tiny{33.32} & \textbf{\tiny{35.32}} & \tiny{2.00} & \tiny{27.58} & \textbf{\tiny{29.56}} & \tiny{1.98} & \textbf{\tiny{21.28}} & \tiny{21.04} & \tiny{-0.24}\\\tiny{impulse\_noise} & \tiny{40.72} & \textbf{\tiny{54.50}} & \tiny{13.78} & \tiny{19.78} & \textbf{\tiny{38.22}} & \tiny{18.44} & \tiny{10.22} & \textbf{\tiny{24.56}} & \tiny{14.34} & \tiny{3.22} & \textbf{\tiny{8.90}} & \tiny{5.68} & \tiny{1.50} & \textbf{\tiny{3.46}} & \tiny{1.96}\\\tiny{jpeg\_compression} & \tiny{63.62} & \textbf{\tiny{65.04}} & \tiny{1.42} & \tiny{58.86} & \textbf{\tiny{62.00}} & \tiny{3.14} & \tiny{55.20} & \textbf{\tiny{58.68}} & \tiny{3.48} & \tiny{44.22} & \textbf{\tiny{49.98}} & \tiny{5.76} & \tiny{32.70} & \textbf{\tiny{39.30}} & \tiny{6.60}\\\tiny{motion\_blur} & \tiny{66.20} & \textbf{\tiny{67.96}} & \tiny{1.76} & \tiny{56.70} & \textbf{\tiny{61.18}} & \tiny{4.48} & \tiny{44.10} & \textbf{\tiny{50.54}} & \tiny{6.44} & \tiny{32.56} & \textbf{\tiny{36.22}} & \tiny{3.66} & \tiny{24.86} & \textbf{\tiny{27.84}} & \tiny{2.98}\\\tiny{pixelate} & \tiny{69.58} & \textbf{\tiny{71.36}} & \tiny{1.78} & \tiny{69.76} & \textbf{\tiny{70.80}} & \tiny{1.04} & \tiny{56.80} & \textbf{\tiny{65.20}} & \tiny{8.40} & \tiny{38.70} & \textbf{\tiny{57.46}} & \tiny{18.76} & \tiny{30.02} & \textbf{\tiny{51.78}} & \tiny{21.76}\\\tiny{saturate} & \tiny{59.04} & \textbf{\tiny{61.32}} & \tiny{2.28} & \tiny{41.98} & \textbf{\tiny{44.48}} & \tiny{2.50} & \tiny{69.42} & \textbf{\tiny{70.70}} & \tiny{1.28} & \tiny{47.96} & \textbf{\tiny{53.02}} & \tiny{5.06} & \tiny{30.04} & \textbf{\tiny{37.38}} & \tiny{7.34}\\\tiny{shot\_noise} & \tiny{49.16} & \textbf{\tiny{59.42}} & \tiny{10.26} & \tiny{25.72} & \textbf{\tiny{40.36}} & \tiny{14.64} & \tiny{10.70} & \textbf{\tiny{21.76}} & \tiny{11.06} & \tiny{4.02} & \textbf{\tiny{7.68}} & \tiny{3.66} & \tiny{2.22} & \textbf{\tiny{4.30}} & \tiny{2.08}\\\tiny{snow} & \tiny{55.42} & \textbf{\tiny{60.36}} & \tiny{4.94} & \tiny{36.30} & \textbf{\tiny{40.02}} & \tiny{3.72} & \tiny{39.88} & \textbf{\tiny{42.92}} & \tiny{3.04} & \textbf{\tiny{30.60}} & \tiny{29.66} & \tiny{-0.94} & \tiny{21.20} & \textbf{\tiny{21.24}} & \tiny{0.04}\\\tiny{spatter} & \tiny{73.18} & \textbf{\tiny{73.66}} & \tiny{0.48} & \tiny{62.36} & \textbf{\tiny{64.72}} & \tiny{2.36} & \tiny{50.00} & \textbf{\tiny{55.48}} & \tiny{5.48} & \textbf{\tiny{49.56}} & \tiny{48.16} & \tiny{-1.40} & \textbf{\tiny{39.02}} & \tiny{37.28} & \tiny{-1.74}\\\tiny{speckle\_noise} & \tiny{53.24} & \textbf{\tiny{62.96}} & \tiny{9.72} & \tiny{41.66} & \textbf{\tiny{54.84}} & \tiny{13.18} & \tiny{18.88} & \textbf{\tiny{30.00}} & \tiny{11.12} & \tiny{12.04} & \textbf{\tiny{17.80}} & \tiny{5.76} & \tiny{7.06} & \textbf{\tiny{9.60}} & \tiny{2.54}\\\tiny{zoom\_blur} & \tiny{55.32} & \textbf{\tiny{57.28}} & \tiny{1.96} & \tiny{47.18} & \textbf{\tiny{50.08}} & \tiny{2.90} & \tiny{43.30} & \textbf{\tiny{46.82}} & \tiny{3.52} & \tiny{38.44} & \textbf{\tiny{41.40}} & \tiny{2.96} & \tiny{34.10} & \textbf{\tiny{37.60}} & \tiny{3.50}\\\tiny{{{$\Sigma$}}} & \tiny{59.54} & \textbf{\tiny{63.89}} & \tiny{4.35} & \tiny{47.04} & \textbf{\tiny{53.36}} & \tiny{6.32} & \tiny{38.44} & \textbf{\tiny{45.14}} & \tiny{6.70} & \tiny{30.04} & \textbf{\tiny{34.66}} & \tiny{4.62} & \tiny{22.33} & \textbf{\tiny{26.02}} & \tiny{3.69}\end{tabular}\caption{Accuracies for EfficientNet w/wo OpticsAugment evaluated on ImageNet-100-c 2D common corruptions~\cite{hendrycks_benchmarking_2019}.}\label{tab:tab:imagenet100c_corruptions_EfficientNet}\end{table*}
\begin{table*}[h]\centering\begin{tabular}{llll|lll|lll|lll|lll}&\multicolumn{3}{c}{\scriptsize{1}}&\multicolumn{3}{c}{\scriptsize{2}}&\multicolumn{3}{c}{\scriptsize{3}}&\multicolumn{3}{c}{\scriptsize{4}}&\multicolumn{3}{c}{\scriptsize{5}}\\\tiny{Corruption} & \tiny{clean} & \tiny{\textbf{ours}} & \tiny{$\Delta$} & \tiny{clean} & \tiny{\textbf{ours}} & \tiny{$\Delta$} & \tiny{clean} & \tiny{\textbf{ours}} & \tiny{$\Delta$} & \tiny{clean} & \tiny{\textbf{ours}} & \tiny{$\Delta$} & \tiny{clean} & \tiny{\textbf{ours}} & \tiny{$\Delta$}\\\hline\tiny{brightness} & \tiny{71.08} & \textbf{\tiny{72.62}} & \tiny{1.54} & \tiny{67.70} & \textbf{\tiny{70.96}} & \tiny{3.26} & \tiny{63.58} & \textbf{\tiny{67.18}} & \tiny{3.60} & \tiny{58.14} & \textbf{\tiny{61.22}} & \tiny{3.08} & \tiny{49.98} & \textbf{\tiny{52.98}} & \tiny{3.00}\\\tiny{contrast} & \tiny{48.70} & \textbf{\tiny{58.90}} & \tiny{10.20} & \tiny{32.92} & \textbf{\tiny{47.06}} & \tiny{14.14} & \tiny{14.88} & \textbf{\tiny{27.10}} & \tiny{12.22} & \tiny{3.78} & \textbf{\tiny{7.10}} & \tiny{3.32} & \tiny{1.84} & \textbf{\tiny{3.10}} & \tiny{1.26}\\\tiny{defocus\_blur} & \tiny{51.96} & \textbf{\tiny{57.96}} & \tiny{6.00} & \tiny{42.92} & \textbf{\tiny{52.80}} & \tiny{9.88} & \tiny{26.34} & \textbf{\tiny{36.26}} & \tiny{9.92} & \tiny{17.36} & \textbf{\tiny{24.10}} & \tiny{6.74} & \tiny{11.94} & \textbf{\tiny{16.18}} & \tiny{4.24}\\\tiny{elastic\_transform} & \tiny{66.50} & \textbf{\tiny{67.52}} & \tiny{1.02} & \tiny{54.58} & \textbf{\tiny{54.68}} & \tiny{0.10} & \tiny{65.56} & \textbf{\tiny{67.52}} & \tiny{1.96} & \tiny{61.90} & \textbf{\tiny{65.52}} & \tiny{3.62} & \tiny{51.68} & \textbf{\tiny{56.60}} & \tiny{4.92}\\\tiny{fog} & \tiny{47.88} & \textbf{\tiny{56.72}} & \tiny{8.84} & \tiny{38.46} & \textbf{\tiny{49.52}} & \tiny{11.06} & \tiny{29.20} & \textbf{\tiny{40.74}} & \tiny{11.54} & \tiny{28.14} & \textbf{\tiny{36.40}} & \tiny{8.26} & \tiny{19.34} & \textbf{\tiny{23.64}} & \tiny{4.30}\\\tiny{frost} & \tiny{56.04} & \textbf{\tiny{58.88}} & \tiny{2.84} & \tiny{39.88} & \textbf{\tiny{41.00}} & \tiny{1.12} & \tiny{27.54} & \textbf{\tiny{28.80}} & \tiny{1.26} & \tiny{26.80} & \textbf{\tiny{27.32}} & \tiny{0.52} & \tiny{20.18} & \textbf{\tiny{21.48}} & \tiny{1.30}\\\tiny{gaussian\_blur} & \tiny{62.74} & \textbf{\tiny{64.66}} & \tiny{1.92} & \tiny{46.44} & \textbf{\tiny{54.92}} & \tiny{8.48} & \tiny{30.66} & \textbf{\tiny{39.44}} & \tiny{8.78} & \tiny{20.00} & \textbf{\tiny{25.70}} & \tiny{5.70} & \tiny{10.62} & \textbf{\tiny{12.42}} & \tiny{1.80}\\\tiny{gaussian\_noise} & \tiny{53.80} & \textbf{\tiny{55.92}} & \tiny{2.12} & \tiny{35.02} & \textbf{\tiny{39.78}} & \tiny{4.76} & \tiny{17.20} & \textbf{\tiny{21.26}} & \tiny{4.06} & \tiny{6.76} & \textbf{\tiny{9.42}} & \tiny{2.66} & \tiny{2.56} & \textbf{\tiny{3.84}} & \tiny{1.28}\\\tiny{glass\_blur} & \tiny{57.68} & \textbf{\tiny{60.14}} & \tiny{2.46} & \tiny{49.16} & \textbf{\tiny{50.70}} & \tiny{1.54} & \tiny{35.46} & \textbf{\tiny{35.54}} & \tiny{0.08} & \tiny{28.58} & \textbf{\tiny{31.26}} & \tiny{2.68} & \tiny{19.68} & \textbf{\tiny{24.00}} & \tiny{4.32}\\\tiny{impulse\_noise} & \tiny{46.78} & \textbf{\tiny{51.76}} & \tiny{4.98} & \tiny{27.10} & \textbf{\tiny{33.56}} & \tiny{6.46} & \tiny{16.80} & \textbf{\tiny{21.50}} & \tiny{4.70} & \tiny{5.78} & \textbf{\tiny{8.48}} & \tiny{2.70} & \tiny{2.54} & \textbf{\tiny{3.50}} & \tiny{0.96}\\\tiny{jpeg\_compression} & \tiny{61.20} & \textbf{\tiny{62.42}} & \tiny{1.22} & \tiny{56.62} & \textbf{\tiny{58.72}} & \tiny{2.10} & \tiny{53.94} & \textbf{\tiny{55.38}} & \tiny{1.44} & \tiny{45.40} & \textbf{\tiny{47.18}} & \tiny{1.78} & \tiny{34.98} & \textbf{\tiny{37.48}} & \tiny{2.50}\\\tiny{motion\_blur} & \tiny{61.50} & \textbf{\tiny{65.74}} & \tiny{4.24} & \tiny{53.38} & \textbf{\tiny{58.16}} & \tiny{4.78} & \tiny{40.06} & \textbf{\tiny{44.98}} & \tiny{4.92} & \tiny{27.80} & \textbf{\tiny{31.96}} & \tiny{4.16} & \tiny{20.94} & \textbf{\tiny{25.16}} & \tiny{4.22}\\\tiny{pixelate} & \tiny{69.04} & \textbf{\tiny{69.12}} & \tiny{0.08} & \tiny{67.76} & \textbf{\tiny{68.40}} & \tiny{0.64} & \tiny{61.80} & \textbf{\tiny{63.96}} & \tiny{2.16} & \tiny{51.62} & \textbf{\tiny{58.12}} & \tiny{6.50} & \tiny{44.00} & \textbf{\tiny{53.60}} & \tiny{9.60}\\\tiny{saturate} & \tiny{55.58} & \textbf{\tiny{57.64}} & \tiny{2.06} & \tiny{40.18} & \textbf{\tiny{41.92}} & \tiny{1.74} & \tiny{67.94} & \textbf{\tiny{68.48}} & \tiny{0.54} & \textbf{\tiny{48.46}} & \tiny{48.46} & \tiny{0.00} & \textbf{\tiny{33.88}} & \tiny{33.56} & \tiny{-0.32}\\\tiny{shot\_noise} & \tiny{49.34} & \textbf{\tiny{54.30}} & \tiny{4.96} & \tiny{31.02} & \textbf{\tiny{36.80}} & \tiny{5.78} & \tiny{17.12} & \textbf{\tiny{22.12}} & \tiny{5.00} & \tiny{6.86} & \textbf{\tiny{9.04}} & \tiny{2.18} & \tiny{3.76} & \textbf{\tiny{5.08}} & \tiny{1.32}\\\tiny{snow} & \tiny{52.78} & \textbf{\tiny{57.84}} & \tiny{5.06} & \tiny{31.14} & \textbf{\tiny{37.48}} & \tiny{6.34} & \tiny{35.48} & \textbf{\tiny{39.78}} & \tiny{4.30} & \tiny{24.74} & \textbf{\tiny{29.14}} & \tiny{4.40} & \tiny{16.48} & \textbf{\tiny{20.14}} & \tiny{3.66}\\\tiny{spatter} & \tiny{71.16} & \textbf{\tiny{71.58}} & \tiny{0.42} & \tiny{60.28} & \textbf{\tiny{62.98}} & \tiny{2.70} & \tiny{49.54} & \textbf{\tiny{53.06}} & \tiny{3.52} & \tiny{45.52} & \textbf{\tiny{48.18}} & \tiny{2.66} & \tiny{33.70} & \textbf{\tiny{35.50}} & \tiny{1.80}\\\tiny{speckle\_noise} & \tiny{53.72} & \textbf{\tiny{57.84}} & \tiny{4.12} & \tiny{43.70} & \textbf{\tiny{49.98}} & \tiny{6.28} & \tiny{22.90} & \textbf{\tiny{29.62}} & \tiny{6.72} & \tiny{15.64} & \textbf{\tiny{20.94}} & \tiny{5.30} & \tiny{9.88} & \textbf{\tiny{13.82}} & \tiny{3.94}\\\tiny{zoom\_blur} & \tiny{50.98} & \textbf{\tiny{54.98}} & \tiny{4.00} & \tiny{44.88} & \textbf{\tiny{47.18}} & \tiny{2.30} & \tiny{39.48} & \textbf{\tiny{43.54}} & \tiny{4.06} & \tiny{34.86} & \textbf{\tiny{38.76}} & \tiny{3.90} & \tiny{30.56} & \textbf{\tiny{34.76}} & \tiny{4.20}\\\tiny{{{$\Sigma$}}} & \tiny{57.29} & \textbf{\tiny{60.87}} & \tiny{3.58} & \tiny{45.43} & \textbf{\tiny{50.35}} & \tiny{4.92} & \tiny{37.66} & \textbf{\tiny{42.43}} & \tiny{4.78} & \tiny{29.38} & \textbf{\tiny{33.07}} & \tiny{3.69} & \tiny{22.03} & \textbf{\tiny{25.10}} & \tiny{3.07}\end{tabular}\caption{Accuracies for MobileNet w/wo OpticsAugment evaluated on ImageNet-100-c 2D common corruptions~\cite{hendrycks_benchmarking_2019}.}\label{tab:tab:imagenet100c_corruptions_MobileNet}\end{table*}
\begin{table*}[h]\centering\begin{tabular}{llll|lll|lll|lll|lll}&\multicolumn{3}{c}{\scriptsize{1}}&\multicolumn{3}{c}{\scriptsize{2}}&\multicolumn{3}{c}{\scriptsize{3}}&\multicolumn{3}{c}{\scriptsize{4}}&\multicolumn{3}{c}{\scriptsize{5}}\\\tiny{Corruption} & \tiny{clean} & \tiny{\textbf{ours}} & \tiny{$\Delta$} & \tiny{clean} & \tiny{\textbf{ours}} & \tiny{$\Delta$} & \tiny{clean} & \tiny{\textbf{ours}} & \tiny{$\Delta$} & \tiny{clean} & \tiny{\textbf{ours}} & \tiny{$\Delta$} & \tiny{clean} & \tiny{\textbf{ours}} & \tiny{$\Delta$}\\\hline\tiny{brightness} & \textbf{\tiny{80.00}} & \tiny{78.96} & \tiny{-1.04} & \textbf{\tiny{77.58}} & \tiny{76.20} & \tiny{-1.38} & \textbf{\tiny{74.30}} & \tiny{71.74} & \tiny{-2.56} & \textbf{\tiny{68.16}} & \tiny{65.16} & \tiny{-3.00} & \textbf{\tiny{58.74}} & \tiny{55.76} & \tiny{-2.98}\\\tiny{contrast} & \tiny{63.62} & \textbf{\tiny{64.56}} & \tiny{0.94} & \tiny{52.90} & \textbf{\tiny{53.70}} & \tiny{0.80} & \tiny{30.66} & \textbf{\tiny{31.94}} & \tiny{1.28} & \tiny{8.56} & \textbf{\tiny{8.58}} & \tiny{0.02} & \tiny{3.02} & \textbf{\tiny{3.26}} & \tiny{0.24}\\\tiny{defocus\_blur} & \tiny{60.30} & \textbf{\tiny{67.52}} & \tiny{7.22} & \tiny{51.82} & \textbf{\tiny{64.18}} & \tiny{12.36} & \tiny{39.78} & \textbf{\tiny{57.46}} & \tiny{17.68} & \tiny{30.04} & \textbf{\tiny{44.38}} & \tiny{14.34} & \tiny{22.42} & \textbf{\tiny{30.80}} & \tiny{8.38}\\\tiny{elastic\_transform} & \tiny{74.44} & \textbf{\tiny{74.46}} & \tiny{0.02} & \tiny{62.92} & \textbf{\tiny{64.22}} & \tiny{1.30} & \tiny{74.34} & \textbf{\tiny{75.20}} & \tiny{0.86} & \tiny{71.62} & \textbf{\tiny{73.02}} & \tiny{1.40} & \tiny{58.74} & \textbf{\tiny{65.06}} & \tiny{6.32}\\\tiny{fog} & \textbf{\tiny{63.14}} & \tiny{62.56} & \tiny{-0.58} & \tiny{54.54} & \textbf{\tiny{54.96}} & \tiny{0.42} & \tiny{44.88} & \textbf{\tiny{45.98}} & \tiny{1.10} & \tiny{41.04} & \textbf{\tiny{44.32}} & \tiny{3.28} & \tiny{28.24} & \textbf{\tiny{32.60}} & \tiny{4.36}\\\tiny{frost} & \textbf{\tiny{68.52}} & \tiny{67.66} & \tiny{-0.86} & \tiny{51.46} & \textbf{\tiny{54.32}} & \tiny{2.86} & \tiny{37.62} & \textbf{\tiny{42.86}} & \tiny{5.24} & \tiny{36.04} & \textbf{\tiny{41.76}} & \tiny{5.72} & \tiny{29.50} & \textbf{\tiny{35.52}} & \tiny{6.02}\\\tiny{gaussian\_blur} & \tiny{71.12} & \textbf{\tiny{73.72}} & \tiny{2.60} & \tiny{55.70} & \textbf{\tiny{65.48}} & \tiny{9.78} & \tiny{42.64} & \textbf{\tiny{58.90}} & \tiny{16.26} & \tiny{33.54} & \textbf{\tiny{50.34}} & \tiny{16.80} & \tiny{20.96} & \textbf{\tiny{24.04}} & \tiny{3.08}\\\tiny{gaussian\_noise} & \textbf{\tiny{67.48}} & \tiny{66.42} & \tiny{-1.06} & \tiny{51.34} & \textbf{\tiny{51.88}} & \tiny{0.54} & \tiny{29.38} & \textbf{\tiny{31.02}} & \tiny{1.64} & \tiny{12.72} & \textbf{\tiny{13.90}} & \tiny{1.18} & \tiny{4.20} & \textbf{\tiny{5.36}} & \tiny{1.16}\\\tiny{glass\_blur} & \tiny{64.54} & \textbf{\tiny{67.68}} & \tiny{3.14} & \tiny{54.66} & \textbf{\tiny{60.58}} & \tiny{5.92} & \tiny{40.18} & \textbf{\tiny{47.34}} & \tiny{7.16} & \tiny{33.90} & \textbf{\tiny{41.76}} & \tiny{7.86} & \tiny{26.98} & \textbf{\tiny{34.60}} & \tiny{7.62}\\\tiny{impulse\_noise} & \tiny{55.94} & \textbf{\tiny{57.74}} & \tiny{1.80} & \tiny{38.50} & \textbf{\tiny{41.00}} & \tiny{2.50} & \tiny{25.28} & \textbf{\tiny{28.64}} & \tiny{3.36} & \tiny{9.54} & \textbf{\tiny{12.74}} & \tiny{3.20} & \tiny{4.06} & \textbf{\tiny{5.56}} & \tiny{1.50}\\\tiny{jpeg\_compression} & \textbf{\tiny{70.52}} & \tiny{70.26} & \tiny{-0.26} & \tiny{66.12} & \textbf{\tiny{66.98}} & \tiny{0.86} & \tiny{62.76} & \textbf{\tiny{64.62}} & \tiny{1.86} & \tiny{50.34} & \textbf{\tiny{56.14}} & \tiny{5.80} & \tiny{37.76} & \textbf{\tiny{45.18}} & \tiny{7.42}\\\tiny{motion\_blur} & \tiny{69.78} & \textbf{\tiny{73.82}} & \tiny{4.04} & \tiny{60.54} & \textbf{\tiny{69.84}} & \tiny{9.30} & \tiny{48.90} & \textbf{\tiny{61.24}} & \tiny{12.34} & \tiny{36.26} & \textbf{\tiny{47.78}} & \tiny{11.52} & \tiny{29.52} & \textbf{\tiny{37.44}} & \tiny{7.92}\\\tiny{pixelate} & \tiny{76.08} & \textbf{\tiny{78.14}} & \tiny{2.06} & \tiny{74.62} & \textbf{\tiny{78.06}} & \tiny{3.44} & \tiny{68.12} & \textbf{\tiny{74.96}} & \tiny{6.84} & \tiny{59.92} & \textbf{\tiny{70.80}} & \tiny{10.88} & \tiny{55.34} & \textbf{\tiny{67.84}} & \tiny{12.50}\\\tiny{saturate} & \textbf{\tiny{66.46}} & \tiny{66.30} & \tiny{-0.16} & \textbf{\tiny{53.22}} & \tiny{52.04} & \tiny{-1.18} & \textbf{\tiny{75.32}} & \tiny{74.04} & \tiny{-1.28} & \textbf{\tiny{57.16}} & \tiny{53.92} & \tiny{-3.24} & \textbf{\tiny{40.72}} & \tiny{36.40} & \tiny{-4.32}\\\tiny{shot\_noise} & \tiny{64.52} & \textbf{\tiny{65.10}} & \tiny{0.58} & \textbf{\tiny{49.62}} & \tiny{49.52} & \tiny{-0.10} & \tiny{30.86} & \textbf{\tiny{32.50}} & \tiny{1.64} & \tiny{12.30} & \textbf{\tiny{14.20}} & \tiny{1.90} & \tiny{6.40} & \textbf{\tiny{7.84}} & \tiny{1.44}\\\tiny{snow} & \tiny{64.30} & \textbf{\tiny{65.32}} & \tiny{1.02} & \tiny{43.38} & \textbf{\tiny{47.36}} & \tiny{3.98} & \tiny{46.30} & \textbf{\tiny{47.28}} & \tiny{0.98} & \tiny{32.76} & \textbf{\tiny{34.08}} & \tiny{1.32} & \tiny{25.06} & \textbf{\tiny{27.30}} & \tiny{2.24}\\\tiny{spatter} & \textbf{\tiny{79.26}} & \tiny{78.28} & \tiny{-0.98} & \textbf{\tiny{69.94}} & \tiny{69.54} & \tiny{-0.40} & \tiny{59.40} & \textbf{\tiny{61.02}} & \tiny{1.62} & \tiny{51.30} & \textbf{\tiny{55.94}} & \tiny{4.64} & \tiny{40.66} & \textbf{\tiny{45.38}} & \tiny{4.72}\\\tiny{speckle\_noise} & \tiny{68.14} & \textbf{\tiny{68.18}} & \tiny{0.04} & \tiny{60.60} & \textbf{\tiny{61.78}} & \tiny{1.18} & \tiny{38.60} & \textbf{\tiny{40.72}} & \tiny{2.12} & \tiny{27.88} & \textbf{\tiny{29.62}} & \tiny{1.74} & \tiny{17.62} & \textbf{\tiny{19.72}} & \tiny{2.10}\\\tiny{zoom\_blur} & \tiny{61.66} & \textbf{\tiny{66.86}} & \tiny{5.20} & \tiny{54.94} & \textbf{\tiny{61.14}} & \tiny{6.20} & \tiny{49.94} & \textbf{\tiny{58.18}} & \tiny{8.24} & \tiny{45.12} & \textbf{\tiny{52.92}} & \tiny{7.80} & \tiny{40.08} & \textbf{\tiny{48.34}} & \tiny{8.26}\\\tiny{{{$\Sigma$}}} & \tiny{67.89} & \textbf{\tiny{69.13}} & \tiny{1.25} & \tiny{57.07} & \textbf{\tiny{60.15}} & \tiny{3.07} & \tiny{48.38} & \textbf{\tiny{52.93}} & \tiny{4.55} & \tiny{37.80} & \textbf{\tiny{42.70}} & \tiny{4.90} & \tiny{28.95} & \textbf{\tiny{33.05}} & \tiny{4.10}\end{tabular}\caption{Accuracies for ResNet101 w/wo OpticsAugment evaluated on ImageNet-100-c 2D common corruptions~\cite{hendrycks_benchmarking_2019}.}\label{tab:tab:imagenet100c_corruptions_ResNet101}\end{table*}
\begin{table*}[h]\centering\begin{tabular}{llll|lll|lll|lll|lll}&\multicolumn{3}{c}{\scriptsize{1}}&\multicolumn{3}{c}{\scriptsize{2}}&\multicolumn{3}{c}{\scriptsize{3}}&\multicolumn{3}{c}{\scriptsize{4}}&\multicolumn{3}{c}{\scriptsize{5}}\\\tiny{Corruption} & \tiny{clean} & \tiny{\textbf{ours}} & \tiny{$\Delta$} & \tiny{clean} & \tiny{\textbf{ours}} & \tiny{$\Delta$} & \tiny{clean} & \tiny{\textbf{ours}} & \tiny{$\Delta$} & \tiny{clean} & \tiny{\textbf{ours}} & \tiny{$\Delta$} & \tiny{clean} & \tiny{\textbf{ours}} & \tiny{$\Delta$}\\\hline\tiny{brightness} & \textbf{\tiny{73.64}} & \tiny{73.18} & \tiny{-0.46} & \textbf{\tiny{69.64}} & \tiny{68.50} & \tiny{-1.14} & \textbf{\tiny{64.04}} & \tiny{61.12} & \tiny{-2.92} & \textbf{\tiny{54.68}} & \tiny{50.54} & \tiny{-4.14} & \textbf{\tiny{42.40}} & \tiny{36.82} & \tiny{-5.58}\\\tiny{contrast} & \tiny{50.40} & \textbf{\tiny{56.58}} & \tiny{6.18} & \tiny{33.88} & \textbf{\tiny{42.60}} & \tiny{8.72} & \tiny{16.38} & \textbf{\tiny{21.86}} & \tiny{5.48} & \tiny{4.88} & \textbf{\tiny{6.78}} & \tiny{1.90} & \tiny{2.66} & \textbf{\tiny{3.10}} & \tiny{0.44}\\\tiny{defocus\_blur} & \tiny{48.84} & \textbf{\tiny{62.54}} & \tiny{13.70} & \tiny{40.30} & \textbf{\tiny{58.68}} & \tiny{18.38} & \tiny{25.28} & \textbf{\tiny{49.46}} & \tiny{24.18} & \tiny{16.46} & \textbf{\tiny{34.54}} & \tiny{18.08} & \tiny{11.68} & \textbf{\tiny{20.82}} & \tiny{9.14}\\\tiny{elastic\_transform} & \tiny{68.02} & \textbf{\tiny{71.26}} & \tiny{3.24} & \tiny{57.98} & \textbf{\tiny{60.30}} & \tiny{2.32} & \tiny{68.12} & \textbf{\tiny{70.98}} & \tiny{2.86} & \tiny{64.54} & \textbf{\tiny{68.80}} & \tiny{4.26} & \tiny{53.80} & \textbf{\tiny{60.92}} & \tiny{7.12}\\\tiny{fog} & \tiny{51.40} & \textbf{\tiny{55.58}} & \tiny{4.18} & \tiny{41.76} & \textbf{\tiny{46.80}} & \tiny{5.04} & \tiny{31.36} & \textbf{\tiny{38.26}} & \tiny{6.90} & \tiny{29.24} & \textbf{\tiny{35.52}} & \tiny{6.28} & \tiny{17.24} & \textbf{\tiny{23.56}} & \tiny{6.32}\\\tiny{frost} & \tiny{57.26} & \textbf{\tiny{58.62}} & \tiny{1.36} & \tiny{37.92} & \textbf{\tiny{40.86}} & \tiny{2.94} & \tiny{26.18} & \textbf{\tiny{28.94}} & \tiny{2.76} & \tiny{24.74} & \textbf{\tiny{29.24}} & \tiny{4.50} & \tiny{18.24} & \textbf{\tiny{22.80}} & \tiny{4.56}\\\tiny{gaussian\_blur} & \tiny{62.92} & \textbf{\tiny{69.40}} & \tiny{6.48} & \tiny{44.28} & \textbf{\tiny{60.64}} & \tiny{16.36} & \tiny{29.20} & \textbf{\tiny{52.02}} & \tiny{22.82} & \tiny{19.70} & \textbf{\tiny{41.48}} & \tiny{21.78} & \tiny{9.98} & \textbf{\tiny{17.06}} & \tiny{7.08}\\\tiny{gaussian\_noise} & \tiny{52.80} & \textbf{\tiny{60.16}} & \tiny{7.36} & \tiny{29.98} & \textbf{\tiny{42.82}} & \tiny{12.84} & \tiny{9.12} & \textbf{\tiny{22.44}} & \tiny{13.32} & \tiny{2.74} & \textbf{\tiny{8.90}} & \tiny{6.16} & \tiny{1.34} & \textbf{\tiny{3.38}} & \tiny{2.04}\\\tiny{glass\_blur} & \tiny{56.02} & \textbf{\tiny{65.42}} & \tiny{9.40} & \tiny{45.18} & \textbf{\tiny{58.72}} & \tiny{13.54} & \tiny{31.06} & \textbf{\tiny{46.34}} & \tiny{15.28} & \tiny{25.14} & \textbf{\tiny{40.94}} & \tiny{15.80} & \tiny{18.56} & \textbf{\tiny{32.40}} & \tiny{13.84}\\\tiny{impulse\_noise} & \tiny{39.94} & \textbf{\tiny{50.30}} & \tiny{10.36} & \tiny{16.52} & \textbf{\tiny{31.00}} & \tiny{14.48} & \tiny{8.30} & \textbf{\tiny{19.24}} & \tiny{10.94} & \tiny{2.18} & \textbf{\tiny{7.10}} & \tiny{4.92} & \tiny{0.96} & \textbf{\tiny{3.22}} & \tiny{2.26}\\\tiny{jpeg\_compression} & \tiny{64.92} & \textbf{\tiny{66.40}} & \tiny{1.48} & \tiny{60.86} & \textbf{\tiny{63.30}} & \tiny{2.44} & \tiny{58.74} & \textbf{\tiny{60.10}} & \tiny{1.36} & \tiny{48.66} & \textbf{\tiny{51.38}} & \tiny{2.72} & \tiny{38.72} & \textbf{\tiny{41.42}} & \tiny{2.70}\\\tiny{motion\_blur} & \tiny{62.32} & \textbf{\tiny{69.18}} & \tiny{6.86} & \tiny{53.26} & \textbf{\tiny{64.08}} & \tiny{10.82} & \tiny{41.32} & \textbf{\tiny{56.18}} & \tiny{14.86} & \tiny{30.74} & \textbf{\tiny{43.20}} & \tiny{12.46} & \tiny{24.20} & \textbf{\tiny{35.00}} & \tiny{10.80}\\\tiny{pixelate} & \tiny{71.12} & \textbf{\tiny{72.80}} & \tiny{1.68} & \tiny{70.48} & \textbf{\tiny{72.60}} & \tiny{2.12} & \tiny{65.46} & \textbf{\tiny{71.74}} & \tiny{6.28} & \tiny{57.70} & \textbf{\tiny{68.72}} & \tiny{11.02} & \tiny{53.32} & \textbf{\tiny{65.38}} & \tiny{12.06}\\\tiny{saturate} & \tiny{55.62} & \textbf{\tiny{55.64}} & \tiny{0.02} & \textbf{\tiny{41.00}} & \tiny{38.56} & \tiny{-2.44} & \textbf{\tiny{66.84}} & \tiny{65.84} & \tiny{-1.00} & \textbf{\tiny{40.98}} & \tiny{35.82} & \tiny{-5.16} & \textbf{\tiny{26.32}} & \tiny{21.88} & \tiny{-4.44}\\\tiny{shot\_noise} & \tiny{51.26} & \textbf{\tiny{57.84}} & \tiny{6.58} & \tiny{28.32} & \textbf{\tiny{40.04}} & \tiny{11.72} & \tiny{10.84} & \textbf{\tiny{22.60}} & \tiny{11.76} & \tiny{2.80} & \textbf{\tiny{9.22}} & \tiny{6.42} & \tiny{1.64} & \textbf{\tiny{5.14}} & \tiny{3.50}\\\tiny{snow} & \tiny{51.98} & \textbf{\tiny{54.38}} & \tiny{2.40} & \tiny{29.66} & \textbf{\tiny{34.36}} & \tiny{4.70} & \tiny{32.56} & \textbf{\tiny{34.70}} & \tiny{2.14} & \tiny{20.02} & \textbf{\tiny{21.36}} & \tiny{1.34} & \tiny{13.96} & \textbf{\tiny{15.20}} & \tiny{1.24}\\\tiny{spatter} & \tiny{73.12} & \textbf{\tiny{73.68}} & \tiny{0.56} & \tiny{61.20} & \textbf{\tiny{64.04}} & \tiny{2.84} & \tiny{47.76} & \textbf{\tiny{53.56}} & \tiny{5.80} & \tiny{37.46} & \textbf{\tiny{45.46}} & \tiny{8.00} & \tiny{26.46} & \textbf{\tiny{36.02}} & \tiny{9.56}\\\tiny{speckle\_noise} & \tiny{56.34} & \textbf{\tiny{62.44}} & \tiny{6.10} & \tiny{46.36} & \textbf{\tiny{54.10}} & \tiny{7.74} & \tiny{20.06} & \textbf{\tiny{30.14}} & \tiny{10.08} & \tiny{11.20} & \textbf{\tiny{19.80}} & \tiny{8.60} & \tiny{5.26} & \textbf{\tiny{12.28}} & \tiny{7.02}\\\tiny{zoom\_blur} & \tiny{55.46} & \textbf{\tiny{65.12}} & \tiny{9.66} & \tiny{49.62} & \textbf{\tiny{61.14}} & \tiny{11.52} & \tiny{44.24} & \textbf{\tiny{56.34}} & \tiny{12.10} & \tiny{40.26} & \textbf{\tiny{51.70}} & \tiny{11.44} & \tiny{35.46} & \textbf{\tiny{45.98}} & \tiny{10.52}\\\tiny{{{$\Sigma$}}} & \tiny{58.07} & \textbf{\tiny{63.19}} & \tiny{5.11} & \tiny{45.17} & \textbf{\tiny{52.80}} & \tiny{7.63} & \tiny{36.68} & \textbf{\tiny{45.36}} & \tiny{8.68} & \tiny{28.11} & \textbf{\tiny{35.29}} & \tiny{7.18} & \tiny{21.17} & \textbf{\tiny{26.44}} & \tiny{5.27}\end{tabular}\caption{Accuracies for ResNeXt50 w/wo OpticsAugment evaluated on ImageNet-100-c 2D common corruptions~\cite{hendrycks_benchmarking_2019}.}\label{tab:tab:imagenet100c_corruptions_ResNeXt50}\end{table*}

\FloatBarrier

\newcommand{\szCommon}{0.7\linewidth}
\begin{figure*}[h]
    \centering 
    \begin{subfigure}{\szCommon}
    \includegraphics[width=\linewidth]{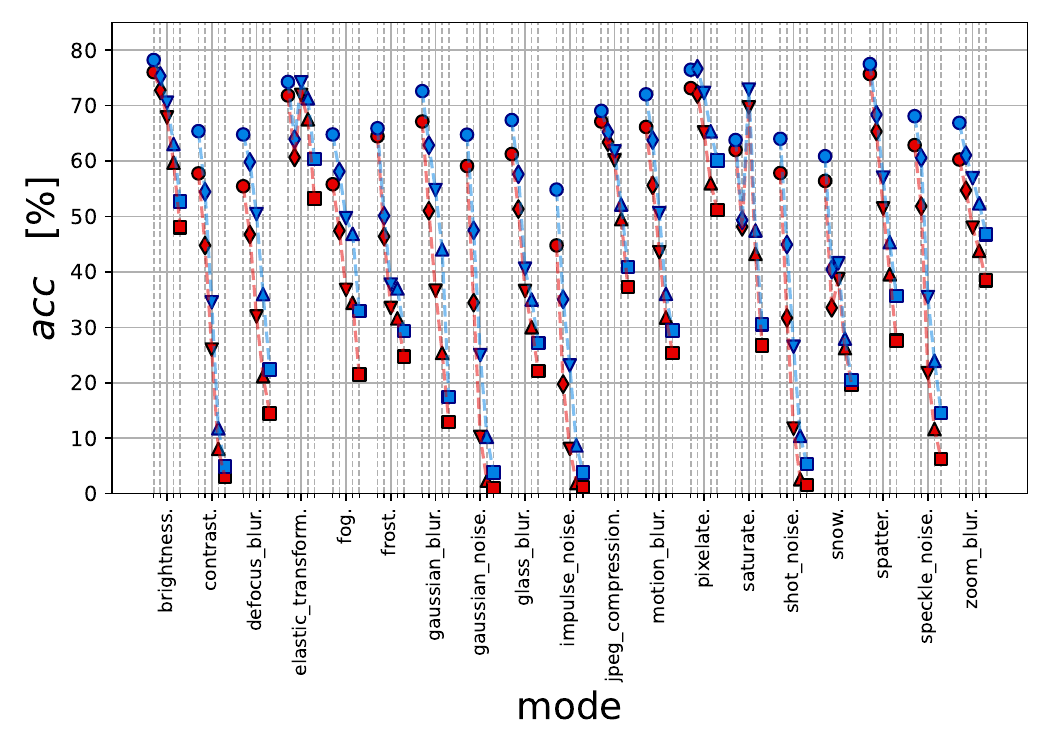}
    \caption{DenseNet161}
    \label{fig:app_imagenet100c_densenet161}
    \end{subfigure} 
    
    \begin{subfigure}{\szCommon}
    \includegraphics[width=\linewidth]{figures/imagenet100-c/eval_rgb/report_acc_over_modesplit_corruptions_highlight_resnext50_32x4d_sgd_clean_resnext50_32x4d_sgd_opticsaugment.pdf}
    \caption{ResNext50}
    \label{fig:app_imagenet100c_resnext50}
    \end{subfigure}
    \caption{Accuracy evaluated on ImageNet-100-C 2D common corruptions for DNNs w/wo OpticsAugment training and all severities 1-5 (circle, diamond, triangles and square markers) at each corruption. \textbf{OpticsAugment (blue)} accuracy compared to the conventionally trained DNN (red):  (a) DenseNet, (b) ResNeXt50.}
    \label{fig:app_imagenet100c_1}
\end{figure*}

\begin{figure*}[h]
    \centering
    \begin{subfigure}{\szCommon}
    \includegraphics[width=\linewidth]{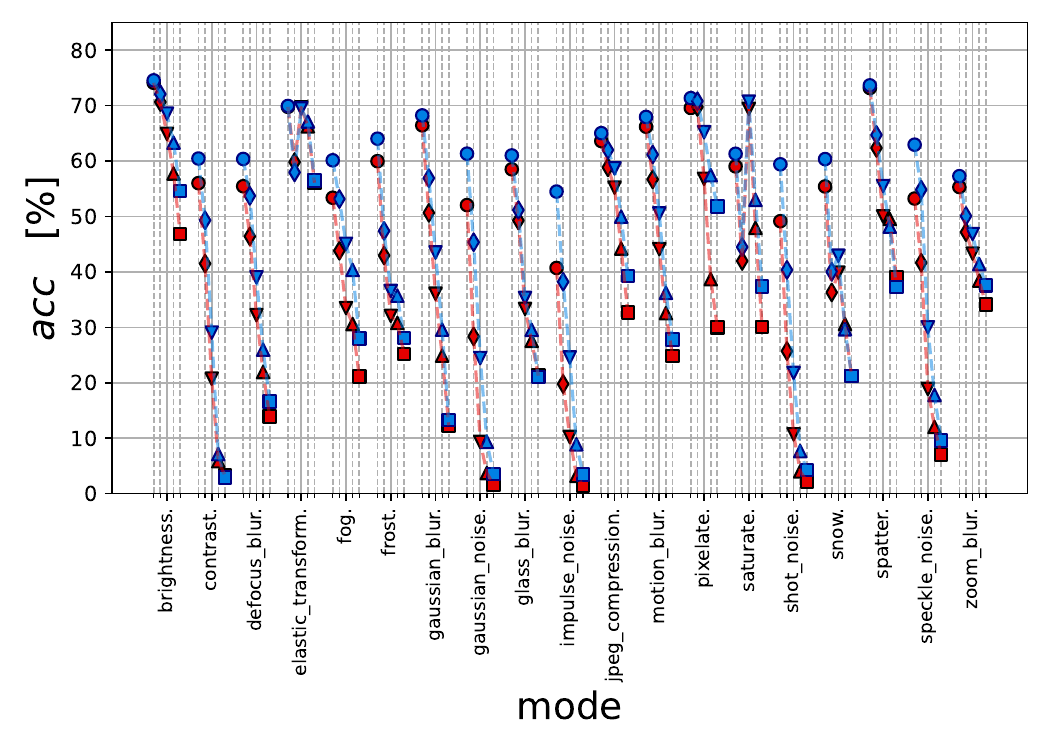}
    \caption{EfficientNet}
    \label{fig:app_imagenet100c_efficientnet_b0}
    \end{subfigure}
    \begin{subfigure}{\szCommon}
    \includegraphics[width=\linewidth]{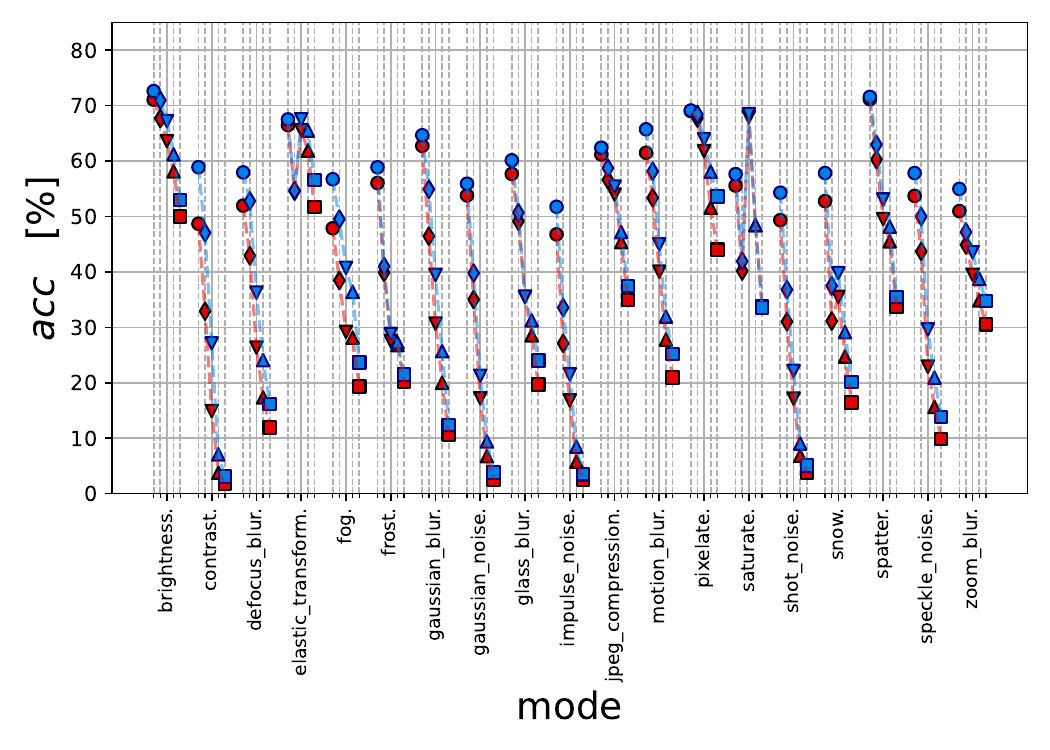}
    \caption{MobileNet}
    \label{fig:app_imagenet100c_mobilenet_v3_large}
    \end{subfigure}   
    \caption{Accuracy evaluated on ImageNet-100-C 2D common corruptions for DNNs w/wo OpticsAugment training and all severities 1-5 (circle, diamond, triangles and square markers) at each corruption. \textbf{OpticsAugment (blue)} accuracy compared to the conventionally trained DNN (red):  (a) EfficientNet, (b) MobileNet.}
    \label{fig:app_imagenet100c_2}
\end{figure*}

\begin{figure*}[h]
    \centering
    \begin{subfigure}{\szCommon}
    \includegraphics[width=\linewidth]{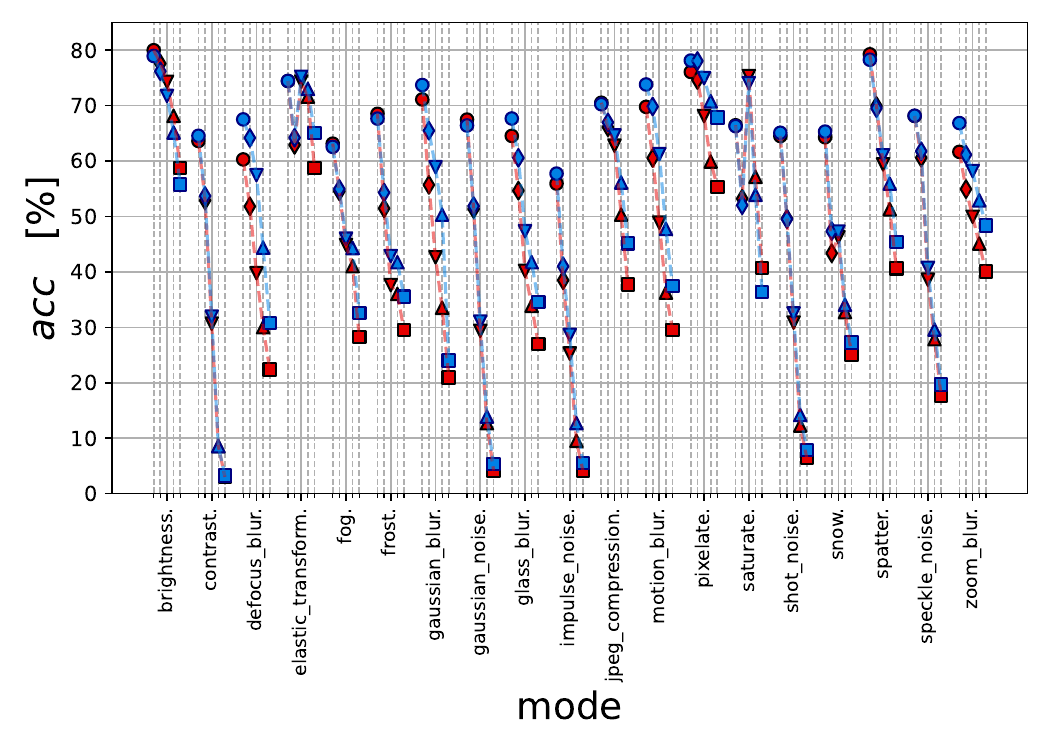}
    \caption{ResNet101}
    \label{fig:app_imagenet100c_resnet101}
    \end{subfigure}    
    \caption{Accuracy evaluated on ImageNet-100-C 2D common corruptions for ResNet101 w/wo OpticsAugment training and all severities 1-5 (circle, diamond, triangles and square markers) at each corruption. 
    \textbf{OpticsAugment (blue) improves} accuracy compared to the conventionally trained DNN (red).}
    \label{fig:app_imagenet100c_3}
\end{figure*}

\vfill
\clearpage
\FloatBarrier
\section{Kernels} 
\label{app:kernels}
All kernels share the same baseline optical wavefront model, which is adapted from~\cite{muller_simulating_2022} and evaluated at the center, i.e. at field 0° with little aberrations, but non-zero. Although isolated Zernike modes are used to generate the different kernels, this ensures a more realistic PSF to avoid for instance a PSF depending solely on coma aberration. The baseline model consists of the wavefront description displayed in Tab.~\ref{tab:baseline_wf_model}.
The different kernels are then generated by adding the isolated Zernike modes from  Tab.~\ref{tab:zernike_fringe_modes} to the baseline wavefront model with Eq.~\ref{eq:zernike_expansion} and retrieving a PSF with Eq.~\ref{eq:propagation}.  Although real lenses may consist of dozens of different balanced Zernike modes, the amplifying of a particular Zernike mode allows for categorization and  benchmarking to particular aberrations. This creates the kernels from Fig.~\ref{fig:kernels_cooke_rgb_iccv}. In practice, a more balanced distribution of coefficients is observed. 
\begin{table}[h]
\centering 
\begin{tabular}{@{}lllll@{}}
Color  &  4 & 9 & 15 & 16 \\
\hline
red 	& 	0.32671		&	  	0.088223	&  -0.061867	& -4.7631E-06 \\
green 	& 	0.11273		  & 	0.095923	&  -0.069497	&  -5.3967E-06 \\
blue 	& 	-0.41772	  & 	0.10825	    & -0.085119	&  -6.7436E-06 

\end{tabular}
\caption{Wavefront baseline model used to produce the kernels (a,d,e) in Fig.~\ref{fig:psf_samples} and Fig.~\ref{fig:kernels_cooke_rgb_iccv}. Other coefficients are zero, each value is in multiples of the wavelength $\lambda$ for RGB color channels red, green and blue: \SIlist{0.6563;0.5876;0.4861}{\micro\metre}. Zernike modes are in Fringe ordering, from left to right: defocus, spherical, secondary spherical and vertical quadrofoil as from~\cite{lakshminarayanan_zernike_2011}.}
\label{tab:baseline_wf_model}
\end{table}

We also include another experimental set of kernels sharing the baseline model from  Tab.~\ref{tab:baseline_wf_model}, but, to further increase chromatic aberrations, with red and blue channels merged. Thus, only the blue and green coefficient values are used and the red channel shares the same coefficients as the blue channel. Wavelength dependent scaling is turned off for the merged channels. The green channel is unchanged. This creates the reddish and greenish PSF kernels from Fig.~\ref{fig:kernels_cooke_rg_iccv} and the kernels (b,c,f) in Fig.~\ref{fig:psf_samples}, why we call the set of kernels RG or OpticsBenchRG.

The defocus blur kernels from~\cite{hendrycks_benchmarking_2019} are reproduced in Fig.~\ref{fig:kernels_baseline_defocus} for all severities to allow for comparison to disk-shaped kernels. As these kernels equally blur all color channels, they are grayscaled. 
\newcommand{\ksize}{0.098}
\newcommand{\ksizeBase}{0.098}
\begin{figure}[h]
\centering\begin{subfigure}{\ksizeBase\linewidth}
\centering
\includegraphics[trim=3.75cm 1.5cm 3.5cm 1.5cm,clip,width=\linewidth]{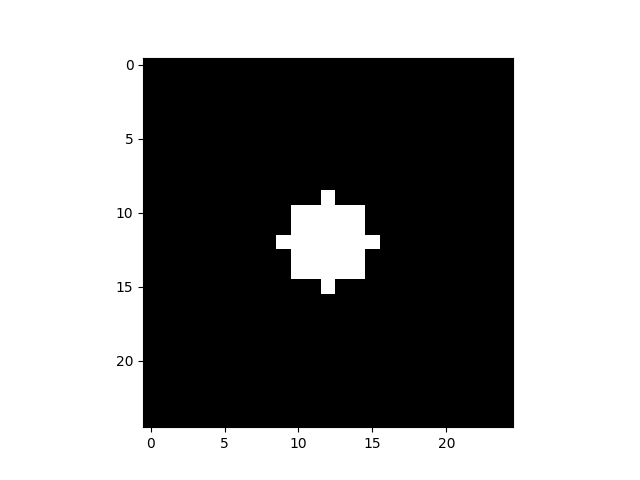}
\end{subfigure}%
\begin{subfigure}{\ksizeBase\linewidth}
\centering
\includegraphics[trim=3.75cm 1.5cm 3.5cm 1.5cm,clip,width=\linewidth]{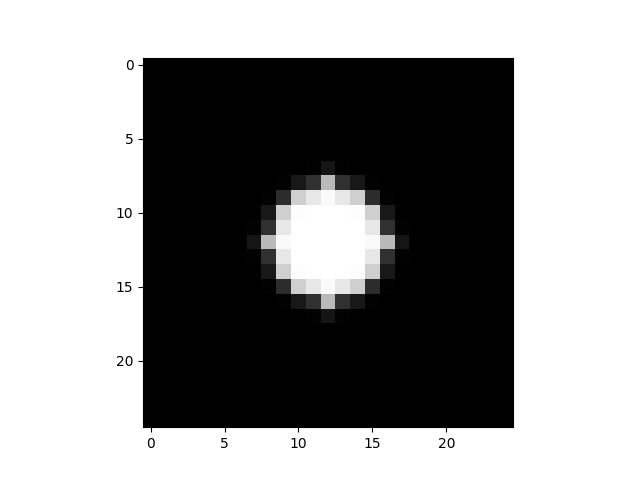}
\end{subfigure}%
\begin{subfigure}{\ksizeBase\linewidth}
\centering
\includegraphics[trim=3.75cm 1.5cm 3.5cm 1.5cm,clip,width=\linewidth]{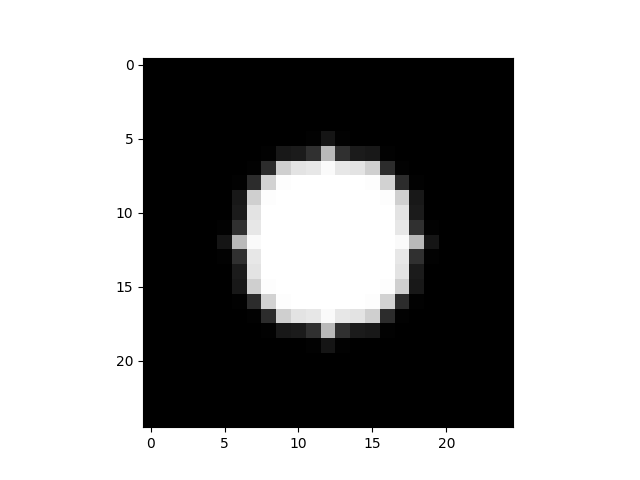}
\end{subfigure}%
\begin{subfigure}{\ksizeBase\linewidth}
\centering
\includegraphics[trim=3.75cm 1.5cm 3.5cm 1.5cm,clip,width=\linewidth]{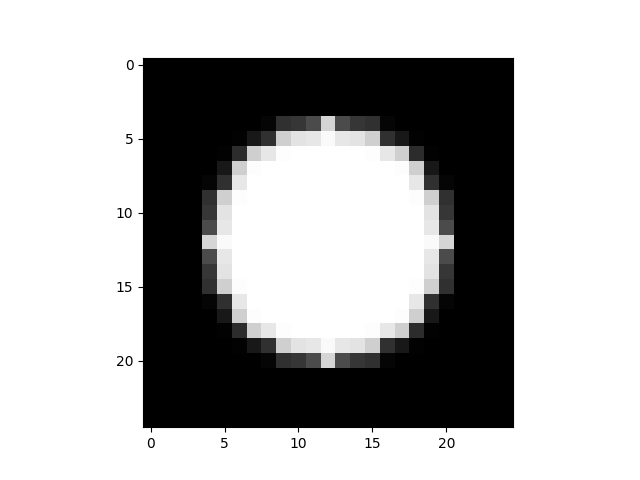}
\end{subfigure}%
\begin{subfigure}{\ksizeBase\linewidth}
\centering
\includegraphics[trim=3.75cm 1.5cm 3.5cm 1.5cm,clip,width=\linewidth]{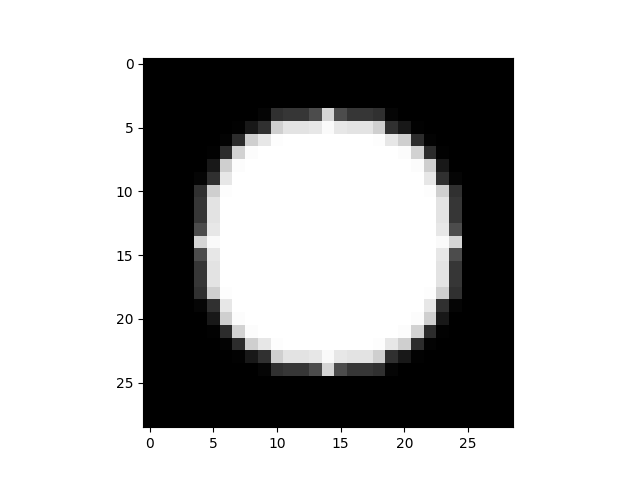}
\end{subfigure}%
\caption{Defocus blur from~\cite{hendrycks_benchmarking_2019} for severities 1-5 used for kernel matching and as \emph{base blur type} for comparisons.}
\label{fig:kernels_baseline_defocus}
\end{figure}


\begin{figure}[h]\begin{subfigure}{\linewidth}\begin{subfigure}{\ksize\linewidth}\centering\includegraphics[width=\linewidth]{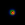}\end{subfigure}\begin{subfigure}{\ksize\linewidth}\centering\includegraphics[width=\linewidth]{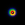}\end{subfigure}\begin{subfigure}{\ksize\linewidth}\centering\includegraphics[width=\linewidth]{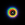}\end{subfigure}\begin{subfigure}{\ksize\linewidth}\centering\includegraphics[width=\linewidth]{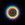}\end{subfigure}\begin{subfigure}{\ksize\linewidth}\centering\includegraphics[width=\linewidth]{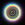}\end{subfigure} \hspace{0.005cm} \begin{subfigure}{\ksize\linewidth}\centering\includegraphics[width=\linewidth]{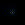}\end{subfigure}\begin{subfigure}{\ksize\linewidth}\centering\includegraphics[width=\linewidth]{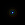}\end{subfigure}\begin{subfigure}{\ksize\linewidth}\centering\includegraphics[width=\linewidth]{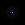}\end{subfigure}\begin{subfigure}{\ksize\linewidth}\centering\includegraphics[width=\linewidth]{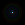}\end{subfigure}\begin{subfigure}{\ksize\linewidth}\centering\includegraphics[width=\linewidth]{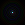}\end{subfigure}\caption{}\end{subfigure}\\\begin{subfigure}{\linewidth}\begin{subfigure}{\ksize\linewidth}\centering\includegraphics[width=\linewidth]{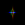}\end{subfigure}\begin{subfigure}{\ksize\linewidth}\centering\includegraphics[width=\linewidth]{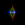}\end{subfigure}\begin{subfigure}{\ksize\linewidth}\centering\includegraphics[width=\linewidth]{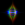}\end{subfigure}\begin{subfigure}{\ksize\linewidth}\centering\includegraphics[width=\linewidth]{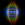}\end{subfigure}\begin{subfigure}{\ksize\linewidth}\centering\includegraphics[width=\linewidth]{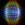}\end{subfigure} \hspace{0.005cm} \begin{subfigure}{\ksize\linewidth}\centering\includegraphics[width=\linewidth]{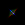}\end{subfigure}\begin{subfigure}{\ksize\linewidth}\centering\includegraphics[width=\linewidth]{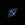}\end{subfigure}\begin{subfigure}{\ksize\linewidth}\centering\includegraphics[width=\linewidth]{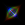}\end{subfigure}\begin{subfigure}{\ksize\linewidth}\centering\includegraphics[width=\linewidth]{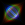}\end{subfigure}\begin{subfigure}{\ksize\linewidth}\centering\includegraphics[width=\linewidth]{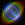}\end{subfigure}\caption{}\end{subfigure}\\\begin{subfigure}{\linewidth}\begin{subfigure}{\ksize\linewidth}\centering\includegraphics[width=\linewidth]{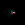}\end{subfigure}\begin{subfigure}{\ksize\linewidth}\centering\includegraphics[width=\linewidth]{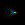}\end{subfigure}\begin{subfigure}{\ksize\linewidth}\centering\includegraphics[width=\linewidth]{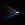}\end{subfigure}\begin{subfigure}{\ksize\linewidth}\centering\includegraphics[width=\linewidth]{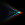}\end{subfigure}\begin{subfigure}{\ksize\linewidth}\centering\includegraphics[width=\linewidth]{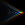}\end{subfigure} \hspace{0.005cm} \begin{subfigure}{\ksize\linewidth}\centering\includegraphics[width=\linewidth]{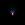}\end{subfigure}\begin{subfigure}{\ksize\linewidth}\centering\includegraphics[width=\linewidth]{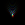}\end{subfigure}\begin{subfigure}{\ksize\linewidth}\centering\includegraphics[width=\linewidth]{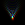}\end{subfigure}\begin{subfigure}{\ksize\linewidth}\centering\includegraphics[width=\linewidth]{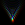}\end{subfigure}\begin{subfigure}{\ksize\linewidth}\centering\includegraphics[width=\linewidth]{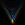}\end{subfigure}\caption{}\end{subfigure}\\\begin{subfigure}{\linewidth}\begin{subfigure}{\ksize\linewidth}\centering\includegraphics[width=\linewidth]{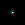}\end{subfigure}\begin{subfigure}{\ksize\linewidth}\centering\includegraphics[width=\linewidth]{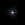}\end{subfigure}\begin{subfigure}{\ksize\linewidth}\centering\includegraphics[width=\linewidth]{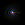}\end{subfigure}\begin{subfigure}{\ksize\linewidth}\centering\includegraphics[width=\linewidth]{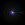}\end{subfigure}\begin{subfigure}{\ksize\linewidth}\centering\includegraphics[width=\linewidth]{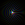}\end{subfigure} \hspace{0.005cm} \begin{subfigure}{\ksize\linewidth}\centering\includegraphics[width=\linewidth]{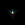}\end{subfigure}\begin{subfigure}{\ksize\linewidth}\centering\includegraphics[width=\linewidth]{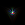}\end{subfigure}\begin{subfigure}{\ksize\linewidth}\centering\includegraphics[width=\linewidth]{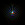}\end{subfigure}\begin{subfigure}{\ksize\linewidth}\centering\includegraphics[width=\linewidth]{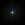}\end{subfigure}\begin{subfigure}{\ksize\linewidth}\centering\includegraphics[width=\linewidth]{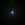}\end{subfigure}\caption{}\end{subfigure}\\\caption{Kernels used to generate OpticsBench. Each row contains the different severities (1-5) for a single corruption using two Zernike modes. Larger kernel size leads to more severe blurring. (a) Defocus \& Spherical, (b) Astigmatism, (c) Coma, (d) Trefoil. All kernels are $l_1$-normalized and therefore have the same energy. }\label{fig:kernels_cooke_rgb_iccv}\end{figure}


\begin{figure}[h]\begin{subfigure}{\linewidth}\begin{subfigure}{\ksize\linewidth}\centering\includegraphics[width=\linewidth]{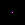}\end{subfigure}\begin{subfigure}{\ksize\linewidth}\centering\includegraphics[width=\linewidth]{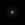}\end{subfigure}\begin{subfigure}{\ksize\linewidth}\centering\includegraphics[width=\linewidth]{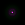}\end{subfigure}\begin{subfigure}{\ksize\linewidth}\centering\includegraphics[width=\linewidth]{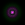}\end{subfigure}\begin{subfigure}{\ksize\linewidth}\centering\includegraphics[width=\linewidth]{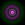}\end{subfigure} \hspace{0.005cm} \begin{subfigure}{\ksize\linewidth}\centering\includegraphics[width=\linewidth]{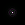}\end{subfigure}\begin{subfigure}{\ksize\linewidth}\centering\includegraphics[width=\linewidth]{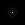}\end{subfigure}\begin{subfigure}{\ksize\linewidth}\centering\includegraphics[width=\linewidth]{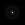}\end{subfigure}\begin{subfigure}{\ksize\linewidth}\centering\includegraphics[width=\linewidth]{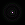}\end{subfigure}\begin{subfigure}{\ksize\linewidth}\centering\includegraphics[width=\linewidth]{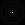}\end{subfigure}\caption{}\end{subfigure}\\\begin{subfigure}{\linewidth}\begin{subfigure}{\ksize\linewidth}\centering\includegraphics[width=\linewidth]{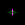}\end{subfigure}\begin{subfigure}{\ksize\linewidth}\centering\includegraphics[width=\linewidth]{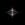}\end{subfigure}\begin{subfigure}{\ksize\linewidth}\centering\includegraphics[width=\linewidth]{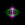}\end{subfigure}\begin{subfigure}{\ksize\linewidth}\centering\includegraphics[width=\linewidth]{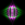}\end{subfigure}\begin{subfigure}{\ksize\linewidth}\centering\includegraphics[width=\linewidth]{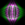}\end{subfigure} \hspace{0.005cm} \begin{subfigure}{\ksize\linewidth}\centering\includegraphics[width=\linewidth]{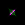}\end{subfigure}\begin{subfigure}{\ksize\linewidth}\centering\includegraphics[width=\linewidth]{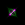}\end{subfigure}\begin{subfigure}{\ksize\linewidth}\centering\includegraphics[width=\linewidth]{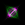}\end{subfigure}\begin{subfigure}{\ksize\linewidth}\centering\includegraphics[width=\linewidth]{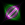}\end{subfigure}\begin{subfigure}{\ksize\linewidth}\centering\includegraphics[width=\linewidth]{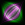}\end{subfigure}\caption{}\end{subfigure}\\\begin{subfigure}{\linewidth}\begin{subfigure}{\ksize\linewidth}\centering\includegraphics[width=\linewidth]{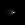}\end{subfigure}\begin{subfigure}{\ksize\linewidth}\centering\includegraphics[width=\linewidth]{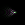}\end{subfigure}\begin{subfigure}{\ksize\linewidth}\centering\includegraphics[width=\linewidth]{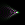}\end{subfigure}\begin{subfigure}{\ksize\linewidth}\centering\includegraphics[width=\linewidth]{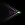}\end{subfigure}\begin{subfigure}{\ksize\linewidth}\centering\includegraphics[width=\linewidth]{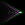}\end{subfigure} \hspace{0.005cm} \begin{subfigure}{\ksize\linewidth}\centering\includegraphics[width=\linewidth]{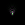}\end{subfigure}\begin{subfigure}{\ksize\linewidth}\centering\includegraphics[width=\linewidth]{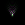}\end{subfigure}\begin{subfigure}{\ksize\linewidth}\centering\includegraphics[width=\linewidth]{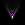}\end{subfigure}\begin{subfigure}{\ksize\linewidth}\centering\includegraphics[width=\linewidth]{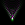}\end{subfigure}\begin{subfigure}{\ksize\linewidth}\centering\includegraphics[width=\linewidth]{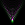}\end{subfigure}\caption{}\end{subfigure}\\\begin{subfigure}{\linewidth}\begin{subfigure}{\ksize\linewidth}\centering\includegraphics[width=\linewidth]{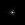}\end{subfigure}\begin{subfigure}{\ksize\linewidth}\centering\includegraphics[width=\linewidth]{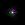}\end{subfigure}\begin{subfigure}{\ksize\linewidth}\centering\includegraphics[width=\linewidth]{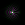}\end{subfigure}\begin{subfigure}{\ksize\linewidth}\centering\includegraphics[width=\linewidth]{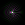}\end{subfigure}\begin{subfigure}{\ksize\linewidth}\centering\includegraphics[width=\linewidth]{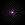}\end{subfigure} \hspace{0.005cm} \begin{subfigure}{\ksize\linewidth}\centering\includegraphics[width=\linewidth]{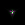}\end{subfigure}\begin{subfigure}{\ksize\linewidth}\centering\includegraphics[width=\linewidth]{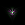}\end{subfigure}\begin{subfigure}{\ksize\linewidth}\centering\includegraphics[width=\linewidth]{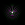}\end{subfigure}\begin{subfigure}{\ksize\linewidth}\centering\includegraphics[width=\linewidth]{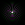}\end{subfigure}\begin{subfigure}{\ksize\linewidth}\centering\includegraphics[width=\linewidth]{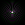}\end{subfigure}\caption{}\end{subfigure}\\\caption{Kernels used to generate OpticsBenchRG. Each row contains the different severities (1-5) for a single corruption using two Zernike modes. Larger kernel size leads to more severe blurring. (a) Defocus \& Spherical, (b) Astigmatism, (c) Coma, (d) Trefoil. All kernels are $l_1$-normalized and therefore have the same energy. }\label{fig:kernels_cooke_rg_iccv}\end{figure}
\FloatBarrier
\section{Image examples} 
\label{app:image_examples}
This section shows images from OpticsBench. Each row contains a single corruption and three image examples with increasing severities (from left to right). The corruptions are sorted as: astigmatism, defocus \& spherical, coma, trefoil. The upper left image represents astigmatism at severity 1, the lower right image shows trefoil at severity 5.

\begin{figure*} \centering\begin{subfigure}{0.065\linewidth}\centering\includegraphics[width=\linewidth]{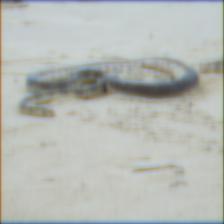}\end{subfigure}\begin{subfigure}{0.065\linewidth}\centering\includegraphics[width=\linewidth]{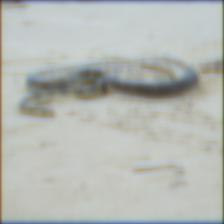}\end{subfigure}\begin{subfigure}{0.065\linewidth}\centering\includegraphics[width=\linewidth]{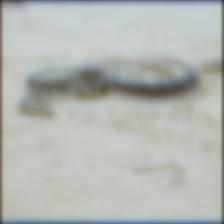}\end{subfigure}\begin{subfigure}{0.065\linewidth}\centering\includegraphics[width=\linewidth]{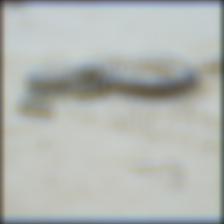}\end{subfigure}\begin{subfigure}{0.065\linewidth}\centering\includegraphics[width=\linewidth]{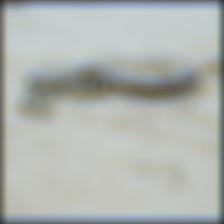}\end{subfigure}\begin{subfigure}{0.065\linewidth}\centering\includegraphics[width=\linewidth]{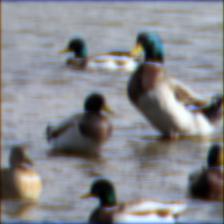}\end{subfigure}\begin{subfigure}{0.065\linewidth}\centering\includegraphics[width=\linewidth]{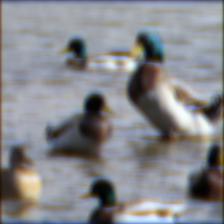}\end{subfigure}\begin{subfigure}{0.065\linewidth}\centering\includegraphics[width=\linewidth]{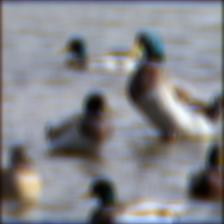}\end{subfigure}\begin{subfigure}{0.065\linewidth}\centering\includegraphics[width=\linewidth]{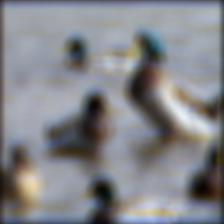}\end{subfigure}\begin{subfigure}{0.065\linewidth}\centering\includegraphics[width=\linewidth]{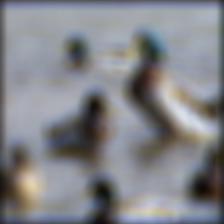}\end{subfigure}\begin{subfigure}{0.065\linewidth}\centering\includegraphics[width=\linewidth]{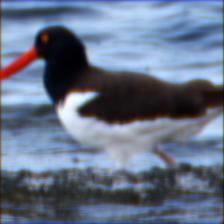}\end{subfigure}\begin{subfigure}{0.065\linewidth}\centering\includegraphics[width=\linewidth]{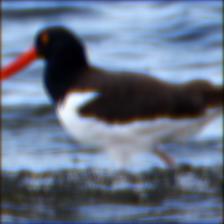}\end{subfigure}\begin{subfigure}{0.065\linewidth}\centering\includegraphics[width=\linewidth]{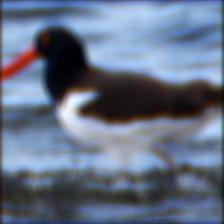}\end{subfigure}\begin{subfigure}{0.065\linewidth}\centering\includegraphics[width=\linewidth]{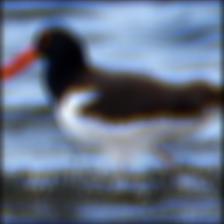}\end{subfigure}\begin{subfigure}{0.065\linewidth}\centering\includegraphics[width=\linewidth]{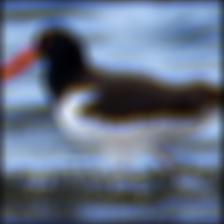}\end{subfigure}\\\begin{subfigure}{0.065\linewidth}\centering\includegraphics[width=\linewidth]{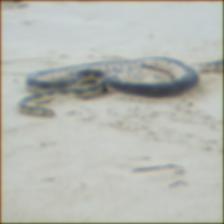}\end{subfigure}\begin{subfigure}{0.065\linewidth}\centering\includegraphics[width=\linewidth]{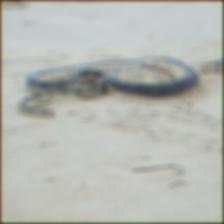}\end{subfigure}\begin{subfigure}{0.065\linewidth}\centering\includegraphics[width=\linewidth]{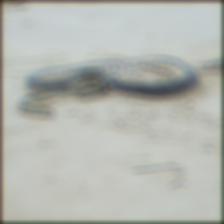}\end{subfigure}\begin{subfigure}{0.065\linewidth}\centering\includegraphics[width=\linewidth]{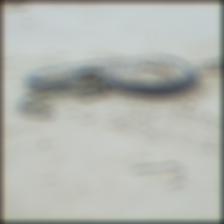}\end{subfigure}\begin{subfigure}{0.065\linewidth}\centering\includegraphics[width=\linewidth]{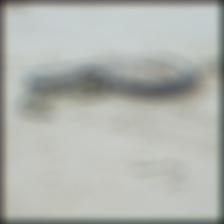}\end{subfigure}\begin{subfigure}{0.065\linewidth}\centering\includegraphics[width=\linewidth]{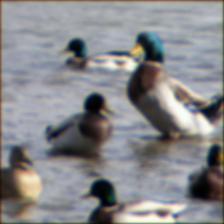}\end{subfigure}\begin{subfigure}{0.065\linewidth}\centering\includegraphics[width=\linewidth]{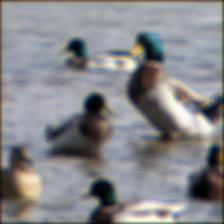}\end{subfigure}\begin{subfigure}{0.065\linewidth}\centering\includegraphics[width=\linewidth]{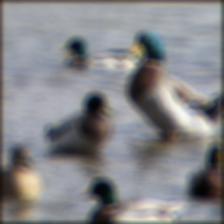}\end{subfigure}\begin{subfigure}{0.065\linewidth}\centering\includegraphics[width=\linewidth]{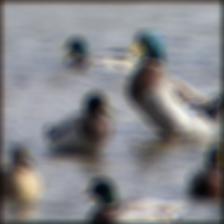}\end{subfigure}\begin{subfigure}{0.065\linewidth}\centering\includegraphics[width=\linewidth]{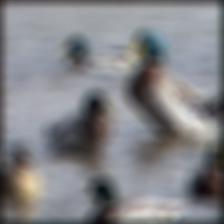}\end{subfigure}\begin{subfigure}{0.065\linewidth}\centering\includegraphics[width=\linewidth]{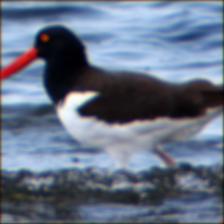}\end{subfigure}\begin{subfigure}{0.065\linewidth}\centering\includegraphics[width=\linewidth]{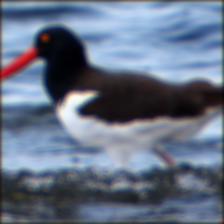}\end{subfigure}\begin{subfigure}{0.065\linewidth}\centering\includegraphics[width=\linewidth]{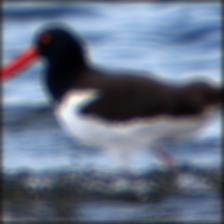}\end{subfigure}\begin{subfigure}{0.065\linewidth}\centering\includegraphics[width=\linewidth]{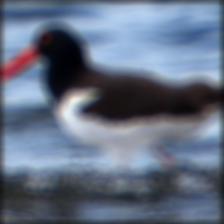}\end{subfigure}\begin{subfigure}{0.065\linewidth}\centering\includegraphics[width=\linewidth]{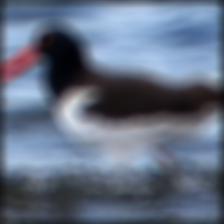}\end{subfigure}\\\begin{subfigure}{0.065\linewidth}\centering\includegraphics[width=\linewidth]{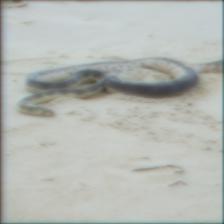}\end{subfigure}\begin{subfigure}{0.065\linewidth}\centering\includegraphics[width=\linewidth]{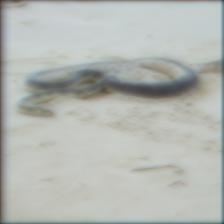}\end{subfigure}\begin{subfigure}{0.065\linewidth}\centering\includegraphics[width=\linewidth]{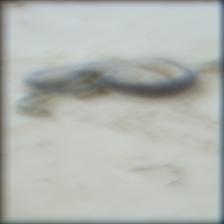}\end{subfigure}\begin{subfigure}{0.065\linewidth}\centering\includegraphics[width=\linewidth]{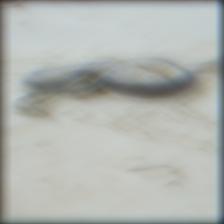}\end{subfigure}\begin{subfigure}{0.065\linewidth}\centering\includegraphics[width=\linewidth]{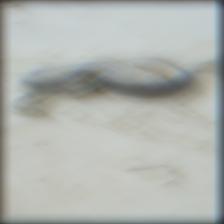}\end{subfigure}\begin{subfigure}{0.065\linewidth}\centering\includegraphics[width=\linewidth]{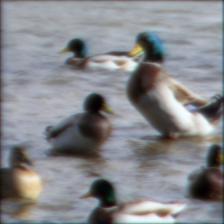}\end{subfigure}\begin{subfigure}{0.065\linewidth}\centering\includegraphics[width=\linewidth]{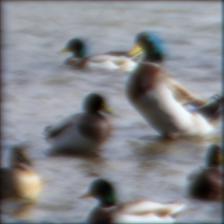}\end{subfigure}\begin{subfigure}{0.065\linewidth}\centering\includegraphics[width=\linewidth]{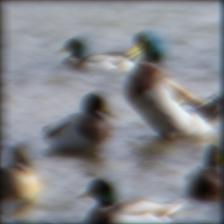}\end{subfigure}\begin{subfigure}{0.065\linewidth}\centering\includegraphics[width=\linewidth]{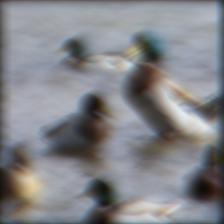}\end{subfigure}\begin{subfigure}{0.065\linewidth}\centering\includegraphics[width=\linewidth]{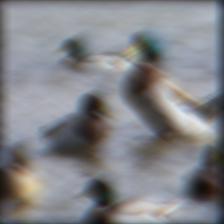}\end{subfigure}\begin{subfigure}{0.065\linewidth}\centering\includegraphics[width=\linewidth]{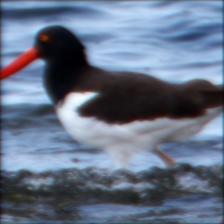}\end{subfigure}\begin{subfigure}{0.065\linewidth}\centering\includegraphics[width=\linewidth]{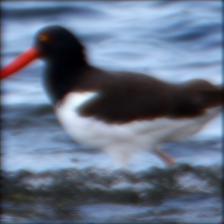}\end{subfigure}\begin{subfigure}{0.065\linewidth}\centering\includegraphics[width=\linewidth]{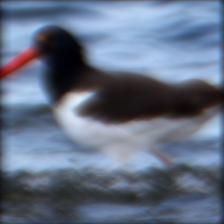}\end{subfigure}\begin{subfigure}{0.065\linewidth}\centering\includegraphics[width=\linewidth]{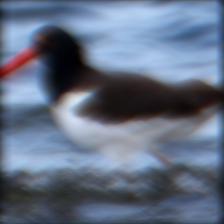}\end{subfigure}\begin{subfigure}{0.065\linewidth}\centering\includegraphics[width=\linewidth]{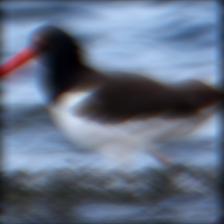}\end{subfigure}\\\begin{subfigure}{0.065\linewidth}\centering\includegraphics[width=\linewidth]{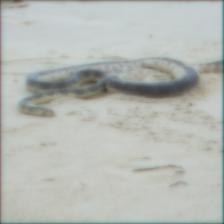}\end{subfigure}\begin{subfigure}{0.065\linewidth}\centering\includegraphics[width=\linewidth]{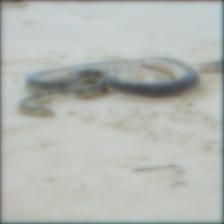}\end{subfigure}\begin{subfigure}{0.065\linewidth}\centering\includegraphics[width=\linewidth]{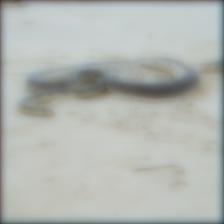}\end{subfigure}\begin{subfigure}{0.065\linewidth}\centering\includegraphics[width=\linewidth]{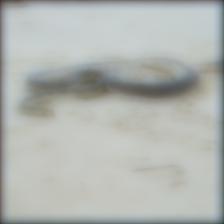}\end{subfigure}\begin{subfigure}{0.065\linewidth}\centering\includegraphics[width=\linewidth]{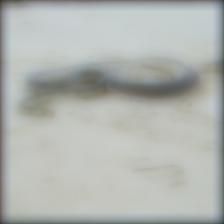}\end{subfigure}\begin{subfigure}{0.065\linewidth}\centering\includegraphics[width=\linewidth]{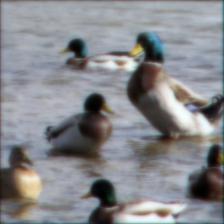}\end{subfigure}\begin{subfigure}{0.065\linewidth}\centering\includegraphics[width=\linewidth]{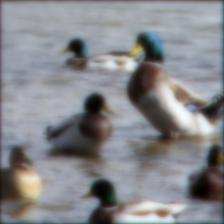}\end{subfigure}\begin{subfigure}{0.065\linewidth}\centering\includegraphics[width=\linewidth]{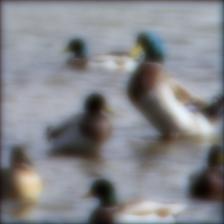}\end{subfigure}\begin{subfigure}{0.065\linewidth}\centering\includegraphics[width=\linewidth]{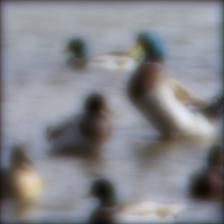}\end{subfigure}\begin{subfigure}{0.065\linewidth}\centering\includegraphics[width=\linewidth]{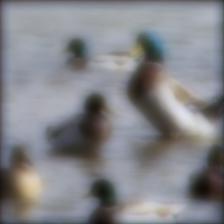}\end{subfigure}\begin{subfigure}{0.065\linewidth}\centering\includegraphics[width=\linewidth]{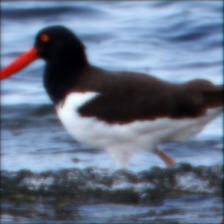}\end{subfigure}\begin{subfigure}{0.065\linewidth}\centering\includegraphics[width=\linewidth]{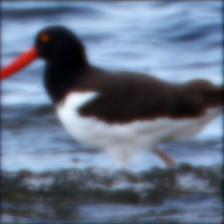}\end{subfigure}\begin{subfigure}{0.065\linewidth}\centering\includegraphics[width=\linewidth]{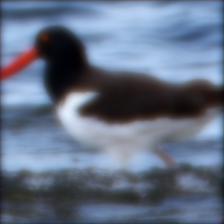}\end{subfigure}\begin{subfigure}{0.065\linewidth}\centering\includegraphics[width=\linewidth]{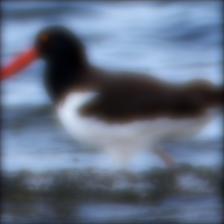}\end{subfigure}\begin{subfigure}{0.065\linewidth}\centering\includegraphics[width=\linewidth]{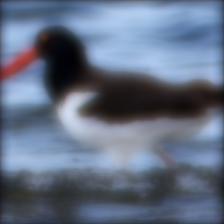}\end{subfigure}\\\begin{subfigure}{0.065\linewidth}\centering\includegraphics[width=\linewidth]{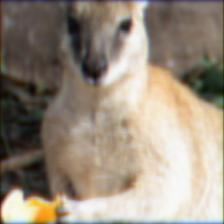}\end{subfigure}\begin{subfigure}{0.065\linewidth}\centering\includegraphics[width=\linewidth]{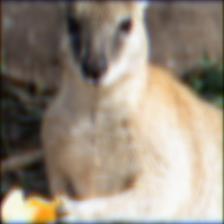}\end{subfigure}\begin{subfigure}{0.065\linewidth}\centering\includegraphics[width=\linewidth]{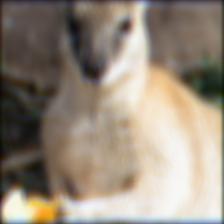}\end{subfigure}\begin{subfigure}{0.065\linewidth}\centering\includegraphics[width=\linewidth]{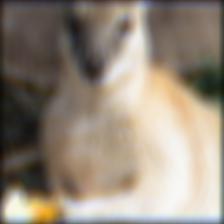}\end{subfigure}\begin{subfigure}{0.065\linewidth}\centering\includegraphics[width=\linewidth]{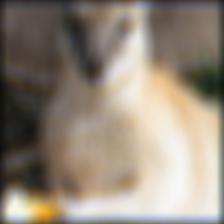}\end{subfigure}\begin{subfigure}{0.065\linewidth}\centering\includegraphics[width=\linewidth]{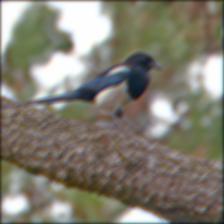}\end{subfigure}\begin{subfigure}{0.065\linewidth}\centering\includegraphics[width=\linewidth]{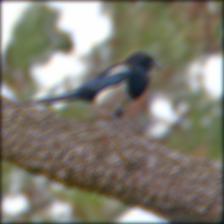}\end{subfigure}\begin{subfigure}{0.065\linewidth}\centering\includegraphics[width=\linewidth]{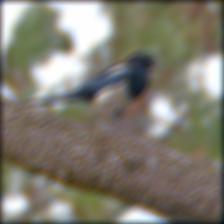}\end{subfigure}\begin{subfigure}{0.065\linewidth}\centering\includegraphics[width=\linewidth]{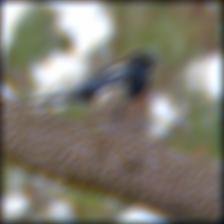}\end{subfigure}\begin{subfigure}{0.065\linewidth}\centering\includegraphics[width=\linewidth]{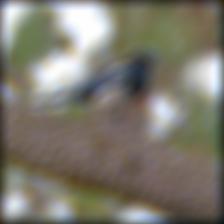}\end{subfigure}\begin{subfigure}{0.065\linewidth}\centering\includegraphics[width=\linewidth]{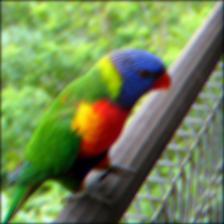}\end{subfigure}\begin{subfigure}{0.065\linewidth}\centering\includegraphics[width=\linewidth]{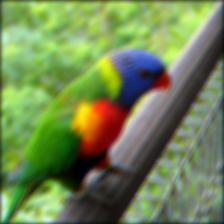}\end{subfigure}\begin{subfigure}{0.065\linewidth}\centering\includegraphics[width=\linewidth]{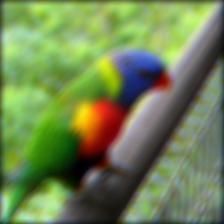}\end{subfigure}\begin{subfigure}{0.065\linewidth}\centering\includegraphics[width=\linewidth]{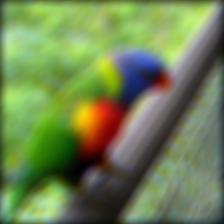}\end{subfigure}\begin{subfigure}{0.065\linewidth}\centering\includegraphics[width=\linewidth]{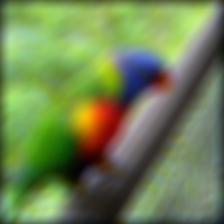}\end{subfigure}\\\begin{subfigure}{0.065\linewidth}\centering\includegraphics[width=\linewidth]{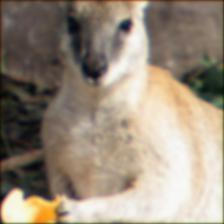}\end{subfigure}\begin{subfigure}{0.065\linewidth}\centering\includegraphics[width=\linewidth]{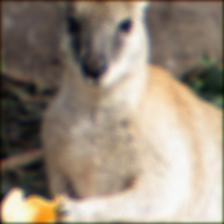}\end{subfigure}\begin{subfigure}{0.065\linewidth}\centering\includegraphics[width=\linewidth]{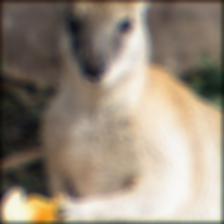}\end{subfigure}\begin{subfigure}{0.065\linewidth}\centering\includegraphics[width=\linewidth]{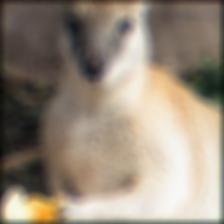}\end{subfigure}\begin{subfigure}{0.065\linewidth}\centering\includegraphics[width=\linewidth]{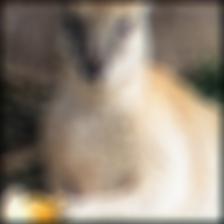}\end{subfigure}\begin{subfigure}{0.065\linewidth}\centering\includegraphics[width=\linewidth]{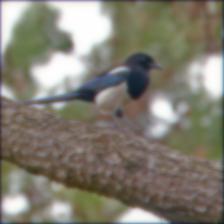}\end{subfigure}\begin{subfigure}{0.065\linewidth}\centering\includegraphics[width=\linewidth]{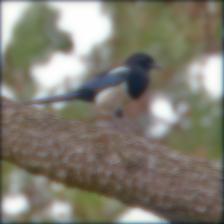}\end{subfigure}\begin{subfigure}{0.065\linewidth}\centering\includegraphics[width=\linewidth]{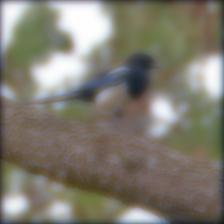}\end{subfigure}\begin{subfigure}{0.065\linewidth}\centering\includegraphics[width=\linewidth]{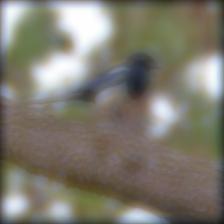}\end{subfigure}\begin{subfigure}{0.065\linewidth}\centering\includegraphics[width=\linewidth]{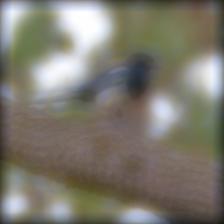}\end{subfigure}\begin{subfigure}{0.065\linewidth}\centering\includegraphics[width=\linewidth]{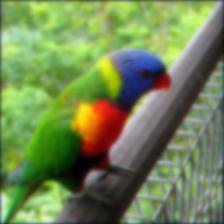}\end{subfigure}\begin{subfigure}{0.065\linewidth}\centering\includegraphics[width=\linewidth]{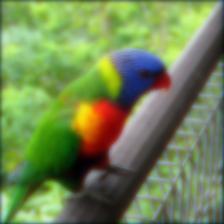}\end{subfigure}\begin{subfigure}{0.065\linewidth}\centering\includegraphics[width=\linewidth]{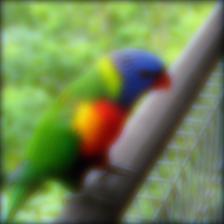}\end{subfigure}\begin{subfigure}{0.065\linewidth}\centering\includegraphics[width=\linewidth]{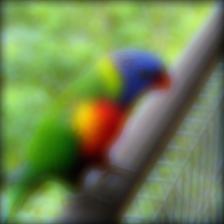}\end{subfigure}\begin{subfigure}{0.065\linewidth}\centering\includegraphics[width=\linewidth]{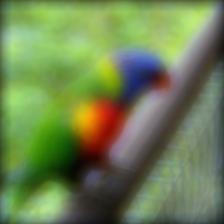}\end{subfigure}\\\begin{subfigure}{0.065\linewidth}\centering\includegraphics[width=\linewidth]{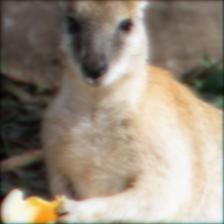}\end{subfigure}\begin{subfigure}{0.065\linewidth}\centering\includegraphics[width=\linewidth]{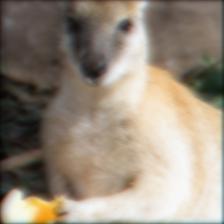}\end{subfigure}\begin{subfigure}{0.065\linewidth}\centering\includegraphics[width=\linewidth]{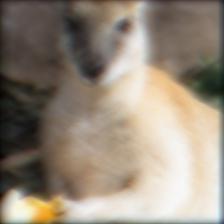}\end{subfigure}\begin{subfigure}{0.065\linewidth}\centering\includegraphics[width=\linewidth]{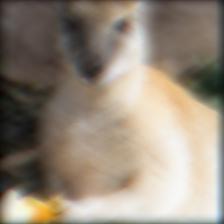}\end{subfigure}\begin{subfigure}{0.065\linewidth}\centering\includegraphics[width=\linewidth]{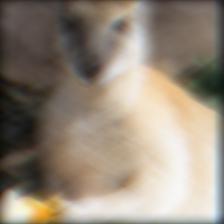}\end{subfigure}\begin{subfigure}{0.065\linewidth}\centering\includegraphics[width=\linewidth]{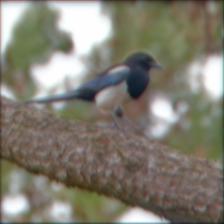}\end{subfigure}\begin{subfigure}{0.065\linewidth}\centering\includegraphics[width=\linewidth]{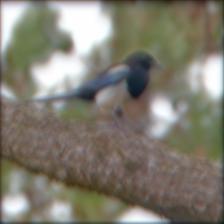}\end{subfigure}\begin{subfigure}{0.065\linewidth}\centering\includegraphics[width=\linewidth]{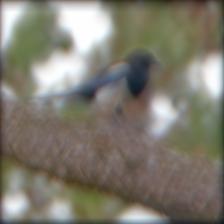}\end{subfigure}\begin{subfigure}{0.065\linewidth}\centering\includegraphics[width=\linewidth]{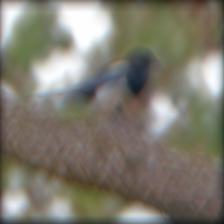}\end{subfigure}\begin{subfigure}{0.065\linewidth}\centering\includegraphics[width=\linewidth]{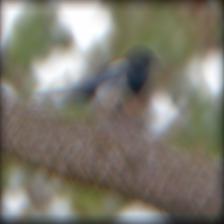}\end{subfigure}\begin{subfigure}{0.065\linewidth}\centering\includegraphics[width=\linewidth]{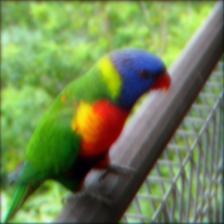}\end{subfigure}\begin{subfigure}{0.065\linewidth}\centering\includegraphics[width=\linewidth]{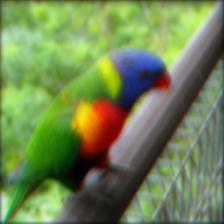}\end{subfigure}\begin{subfigure}{0.065\linewidth}\centering\includegraphics[width=\linewidth]{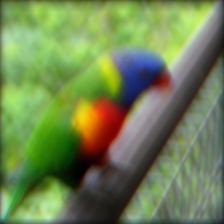}\end{subfigure}\begin{subfigure}{0.065\linewidth}\centering\includegraphics[width=\linewidth]{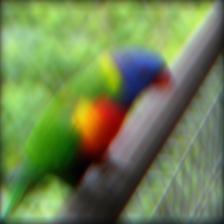}\end{subfigure}\begin{subfigure}{0.065\linewidth}\centering\includegraphics[width=\linewidth]{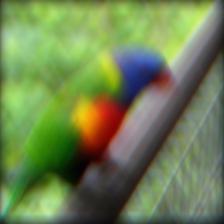}\end{subfigure}\\\begin{subfigure}{0.065\linewidth}\centering\includegraphics[width=\linewidth]{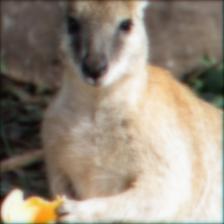}\end{subfigure}\begin{subfigure}{0.065\linewidth}\centering\includegraphics[width=\linewidth]{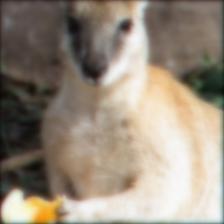}\end{subfigure}\begin{subfigure}{0.065\linewidth}\centering\includegraphics[width=\linewidth]{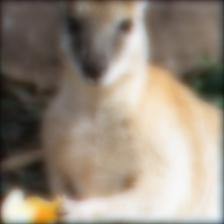}\end{subfigure}\begin{subfigure}{0.065\linewidth}\centering\includegraphics[width=\linewidth]{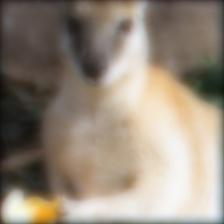}\end{subfigure}\begin{subfigure}{0.065\linewidth}\centering\includegraphics[width=\linewidth]{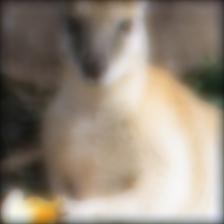}\end{subfigure}\begin{subfigure}{0.065\linewidth}\centering\includegraphics[width=\linewidth]{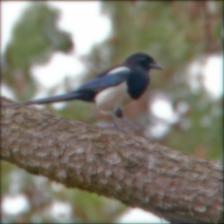}\end{subfigure}\begin{subfigure}{0.065\linewidth}\centering\includegraphics[width=\linewidth]{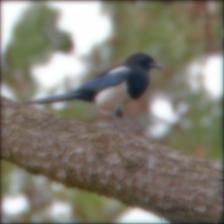}\end{subfigure}\begin{subfigure}{0.065\linewidth}\centering\includegraphics[width=\linewidth]{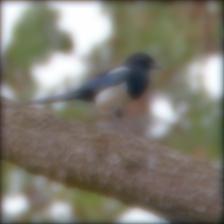}\end{subfigure}\begin{subfigure}{0.065\linewidth}\centering\includegraphics[width=\linewidth]{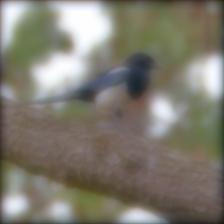}\end{subfigure}\begin{subfigure}{0.065\linewidth}\centering\includegraphics[width=\linewidth]{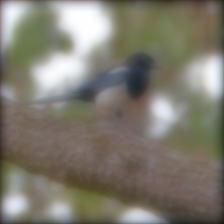}\end{subfigure}\begin{subfigure}{0.065\linewidth}\centering\includegraphics[width=\linewidth]{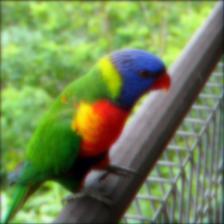}\end{subfigure}\begin{subfigure}{0.065\linewidth}\centering\includegraphics[width=\linewidth]{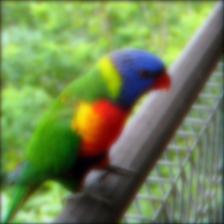}\end{subfigure}\begin{subfigure}{0.065\linewidth}\centering\includegraphics[width=\linewidth]{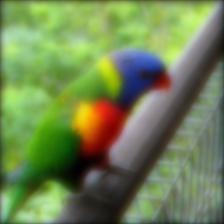}\end{subfigure}\begin{subfigure}{0.065\linewidth}\centering\includegraphics[width=\linewidth]{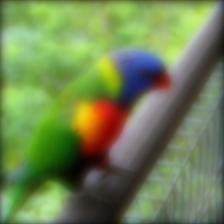}\end{subfigure}\begin{subfigure}{0.065\linewidth}\centering\includegraphics[width=\linewidth]{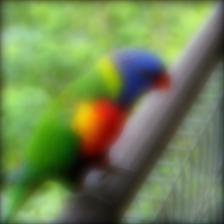}\end{subfigure}\\\begin{subfigure}{0.065\linewidth}\centering\includegraphics[width=\linewidth]{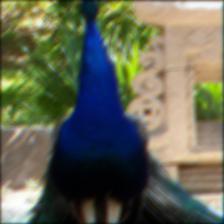}\end{subfigure}\begin{subfigure}{0.065\linewidth}\centering\includegraphics[width=\linewidth]{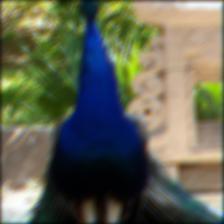}\end{subfigure}\begin{subfigure}{0.065\linewidth}\centering\includegraphics[width=\linewidth]{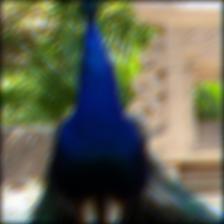}\end{subfigure}\begin{subfigure}{0.065\linewidth}\centering\includegraphics[width=\linewidth]{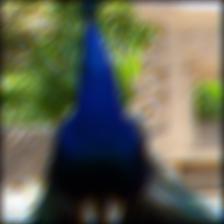}\end{subfigure}\begin{subfigure}{0.065\linewidth}\centering\includegraphics[width=\linewidth]{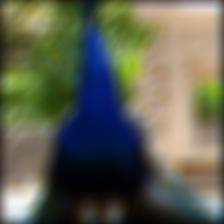}\end{subfigure}\begin{subfigure}{0.065\linewidth}\centering\includegraphics[width=\linewidth]{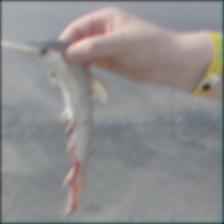}\end{subfigure}\begin{subfigure}{0.065\linewidth}\centering\includegraphics[width=\linewidth]{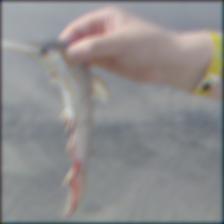}\end{subfigure}\begin{subfigure}{0.065\linewidth}\centering\includegraphics[width=\linewidth]{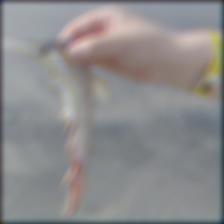}\end{subfigure}\begin{subfigure}{0.065\linewidth}\centering\includegraphics[width=\linewidth]{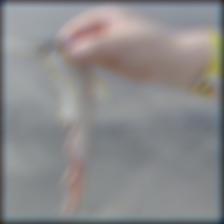}\end{subfigure}\begin{subfigure}{0.065\linewidth}\centering\includegraphics[width=\linewidth]{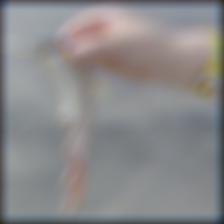}\end{subfigure}\begin{subfigure}{0.065\linewidth}\centering\includegraphics[width=\linewidth]{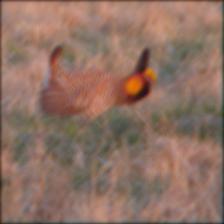}\end{subfigure}\begin{subfigure}{0.065\linewidth}\centering\includegraphics[width=\linewidth]{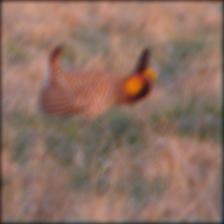}\end{subfigure}\begin{subfigure}{0.065\linewidth}\centering\includegraphics[width=\linewidth]{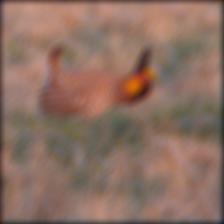}\end{subfigure}\begin{subfigure}{0.065\linewidth}\centering\includegraphics[width=\linewidth]{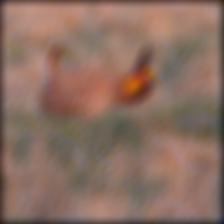}\end{subfigure}\begin{subfigure}{0.065\linewidth}\centering\includegraphics[width=\linewidth]{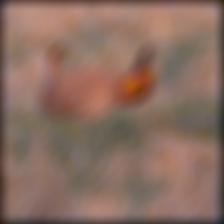}\end{subfigure}\\\begin{subfigure}{0.065\linewidth}\centering\includegraphics[width=\linewidth]{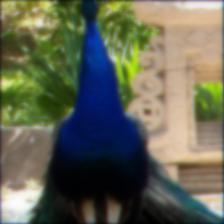}\end{subfigure}\begin{subfigure}{0.065\linewidth}\centering\includegraphics[width=\linewidth]{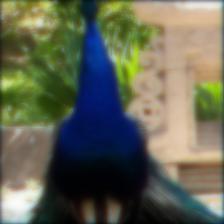}\end{subfigure}\begin{subfigure}{0.065\linewidth}\centering\includegraphics[width=\linewidth]{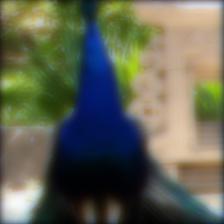}\end{subfigure}\begin{subfigure}{0.065\linewidth}\centering\includegraphics[width=\linewidth]{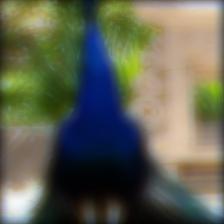}\end{subfigure}\begin{subfigure}{0.065\linewidth}\centering\includegraphics[width=\linewidth]{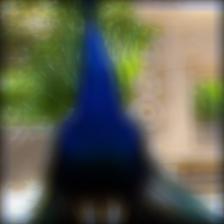}\end{subfigure}\begin{subfigure}{0.065\linewidth}\centering\includegraphics[width=\linewidth]{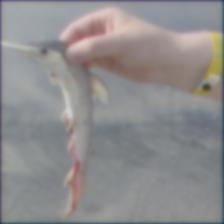}\end{subfigure}\begin{subfigure}{0.065\linewidth}\centering\includegraphics[width=\linewidth]{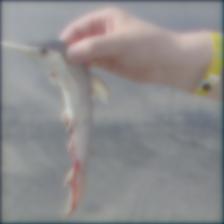}\end{subfigure}\begin{subfigure}{0.065\linewidth}\centering\includegraphics[width=\linewidth]{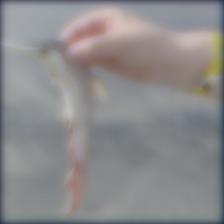}\end{subfigure}\begin{subfigure}{0.065\linewidth}\centering\includegraphics[width=\linewidth]{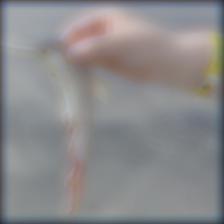}\end{subfigure}\begin{subfigure}{0.065\linewidth}\centering\includegraphics[width=\linewidth]{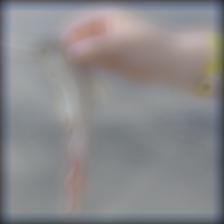}\end{subfigure}\begin{subfigure}{0.065\linewidth}\centering\includegraphics[width=\linewidth]{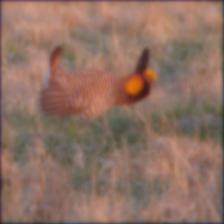}\end{subfigure}\begin{subfigure}{0.065\linewidth}\centering\includegraphics[width=\linewidth]{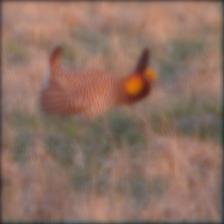}\end{subfigure}\begin{subfigure}{0.065\linewidth}\centering\includegraphics[width=\linewidth]{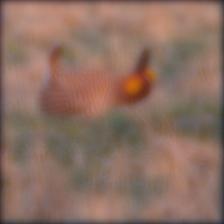}\end{subfigure}\begin{subfigure}{0.065\linewidth}\centering\includegraphics[width=\linewidth]{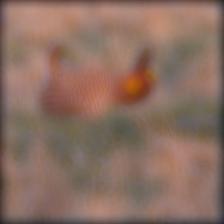}\end{subfigure}\begin{subfigure}{0.065\linewidth}\centering\includegraphics[width=\linewidth]{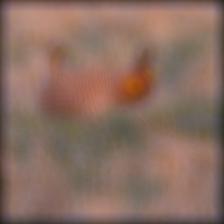}\end{subfigure}\\\begin{subfigure}{0.065\linewidth}\centering\includegraphics[width=\linewidth]{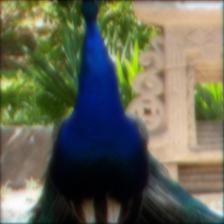}\end{subfigure}\begin{subfigure}{0.065\linewidth}\centering\includegraphics[width=\linewidth]{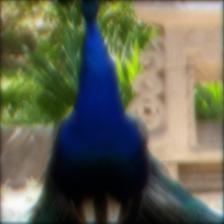}\end{subfigure}\begin{subfigure}{0.065\linewidth}\centering\includegraphics[width=\linewidth]{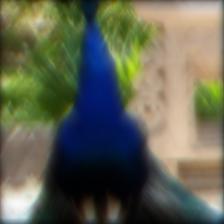}\end{subfigure}\begin{subfigure}{0.065\linewidth}\centering\includegraphics[width=\linewidth]{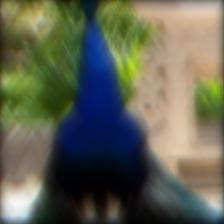}\end{subfigure}\begin{subfigure}{0.065\linewidth}\centering\includegraphics[width=\linewidth]{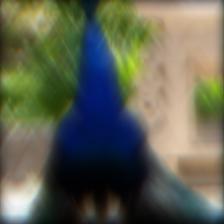}\end{subfigure}\begin{subfigure}{0.065\linewidth}\centering\includegraphics[width=\linewidth]{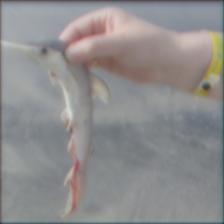}\end{subfigure}\begin{subfigure}{0.065\linewidth}\centering\includegraphics[width=\linewidth]{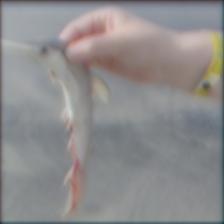}\end{subfigure}\begin{subfigure}{0.065\linewidth}\centering\includegraphics[width=\linewidth]{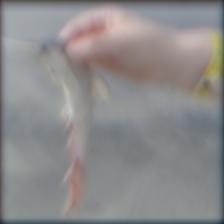}\end{subfigure}\begin{subfigure}{0.065\linewidth}\centering\includegraphics[width=\linewidth]{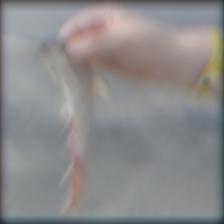}\end{subfigure}\begin{subfigure}{0.065\linewidth}\centering\includegraphics[width=\linewidth]{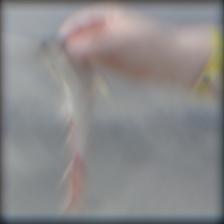}\end{subfigure}\begin{subfigure}{0.065\linewidth}\centering\includegraphics[width=\linewidth]{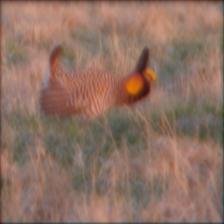}\end{subfigure}\begin{subfigure}{0.065\linewidth}\centering\includegraphics[width=\linewidth]{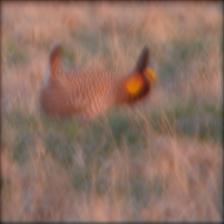}\end{subfigure}\begin{subfigure}{0.065\linewidth}\centering\includegraphics[width=\linewidth]{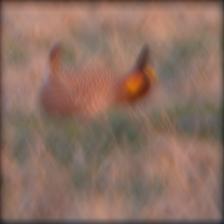}\end{subfigure}\begin{subfigure}{0.065\linewidth}\centering\includegraphics[width=\linewidth]{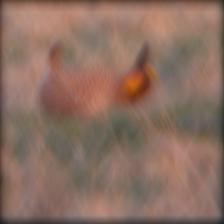}\end{subfigure}\begin{subfigure}{0.065\linewidth}\centering\includegraphics[width=\linewidth]{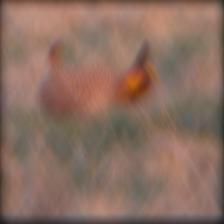}\end{subfigure}\\\begin{subfigure}{0.065\linewidth}\centering\includegraphics[width=\linewidth]{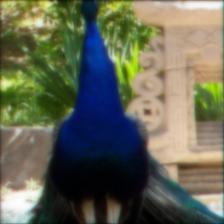}\end{subfigure}\begin{subfigure}{0.065\linewidth}\centering\includegraphics[width=\linewidth]{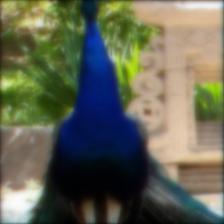}\end{subfigure}\begin{subfigure}{0.065\linewidth}\centering\includegraphics[width=\linewidth]{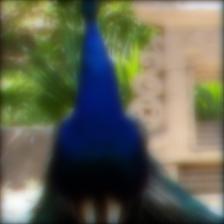}\end{subfigure}\begin{subfigure}{0.065\linewidth}\centering\includegraphics[width=\linewidth]{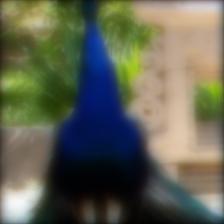}\end{subfigure}\begin{subfigure}{0.065\linewidth}\centering\includegraphics[width=\linewidth]{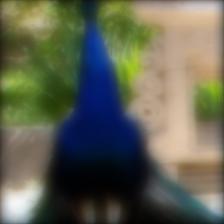}\end{subfigure}\begin{subfigure}{0.065\linewidth}\centering\includegraphics[width=\linewidth]{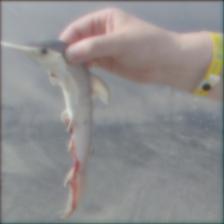}\end{subfigure}\begin{subfigure}{0.065\linewidth}\centering\includegraphics[width=\linewidth]{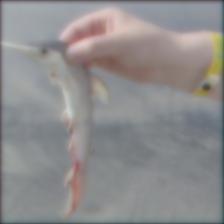}\end{subfigure}\begin{subfigure}{0.065\linewidth}\centering\includegraphics[width=\linewidth]{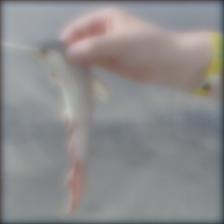}\end{subfigure}\begin{subfigure}{0.065\linewidth}\centering\includegraphics[width=\linewidth]{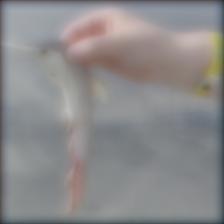}\end{subfigure}\begin{subfigure}{0.065\linewidth}\centering\includegraphics[width=\linewidth]{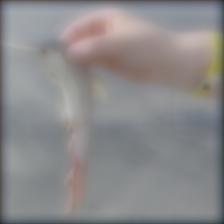}\end{subfigure}\begin{subfigure}{0.065\linewidth}\centering\includegraphics[width=\linewidth]{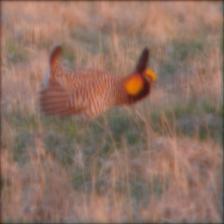}\end{subfigure}\begin{subfigure}{0.065\linewidth}\centering\includegraphics[width=\linewidth]{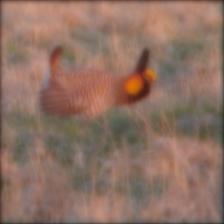}\end{subfigure}\begin{subfigure}{0.065\linewidth}\centering\includegraphics[width=\linewidth]{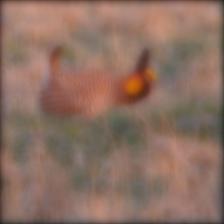}\end{subfigure}\begin{subfigure}{0.065\linewidth}\centering\includegraphics[width=\linewidth]{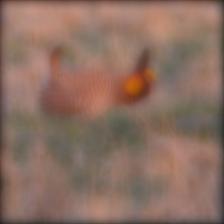}\end{subfigure}\begin{subfigure}{0.065\linewidth}\centering\includegraphics[width=\linewidth]{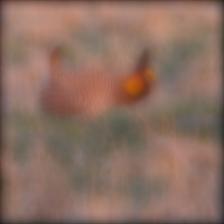}\end{subfigure}\caption{Image examples from ImageNet-1k OpticsBench.}\label{fig:image_examples}\end{figure*}
\FloatBarrier
\section{OpticsBenchRG}
\label{app:opticsbenchRG}
This appendix includes exemplary evaluations on OpticsBenchRG on ImageNet and ImageNet-100. The benchmark consists of images blurred with the reddish and greenish optical kernels from Fig.~\ref{fig:kernels_cooke_rg_iccv} in App.~\ref{app:kernels}.
Tab.~\ref{tab:tab:imagenet100_avg_absolute_rg} lists the average accuracies for all OpticsBenchRG corruptions. OpticsAugment (ours) achieves constantly better results. For selected DNNs Fig.~\ref{fig:app_rg_optics_augment2} and Fig.~\ref{fig:app_rg_optics_augment} show the accuracies separately for each corruption and w/wo OpticsAugment (out of domain kernels compared to OpticsBenchRG) training. 

\begin{figure*}[h]
\begin{subfigure}{0.95\linewidth}
    \centering    
    \includegraphics[width=\linewidth]{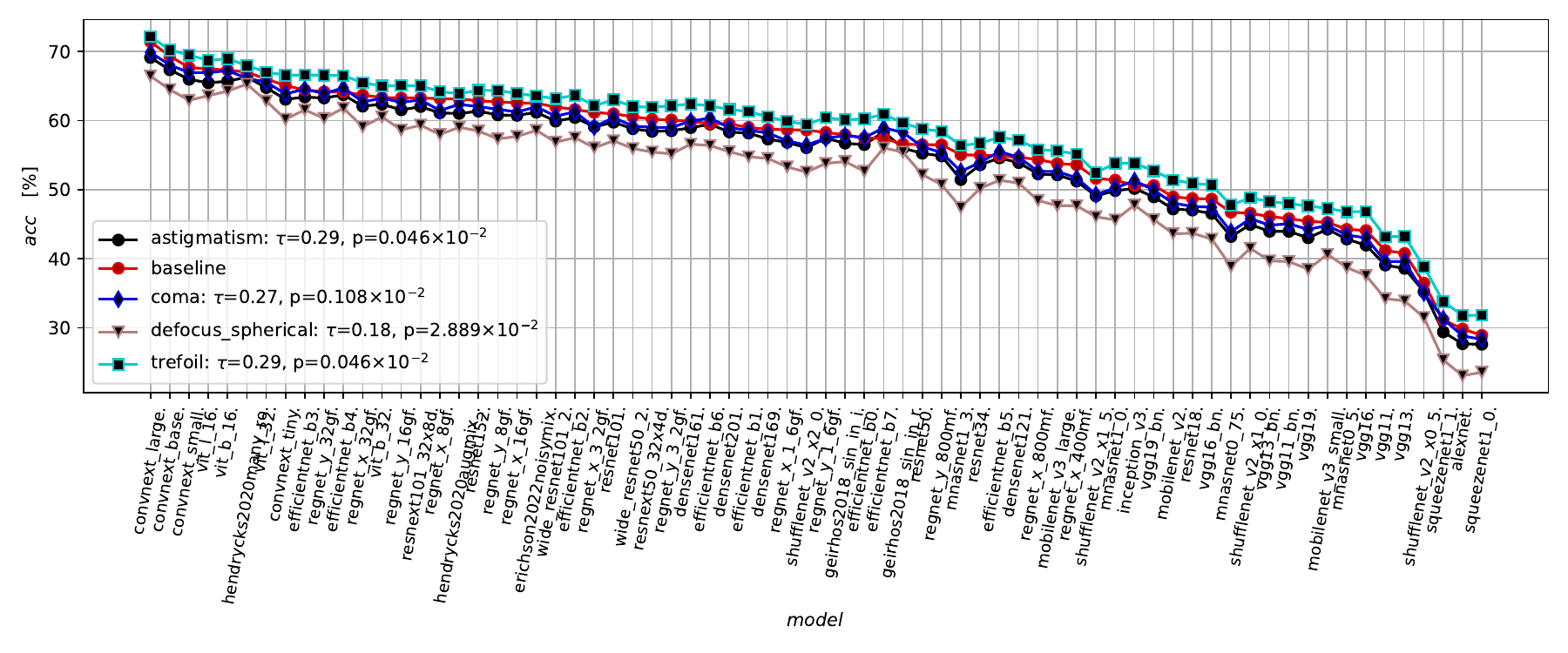}
    \caption{}
    \label{fig:ranking_sev1}
\end{subfigure}
\begin{subfigure}{0.95\linewidth}
    \centering    
    \includegraphics[width=\linewidth]{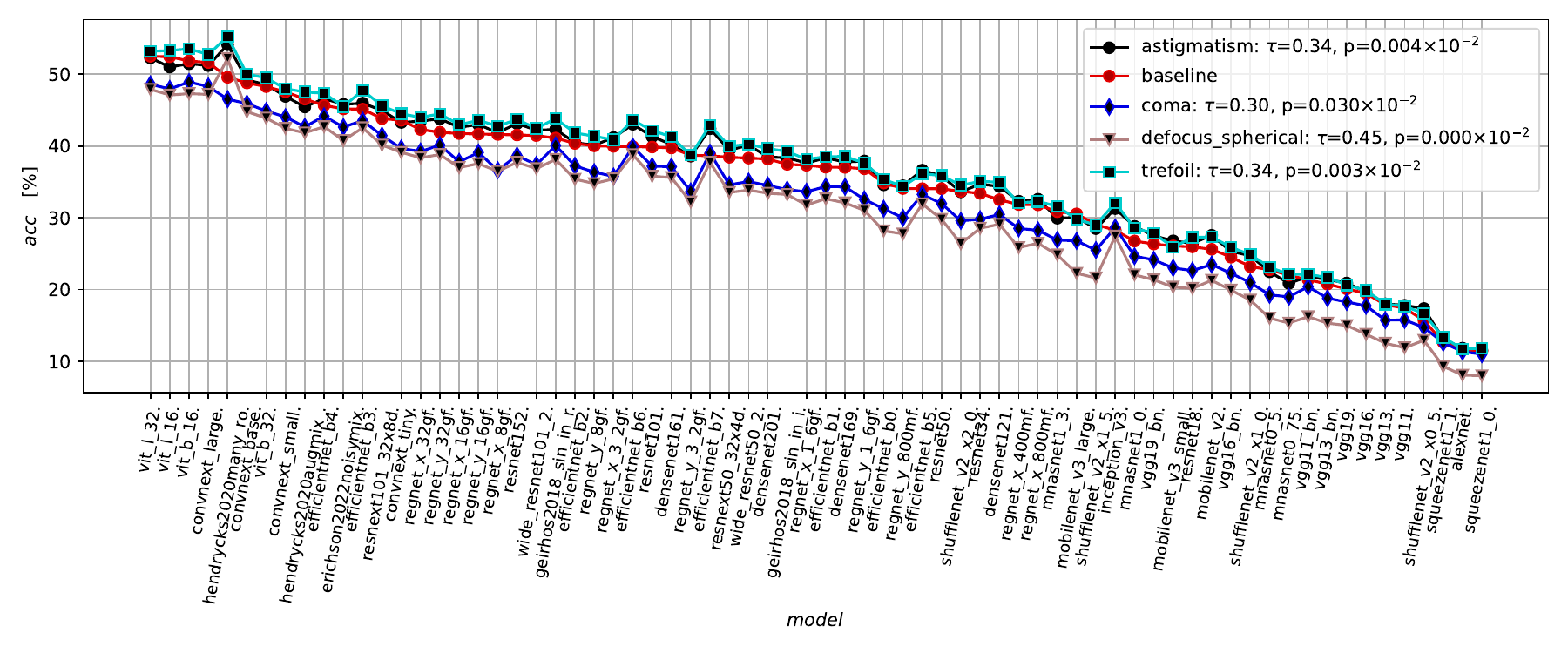}
    \caption{}
    \label{fig:ranking_sev3}
\end{subfigure}
\begin{subfigure}{0.95\linewidth}
    \centering    
    \includegraphics[width=\linewidth]{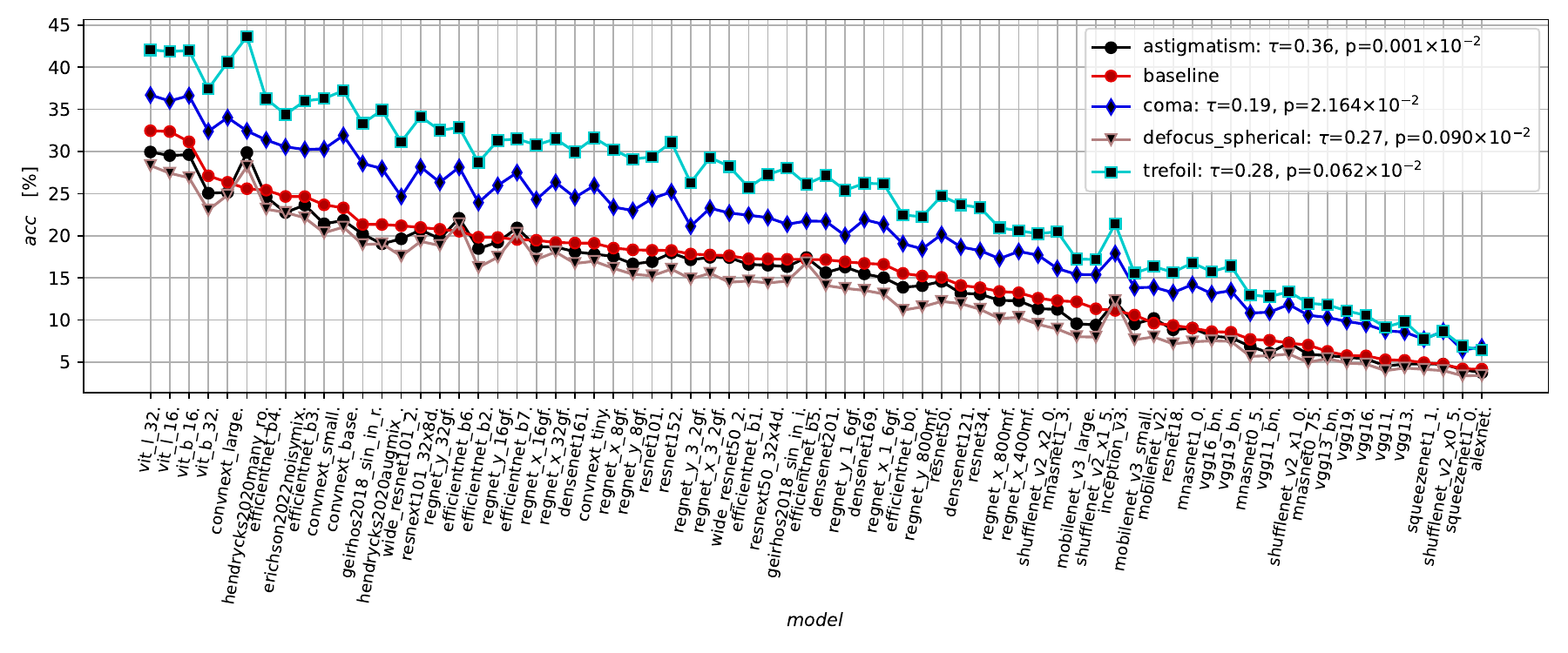}
    \caption{}
    \label{fig:ranking_sev5}
\end{subfigure}
    \caption{Ranking comparison of baseline and all corruptions for severities 1,3 and 5 (a-c).}
    \label{fig:ranking_sev135_rg}
\end{figure*}

Additionally, ranking comparisons on ImageNet and OpticsBenchRG for the 70 pretrained DNNs (5 from RobustBench leaderboard and 65 from PyTorch) are shown for selected severities in Fig.~\ref{fig:ranking_sev135_rg} for comparison with App.~\ref{app:ranking}.
%
\begin{table}[h]\centering\footnotesize\begin{tabular}{@{}llllll@{}}Model & 1 & 2 & 3 & 4 & 5 \\ \hline \\DenseNet \textbf{(ours)} & \textbf{64.78} & \textbf{59.41} & \textbf{47.75} & \textbf{36.41} & \textbf{29.66} \\ DenseNet & 54.53 & 45.09 & 31.37 & 22.73 & 18.43\\EfficientNet \textbf{(ours)} & \textbf{60.55} & \textbf{54.23} & \textbf{42.50} & \textbf{32.41} & \textbf{26.13} \\ EfficientNet & 53.38 & 43.99 & 30.91 & 22.20 & 17.61\\MobileNet \textbf{(ours)} & \textbf{56.55} & \textbf{50.49} & \textbf{36.58} & \textbf{25.95} & \textbf{20.91} \\ MobileNet & 49.71 & 39.56 & 25.60 & 18.81 & 15.55\\ResNet101 \textbf{(ours)} & \textbf{67.95} & \textbf{63.90} & \textbf{54.34} & \textbf{43.11} & \textbf{34.75} \\ ResNet101 & 60.42 & 52.44 & 41.21 & 33.31 & 27.85\\ResNeXt50 \textbf{(ours)} & \textbf{59.59} & \textbf{54.50} & \textbf{43.57} & \textbf{31.91} & \textbf{25.04} \\ ResNeXt50 & 47.62 & 37.87 & 25.66 & 18.61 & 15.29\\\end{tabular}\caption{Accuracies w/wo OpticsAugment evaluated on ImageNet-100 OpticsBenchRG. Average over all corruptions. Even when changing the optics corruption (\ie in an out of domain setting), the proposed augmentation consistently leads to higher classification accuracies.}\label{tab:tab:imagenet100_avg_absolute_rg}\end{table}
\FloatBarrier

\newcommand{\sizeRGbench}{0.9\linewidth}

\begin{figure}[h]
    \centering
    \includegraphics[width=\sizeRGbench]{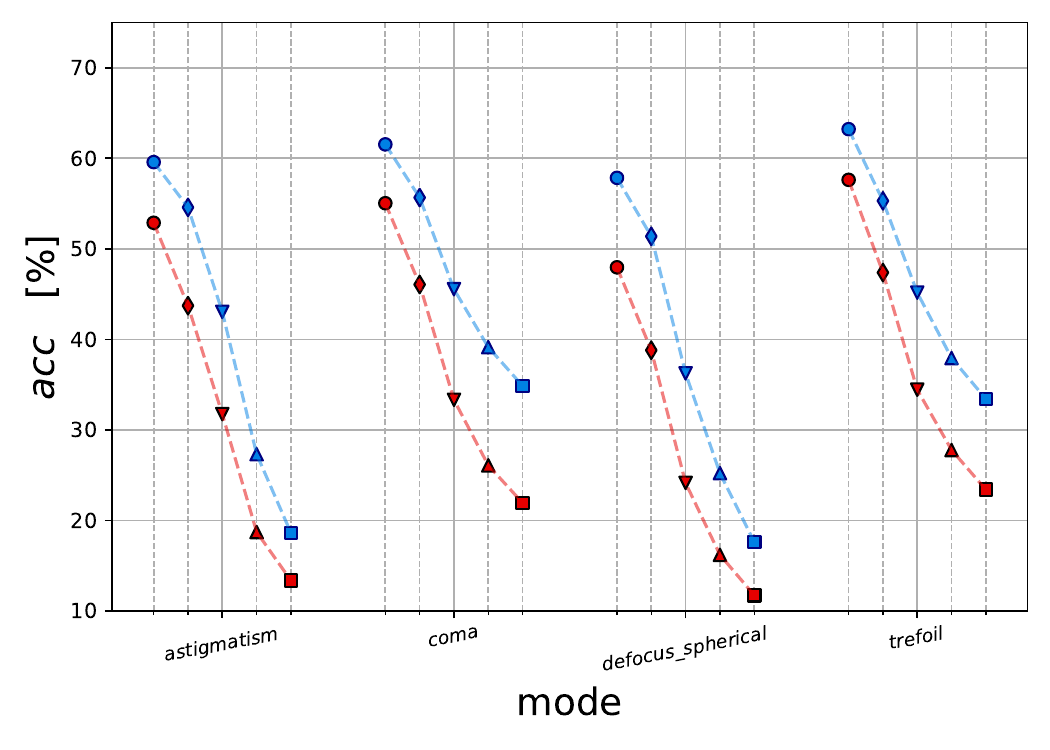}
    \caption{Accuracy evaluated on \textbf{OpticsBenchRG}-ImageNet-100 for EfficientNet w/wo OpticsAugment training and all severities 1-5 (circle to square markers) at each corruption. Although the exact kernels haven't been visible during training, still 
    \textbf{OpticsAugment (blue) improves} accuracy compared to the conventionally trained DNN (red).}
    \label{fig:app_rg_optics_augment2}
\end{figure}
\begin{figure}[h]
    \centering
    \begin{subfigure}{\sizeRGbench}
    \includegraphics[width=\linewidth]{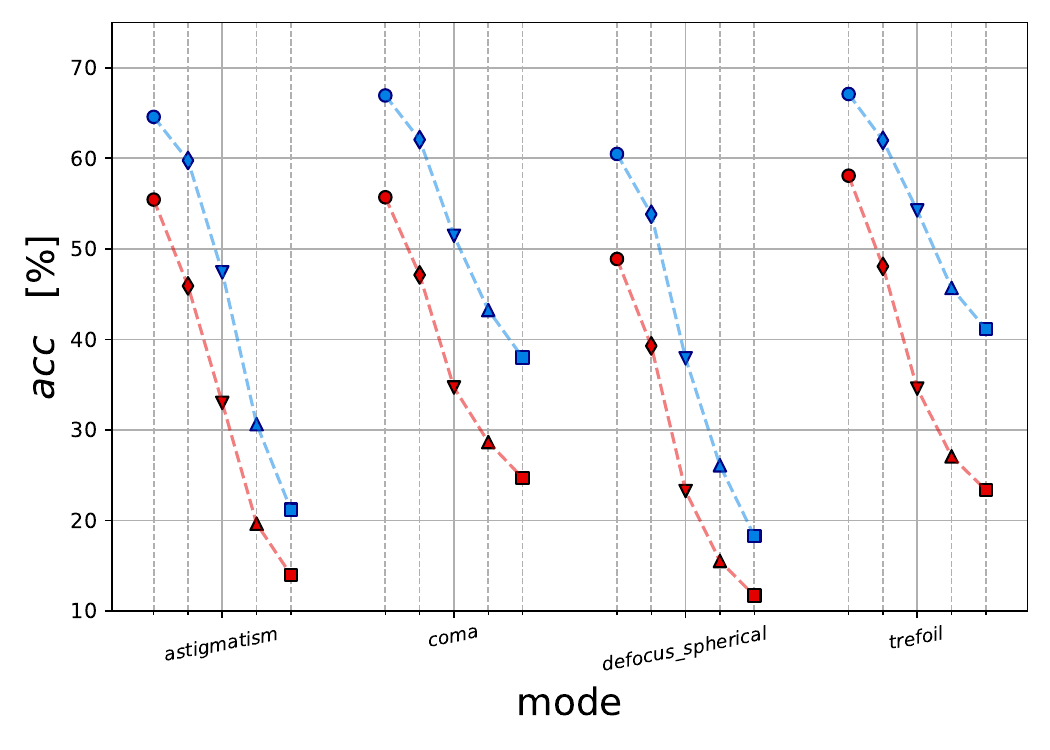}
    \caption{DenseNet161}
    \label{fig:app_rg_imagenet100_densenet161}
    \end{subfigure}   
    \begin{subfigure}{\sizeRGbench}
    \includegraphics[width=\linewidth]{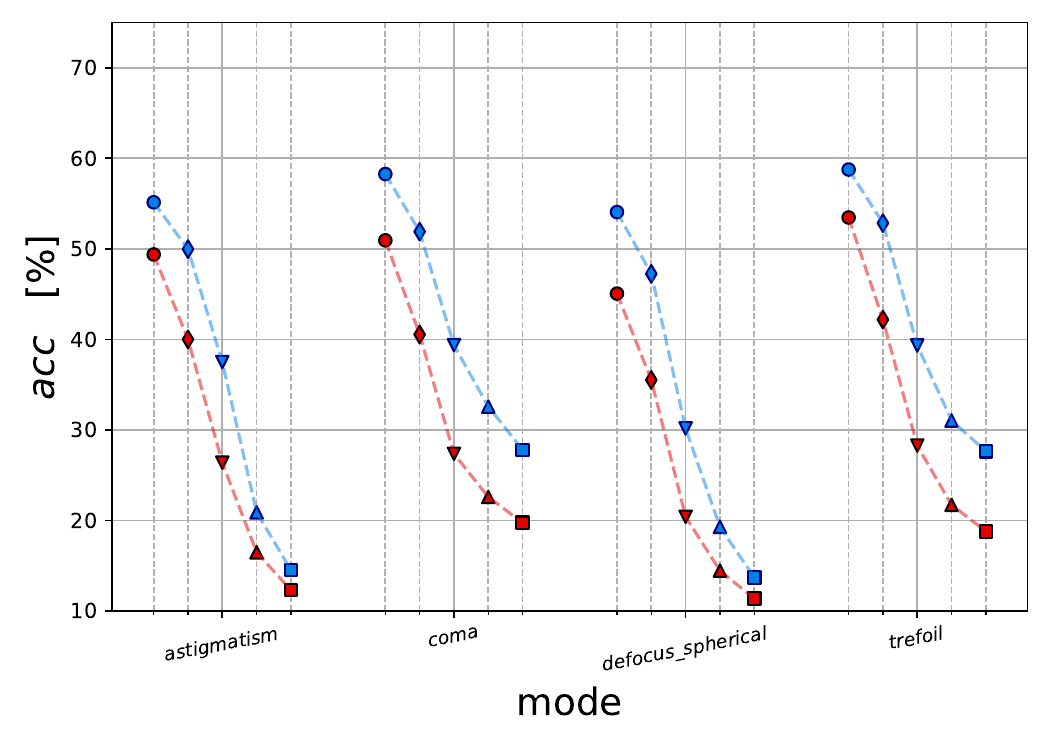}
    \caption{MobileNet}
    \label{fig:app_rg_imagenet100_mobilenet_v3_large}
    \end{subfigure}   
    \begin{subfigure}{\sizeRGbench}
    \includegraphics[width=\linewidth]{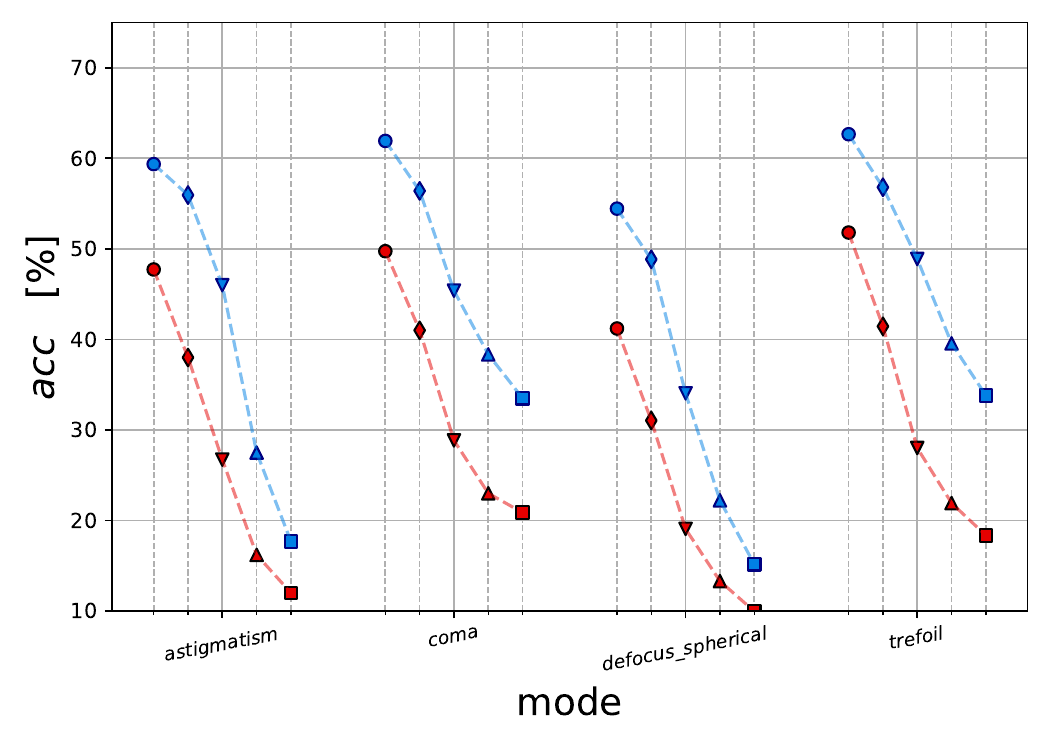}
    \caption{ResNeXt50}
    \label{fig:app_rg_imagenet100_resnext50}
    \end{subfigure}
    \caption{\textbf{OpticsBenchRG}-ImageNet-100 for DNNs w/wo OpticsAugment training and all severities 1-5 (circle to square markers). Although the exact kernels haven't been visible during training, \textbf{OpticsAugment (blue) improves} accuracy compared to the conventionally trained DNN (red):  (a) DenseNet161, (b) MobileNet and (c) ResNeXt50.}
    \label{fig:app_rg_optics_augment}
\end{figure}
\FloatBarrier

\FloatBarrier
\section{Implementation and additional analysis}
\label{app:additional_analysis}
Here, we include additional analysis on various datasets and give further implementation details.

\subsection{Implementation details}
We also provide code for a more detailed insight into the structure and run experiments. The code is organized into two main parts: Benchmark (OpticsBench and OpticsBenchRG) and training (OpticsAugment, variations and baseline training without additional augmentation). The whole code submission uses Python $3.6+$ for downward compatibility for training on a high performance cluster (HPC) and Python $\geq 3.8$ for the benchmark and pretrained variants. The latter allows to use a more recent pytorch and torchvision version to include VisionTransformer networks and other types.  

Training is run on a HPC using slurm job scheduling and V100 GPUs. 
Training on ImageNet-100 required single V100 GPUs, but also multi GPU training had been utilized. The training for 90 epochs takes about one day for smaller DNNs such as EfficientNet and two days for larger DNNs such as ResNet101. The exact hyperparameter settings can be found in the code submission in recipes, however the particular batch size had been adjusted to increase the speed. For OpticsAugment training $\alpha=1.0$ and $severity=3$ is set. For training the ImageNet-100 train split is split again into a validation and train split to avoid any overlap with the original validation dataset, which is used as test dataset.

\subsection{Additional analysis}

First, an evaluation of adversarial robustness for different ImageNet-100 trained DNNs is presented in Tab.~\ref{tab:app:adversarial_robustness}. Ours uses the OpticsAugment training scheme and is compared to a conventionally trained DNN on the same dataset. To allow for evaluation on ImageNet's validation dataset, the train set is split into a validation and train split. 
To lower the computational resources needed for the computation, 1000 validation images are randomly selected and saved as test dataset for adversarial robustness. The attacks had been lowered to $l_2$ and $\epsilon=4/255$ to avoid exclusively successful attacks. Still, with this setting no clear trend can be observed, the overall robustness to the attacks is low, but on average OpticsAugment does not lower adversarial robustness compared to a conventionally trained DNN. The evaluation for each DNN takes several hours on a NVIDIA GeForce 3080Ti 12GB VRAM GPU. 

\begin{table}[]
    \centering
    \begin{tabular}{l l l l}
       DNN  & Robust Acc & APGD-CE & APGD-DLR \\
       \hline
       DenseNet  & 5.2 & 13.9 & 5.7  \\ 
       DenseNet (ours)  & 6.4 & 14.5 & 6.4   \\ 
       EfficientNet  & 2.1 & 8.8 & 2.4 \\ 
       EfficientNet (ours)  & 1.7 & 8.4 & 1.9 \\ 
       MobileNet  & 1.2 & 6.2 & 1.6 \\ 
       MobileNet (ours)  & 1.8 & 7.6 & 2.2 \\ 
       ResNeXt50  & 1.2 & 11.3 & 1.6 \\ 
       ResNeXt50 (ours)  & 3.1 & 8.9 & 3.6  \\ 
    \end{tabular}
    \caption{Adversarial robustness in \% to adversarial attacks using APGD-CE and APGD-DLR from AutoAttack~\cite{croce_reliable_2020}, $l_2$ and $\epsilon=4/255$, batch size $32$ and $5$ restarts on $1000$ validation images of ImageNet-100.}
    \label{tab:app:adversarial_robustness}
\end{table}

\begin{figure*}
\centering
\begin{subfigure}{0.6\linewidth}
    \centering
    \includegraphics[width=\linewidth]{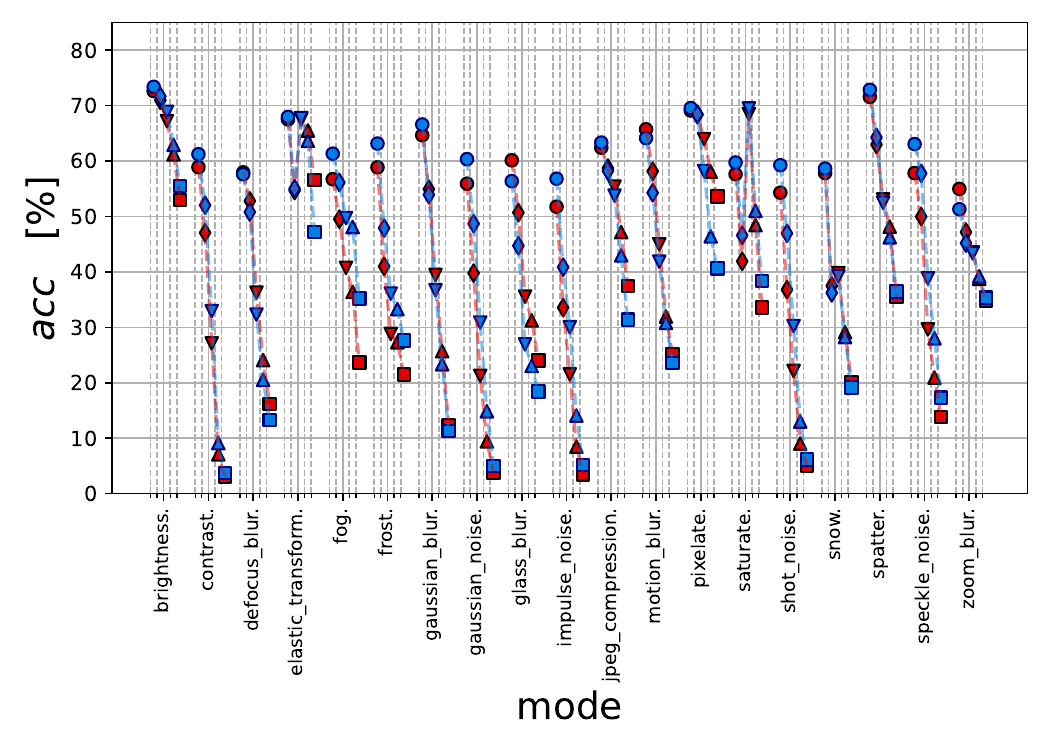}
    \caption{MobileNet}
\end{subfigure}
\begin{subfigure}{0.6\linewidth}
    \centering
    \includegraphics[width=\linewidth]{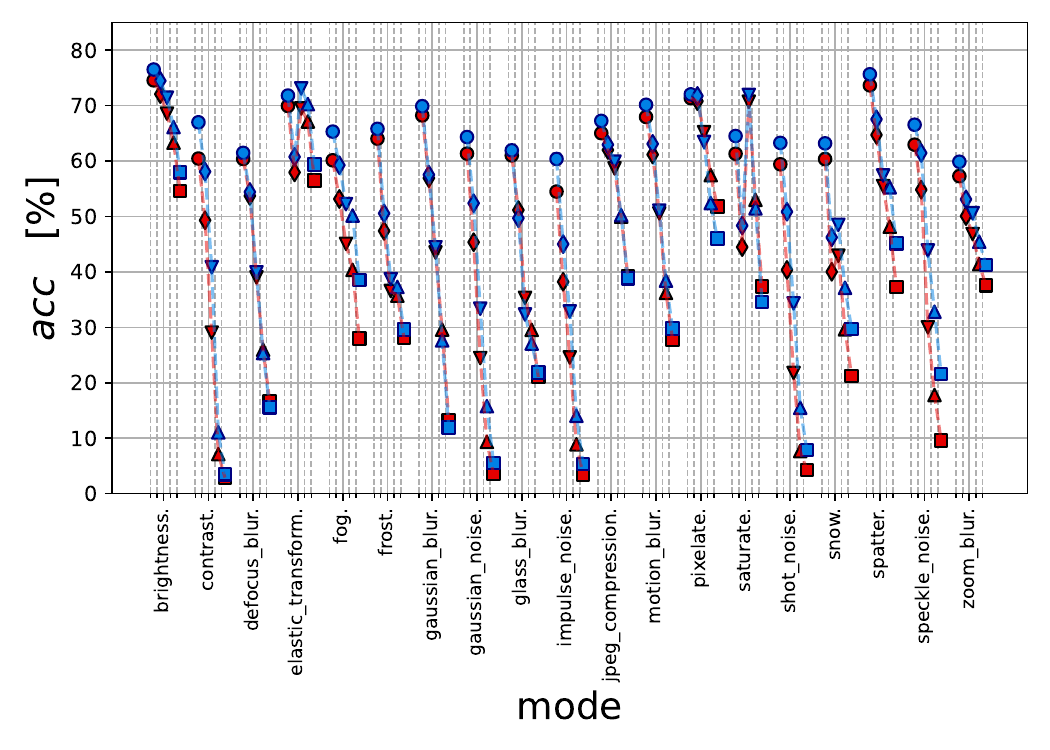}
    \caption{EfficientNet}
\end{subfigure}
    \caption{Pipelining of AugMix~\cite{hendrycks_augmix_2020} and OpticsAugment: Blue represents now a cascaded application of AugMix and OpticsAugment. Red represents the OpticsAugment trained version from  Fig.~\ref{fig:app_imagenet100c_mobilenet_v3_large} and~\ref{fig:app_imagenet100c_efficientnet_b0} respectively. (a) MobileNet and (b) EfficientNet. Evaluated on 2D common corruptions on ImageNet-100-C~\cite{hendrycks_benchmarking_2019}.}
    \label{fig:pipelining_common}
\end{figure*}

Additionally, in Tab.~\ref{tab:tab:imagenet100_corruptions_MobileNet_pipelined} and~\ref{tab:tab:imagenet100_corruptions_EfficientNet_pipelined} the results for a pipelining of AugMix and OpticsAugment during ImageNet-100 training are listed for an evaluation on OpticsBenchRG. EfficientNet and MobileNet are either trained with only OpticsAugment (red) or with OpticsAugment and AugMix~\cite{hendrycks_augmix_2020}. Fig.~\ref{fig:pipelining_common} visualizes the same DNNs on 2D common corruptions. This shows another application scenario of OpticsAugment.

\begin{table*}[h]\centering\begin{tabular}{llll|lll|lll|lll|lll}&\multicolumn{3}{c}{\scriptsize{1}}&\multicolumn{3}{c}{\scriptsize{2}}&\multicolumn{3}{c}{\scriptsize{3}}&\multicolumn{3}{c}{\scriptsize{4}}&\multicolumn{3}{c}{\scriptsize{5}}\\\tiny{Corruption} & \tiny{ours \& AugMix} & \tiny{ours} & \tiny{$\Delta$} & \tiny{ours \& AugMix} & \tiny{ours} & \tiny{$\Delta$} & \tiny{ours \& AugMix} & \tiny{ours} & \tiny{$\Delta$} & \tiny{ours \& AugMix} & \tiny{ours} & \tiny{$\Delta$} & \tiny{ours \& AugMix} & \tiny{ours} & \tiny{$\Delta$}\\\hline\tiny{astigmatism} & \tiny{59.52} & \textbf{\tiny{59.58}} & \tiny{0.06} & \tiny{53.00} & \textbf{\tiny{54.58}} & \tiny{1.58} & \tiny{38.82} & \textbf{\tiny{43.04}} & \tiny{4.22} & \tiny{23.38} & \textbf{\tiny{27.32}} & \tiny{3.94} & \tiny{16.06} & \textbf{\tiny{18.62}} & \tiny{2.56}\\\tiny{coma} & \textbf{\tiny{64.96}} & \tiny{61.54} & \tiny{-3.42} & \textbf{\tiny{57.74}} & \tiny{55.66} & \tiny{-2.08} & \textbf{\tiny{46.32}} & \tiny{45.54} & \tiny{-0.78} & \tiny{37.70} & \textbf{\tiny{39.14}} & \tiny{1.44} & \tiny{32.50} & \textbf{\tiny{34.86}} & \tiny{2.36}\\\tiny{defocus\_spherical} & \textbf{\tiny{58.00}} & \tiny{57.84} & \tiny{-0.16} & \tiny{50.38} & \textbf{\tiny{51.38}} & \tiny{1.00} & \tiny{33.14} & \textbf{\tiny{36.24}} & \tiny{3.10} & \tiny{22.64} & \textbf{\tiny{25.22}} & \tiny{2.58} & \textbf{\tiny{17.98}} & \tiny{17.62} & \tiny{-0.36}\\\tiny{trefoil} & \textbf{\tiny{65.30}} & \tiny{63.22} & \tiny{-2.08} & \textbf{\tiny{57.42}} & \tiny{55.30} & \tiny{-2.12} & \tiny{45.12} & \textbf{\tiny{45.16}} & \tiny{0.04} & \textbf{\tiny{38.30}} & \tiny{37.94} & \tiny{-0.36} & \textbf{\tiny{34.82}} & \tiny{33.42} & \tiny{-1.40}\\\tiny{{{$\Sigma$}}} & \textbf{\tiny{61.95}} & \tiny{60.55} & \tiny{-1.40} & \textbf{\tiny{54.64}} & \tiny{54.23} & \tiny{-0.40} & \tiny{40.85} & \textbf{\tiny{42.50}} & \tiny{1.64} & \tiny{30.50} & \textbf{\tiny{32.41}} & \tiny{1.90} & \tiny{25.34} & \textbf{\tiny{26.13}} & \tiny{0.79}\end{tabular}\caption{Accuracies for EfficientNet \& OpticsAugment w/wo AugMix. Evaluated on ImageNet-100 OpticsBenchRG.}\label{tab:tab:imagenet100_corruptions_EfficientNet_pipelined}\end{table*}
\begin{table*}[h]\centering\begin{tabular}{llll|lll|lll|lll|lll}&\multicolumn{3}{c}{\scriptsize{1}}&\multicolumn{3}{c}{\scriptsize{2}}&\multicolumn{3}{c}{\scriptsize{3}}&\multicolumn{3}{c}{\scriptsize{4}}&\multicolumn{3}{c}{\scriptsize{5}}\\\tiny{Corruption} & \tiny{ours \& AugMix} & \tiny{ours} & \tiny{$\Delta$} & \tiny{ours \& AugMix} & \tiny{ours} & \tiny{$\Delta$} & \tiny{ours \& AugMix} & \tiny{ours} & \tiny{$\Delta$} & \tiny{ours \& AugMix} & \tiny{ours} & \tiny{$\Delta$} & \tiny{ours \& AugMix} & \tiny{ours} & \tiny{$\Delta$}\\\hline\tiny{astigmatism} & \textbf{\tiny{56.14}} & \tiny{55.14} & \tiny{-1.00} & \tiny{48.26} & \textbf{\tiny{49.96}} & \tiny{1.70} & \tiny{32.64} & \textbf{\tiny{37.48}} & \tiny{4.84} & \tiny{18.72} & \textbf{\tiny{20.90}} & \tiny{2.18} & \tiny{12.98} & \textbf{\tiny{14.52}} & \tiny{1.54}\\\tiny{coma} & \textbf{\tiny{60.12}} & \tiny{58.26} & \tiny{-1.86} & \textbf{\tiny{53.12}} & \tiny{51.90} & \tiny{-1.22} & \textbf{\tiny{39.46}} & \tiny{39.36} & \tiny{-0.10} & \tiny{32.08} & \textbf{\tiny{32.58}} & \tiny{0.50} & \tiny{26.50} & \textbf{\tiny{27.80}} & \tiny{1.30}\\\tiny{defocus\_spherical} & \textbf{\tiny{55.50}} & \tiny{54.06} & \tiny{-1.44} & \tiny{47.14} & \textbf{\tiny{47.24}} & \tiny{0.10} & \tiny{29.12} & \textbf{\tiny{30.16}} & \tiny{1.04} & \textbf{\tiny{20.70}} & \tiny{19.30} & \tiny{-1.40} & \textbf{\tiny{15.20}} & \tiny{13.68} & \tiny{-1.52}\\\tiny{trefoil} & \textbf{\tiny{61.48}} & \tiny{58.76} & \tiny{-2.72} & \textbf{\tiny{53.54}} & \tiny{52.84} & \tiny{-0.70} & \textbf{\tiny{39.36}} & \tiny{39.32} & \tiny{-0.04} & \textbf{\tiny{32.40}} & \tiny{31.04} & \tiny{-1.36} & \textbf{\tiny{29.38}} & \tiny{27.64} & \tiny{-1.74}\\\tiny{{{$\Sigma$}}} & \textbf{\tiny{58.31}} & \tiny{56.55} & \tiny{-1.76} & \textbf{\tiny{50.51}} & \tiny{50.49} & \tiny{-0.03} & \tiny{35.14} & \textbf{\tiny{36.58}} & \tiny{1.44} & \textbf{\tiny{25.98}} & \tiny{25.95} & \tiny{-0.02} & \textbf{\tiny{21.02}} & \tiny{20.91} & \tiny{-0.10}\end{tabular}\caption{Accuracies for MobileNet \& OpticsAugment w/wo AugMix. Evaluated on ImageNet-100 OpticsBenchRG.}\label{tab:tab:imagenet100_corruptions_MobileNet_pipelined}\end{table*}


\begin{table*}[h]\centering\begin{tabular}{llll|lll|lll|lll|lll}&\multicolumn{3}{c}{\scriptsize{1}}&\multicolumn{3}{c}{\scriptsize{2}}&\multicolumn{3}{c}{\scriptsize{3}}&\multicolumn{3}{c}{\scriptsize{4}}&\multicolumn{3}{c}{\scriptsize{5}}\\\tiny{Corruption} & \tiny{ours \& AugMix} & \tiny{ours} & \tiny{$\Delta$} & \tiny{ours \& AugMix} & \tiny{ours} & \tiny{$\Delta$} & \tiny{ours \& AugMix} & \tiny{ours} & \tiny{$\Delta$} & \tiny{ours \& AugMix} & \tiny{ours} & \tiny{$\Delta$} & \tiny{ours \& AugMix} & \tiny{ours} & \tiny{$\Delta$}\\\hline\tiny{brightness} & \textbf{\tiny{76.52}} & \tiny{74.56} & \tiny{-1.96} & \textbf{\tiny{74.48}} & \tiny{72.08} & \tiny{-2.40} & \textbf{\tiny{71.42}} & \tiny{68.56} & \tiny{-2.86} & \textbf{\tiny{66.14}} & \tiny{63.32} & \tiny{-2.82} & \textbf{\tiny{57.98}} & \tiny{54.60} & \tiny{-3.38}\\\tiny{contrast} & \textbf{\tiny{67.00}} & \tiny{60.46} & \tiny{-6.54} & \textbf{\tiny{58.10}} & \tiny{49.32} & \tiny{-8.78} & \textbf{\tiny{40.84}} & \tiny{29.04} & \tiny{-11.80} & \textbf{\tiny{11.06}} & \tiny{7.14} & \tiny{-3.92} & \textbf{\tiny{3.50}} & \tiny{2.90} & \tiny{-0.60}\\\tiny{defocus\_blur} & \textbf{\tiny{61.48}} & \tiny{60.38} & \tiny{-1.10} & \textbf{\tiny{54.42}} & \tiny{53.72} & \tiny{-0.70} & \textbf{\tiny{39.92}} & \tiny{39.02} & \tiny{-0.90} & \tiny{25.38} & \textbf{\tiny{25.98}} & \tiny{0.60} & \tiny{15.58} & \textbf{\tiny{16.64}} & \tiny{1.06}\\\tiny{elastic\_transform} & \textbf{\tiny{71.82}} & \tiny{69.94} & \tiny{-1.88} & \textbf{\tiny{60.76}} & \tiny{57.96} & \tiny{-2.80} & \textbf{\tiny{73.10}} & \tiny{69.48} & \tiny{-3.62} & \textbf{\tiny{70.28}} & \tiny{67.14} & \tiny{-3.14} & \textbf{\tiny{59.38}} & \tiny{56.50} & \tiny{-2.88}\\\tiny{fog} & \textbf{\tiny{65.32}} & \tiny{60.14} & \tiny{-5.18} & \textbf{\tiny{59.26}} & \tiny{53.16} & \tiny{-6.10} & \textbf{\tiny{52.22}} & \tiny{45.06} & \tiny{-7.16} & \textbf{\tiny{50.18}} & \tiny{40.38} & \tiny{-9.80} & \textbf{\tiny{38.58}} & \tiny{28.02} & \tiny{-10.56}\\\tiny{frost} & \textbf{\tiny{65.82}} & \tiny{64.04} & \tiny{-1.78} & \textbf{\tiny{50.56}} & \tiny{47.40} & \tiny{-3.16} & \textbf{\tiny{38.68}} & \tiny{36.56} & \tiny{-2.12} & \textbf{\tiny{37.30}} & \tiny{35.72} & \tiny{-1.58} & \textbf{\tiny{29.62}} & \tiny{28.08} & \tiny{-1.54}\\\tiny{gaussian\_blur} & \textbf{\tiny{69.90}} & \tiny{68.24} & \tiny{-1.66} & \textbf{\tiny{57.54}} & \tiny{56.92} & \tiny{-0.62} & \textbf{\tiny{44.44}} & \tiny{43.50} & \tiny{-0.94} & \tiny{27.66} & \textbf{\tiny{29.60}} & \tiny{1.94} & \tiny{11.90} & \textbf{\tiny{13.32}} & \tiny{1.42}\\\tiny{gaussian\_noise} & \textbf{\tiny{64.34}} & \tiny{61.34} & \tiny{-3.00} & \textbf{\tiny{52.40}} & \tiny{45.36} & \tiny{-7.04} & \textbf{\tiny{33.36}} & \tiny{24.46} & \tiny{-8.90} & \textbf{\tiny{15.78}} & \tiny{9.34} & \tiny{-6.44} & \textbf{\tiny{5.52}} & \tiny{3.56} & \tiny{-1.96}\\\tiny{glass\_blur} & \textbf{\tiny{61.94}} & \tiny{61.02} & \tiny{-0.92} & \tiny{49.70} & \textbf{\tiny{51.16}} & \tiny{1.46} & \tiny{32.32} & \textbf{\tiny{35.32}} & \tiny{3.00} & \tiny{27.08} & \textbf{\tiny{29.56}} & \tiny{2.48} & \textbf{\tiny{21.84}} & \tiny{21.04} & \tiny{-0.80}\\\tiny{impulse\_noise} & \textbf{\tiny{60.38}} & \tiny{54.50} & \tiny{-5.88} & \textbf{\tiny{45.02}} & \tiny{38.22} & \tiny{-6.80} & \textbf{\tiny{32.86}} & \tiny{24.56} & \tiny{-8.30} & \textbf{\tiny{14.06}} & \tiny{8.90} & \tiny{-5.16} & \textbf{\tiny{5.32}} & \tiny{3.46} & \tiny{-1.86}\\\tiny{jpeg\_compression} & \textbf{\tiny{67.26}} & \tiny{65.04} & \tiny{-2.22} & \textbf{\tiny{62.96}} & \tiny{62.00} & \tiny{-0.96} & \textbf{\tiny{59.88}} & \tiny{58.68} & \tiny{-1.20} & \textbf{\tiny{50.24}} & \tiny{49.98} & \tiny{-0.26} & \tiny{38.86} & \textbf{\tiny{39.30}} & \tiny{0.44}\\\tiny{motion\_blur} & \textbf{\tiny{70.16}} & \tiny{67.96} & \tiny{-2.20} & \textbf{\tiny{63.10}} & \tiny{61.18} & \tiny{-1.92} & \textbf{\tiny{51.04}} & \tiny{50.54} & \tiny{-0.50} & \textbf{\tiny{38.42}} & \tiny{36.22} & \tiny{-2.20} & \textbf{\tiny{29.78}} & \tiny{27.84} & \tiny{-1.94}\\\tiny{pixelate} & \textbf{\tiny{72.04}} & \tiny{71.36} & \tiny{-0.68} & \textbf{\tiny{71.76}} & \tiny{70.80} & \tiny{-0.96} & \tiny{63.44} & \textbf{\tiny{65.20}} & \tiny{1.76} & \tiny{52.42} & \textbf{\tiny{57.46}} & \tiny{5.04} & \tiny{46.02} & \textbf{\tiny{51.78}} & \tiny{5.76}\\\tiny{saturate} & \textbf{\tiny{64.52}} & \tiny{61.32} & \tiny{-3.20} & \textbf{\tiny{48.36}} & \tiny{44.48} & \tiny{-3.88} & \textbf{\tiny{71.92}} & \tiny{70.70} & \tiny{-1.22} & \tiny{51.48} & \textbf{\tiny{53.02}} & \tiny{1.54} & \tiny{34.58} & \textbf{\tiny{37.38}} & \tiny{2.80}\\\tiny{shot\_noise} & \textbf{\tiny{63.28}} & \tiny{59.42} & \tiny{-3.86} & \textbf{\tiny{50.86}} & \tiny{40.36} & \tiny{-10.50} & \textbf{\tiny{34.32}} & \tiny{21.76} & \tiny{-12.56} & \textbf{\tiny{15.50}} & \tiny{7.68} & \tiny{-7.82} & \textbf{\tiny{7.90}} & \tiny{4.30} & \tiny{-3.60}\\\tiny{snow} & \textbf{\tiny{63.20}} & \tiny{60.36} & \tiny{-2.84} & \textbf{\tiny{46.24}} & \tiny{40.02} & \tiny{-6.22} & \textbf{\tiny{48.48}} & \tiny{42.92} & \tiny{-5.56} & \textbf{\tiny{37.14}} & \tiny{29.66} & \tiny{-7.48} & \textbf{\tiny{29.72}} & \tiny{21.24} & \tiny{-8.48}\\\tiny{spatter} & \textbf{\tiny{75.66}} & \tiny{73.66} & \tiny{-2.00} & \textbf{\tiny{67.54}} & \tiny{64.72} & \tiny{-2.82} & \textbf{\tiny{57.40}} & \tiny{55.48} & \tiny{-1.92} & \textbf{\tiny{55.28}} & \tiny{48.16} & \tiny{-7.12} & \textbf{\tiny{45.18}} & \tiny{37.28} & \tiny{-7.90}\\\tiny{speckle\_noise} & \textbf{\tiny{66.56}} & \tiny{62.96} & \tiny{-3.60} & \textbf{\tiny{61.46}} & \tiny{54.84} & \tiny{-6.62} & \textbf{\tiny{43.86}} & \tiny{30.00} & \tiny{-13.86} & \textbf{\tiny{32.82}} & \tiny{17.80} & \tiny{-15.02} & \textbf{\tiny{21.58}} & \tiny{9.60} & \tiny{-11.98}\\\tiny{zoom\_blur} & \textbf{\tiny{59.88}} & \tiny{57.28} & \tiny{-2.60} & \textbf{\tiny{53.10}} & \tiny{50.08} & \tiny{-3.02} & \textbf{\tiny{50.58}} & \tiny{46.82} & \tiny{-3.76} & \textbf{\tiny{45.46}} & \tiny{41.40} & \tiny{-4.06} & \textbf{\tiny{41.28}} & \tiny{37.60} & \tiny{-3.68}\\\tiny{{{$\Sigma$}}} & \textbf{\tiny{66.69}} & \tiny{63.89} & \tiny{-2.79} & \textbf{\tiny{57.24}} & \tiny{53.36} & \tiny{-3.89} & \textbf{\tiny{49.48}} & \tiny{45.14} & \tiny{-4.34} & \textbf{\tiny{38.09}} & \tiny{34.66} & \tiny{-3.43} & \textbf{\tiny{28.64}} & \tiny{26.02} & \tiny{-2.61}\end{tabular}\caption{Accuracies for EfficientNet with pipelining of AugMix~\cite{hendrycks_augmix_2020} \& OpticsAugment and only OpticsAugment evaluated on ImageNet-100-c 2D common corruptions~\cite{hendrycks_benchmarking_2019}.}\label{tab:tab:imagenet100c_corruptions_EfficientNet}\end{table*}

\end{document}